%% file: main.tex
\pdfoutput=1

\documentclass[11pt]{article}

\usepackage[final]{acl}

\usepackage{times}
\usepackage{latexsym}

\usepackage[T1]{fontenc}

\usepackage[utf8]{inputenc}

\usepackage{microtype}

\usepackage{inconsolata}

\usepackage{graphicx}

\usepackage{multirow}
\usepackage{booktabs}
\usepackage{xspace}
\usepackage{enumerate}
\usepackage{comment}
\usepackage{makecell}
\usepackage{amssymb}
\usepackage{colortbl}  %
\usepackage{xcolor}    %
\usepackage{makecell,booktabs}
\usepackage{makecell}
\usepackage[most]{tcolorbox}
\usepackage{subcaption}

\usepackage{changepage,threeparttable}

\usepackage{listings}

\title{\datasetname: A Legal NLP Benchmark for Assisting with Legal Briefs}

\author{Jesse Woo$^{\dagger *}$ \quad Fateme Hashemi Chaleshtori$^{\ddag *}$  \quad Ana Marasovi\'c$^{\ddag **}$ \quad Kenneth Marino$^{\ddag **}$ \\
$^\dagger$NYU School of Law\\
$^\ddag$University of Utah\\
\texttt{\small jesse.woo@nyu.edu \quad fateme.hashemi@utah.edu \quad ana.marasovic@utah.edu \quad kenneth.marino28@gmail.com}  
}

\newcommand{\datasetname}{\textsc{BriefMe}\xspace}

\providecommand{\sectionvspace}{\vspace{-0.02cm}}

\usepackage{enumitem}

\newlist{compactitem}{enumerate}{3} %
\setlist[compactitem,1]{label=\textbullet, nosep, leftmargin=*}
\newlist{indenteditem}{enumerate}{3}
\setlist[indenteditem,1]{label=\textbullet, nosep, leftmargin=2em}

\newcommand\sect[1]{\S\ref{#1}}

\begin{document}
\maketitle

\begingroup
\let\thefootnote\relax\footnotetext{$^{*}$Equal contribution as first authors. Portions of this work were conducted by Jesse Woo while at Columbia University.}
\let\thefootnote\relax\footnotetext{$^{**}$Equal contribution as last authors.}
\endgroup

\begin{abstract}
\sectionvspace
A core part of legal work that has been under-explored in Legal NLP is the writing and editing of \textit{legal briefs}. This requires not only a thorough understanding of the law of a jurisdiction, from judgments to statutes, but also the ability to make new arguments to try to expand the law in a new direction and make novel and creative arguments that are persuasive to judges. To capture and evaluate these legal skills in language models, we introduce 
\datasetname, a new dataset focused on legal briefs. It contains three tasks for language models to assist legal professionals in writing briefs: \textit{argument summarization}, \textit{argument completion}, and \textit{case retrieval}. In this work, we describe the creation of these tasks, analyze them, and show how current models perform. We see that today's large language models (LLMs) are already quite good at the summarization and guided completion tasks, even beating human-generated headings. 
Yet, they perform poorly on other tasks in our benchmark: realistic argument completion and retrieving relevant legal cases. 
We hope this dataset encourages more development in Legal NLP in ways that will specifically aid people in performing legal work. 
\end{abstract}

\sectionvspace
\section{Introduction}
\sectionvspace

\input{figures/tasks_sample_fig}

Legal work involves reading, writing and reasoning about large volumes of text \cite{Zhong2020LegalAI}. %
For instance, in litigation, attorneys must search and interpret prior case law to construct convincing arguments  
to stake out a strong position. 
They present these to a judge in the form of a document called a \emph{brief}. %
Given their general text processing and generation strengths, large language models (LLMs) could assist legal professionals with brief drafting. %
To assess this, we develop a suite of three tasks to test LLMs' capabilities in supporting brief writing tasks: (1) extreme summarizing of an argument in a brief section, (2) completing a missing argument, and (3) retrieving a missing case citation.

Most prior work in Legal NLP focuses on judicial opinions\,---\,documents that present the court's reasoning as a synthesis of various legal considerations, ultimately framing the outcome as the natural conclusion given the legal landscape. 
We focus instead on briefs, which encapsulate the core of an attorney's work and thought process in advocating for a particular interpretation of the law,  especially at the appeals stage. %
This is more appropriate if the goal is to develop LLMs that assist attorneys.

As seen in Figure \ref{fig:tasks-overview-fig}, briefs are split into multiple sections that constitute sub-arguments about the state of the law as interpreted by that attorney. %
Effective arguments fit together into a cohesive whole that logically walks the judge step-by-step to the attorney's preferred outcome. 
A model that could assist a legal professional in drafting briefs could offer notable productivity gains. %
Although there is a growing number of legal NLP datasets, none contain legal briefs in significant numbers 
or explicitly capture the logic of legal argumentation (see \sect{sec:related}).  

To this end, we present \textbf{\datasetname}, %
a dataset of Supreme Court of the United States (SCOTUS) briefs with annotations to support three tasks. 
We construct \datasetname from briefs hosted on the Supreme Court's website.  
We formulate the tasks based on the structure of SCOTUS briefs. %
This structure follows a convention where each section and subsection state a concise version of the arguments that are contained in that section, with each section building toward the ultimate argument. 
See an example in Figure \ref{fig:tasks-overview-fig}. 

The first task, \emph{argument summarization}, requires generating a section heading that captures the essence of presented legal arguments. Doing so assists attorneys in developing and structuring their briefs.  %
The goal of the second task, \emph{argument completion}, is to identify and generate a missing argument in a sequence of brief headings. This, in turn, supports attorneys in developing complete and logically coherent arguments, while also serving as a test of models' ``understanding'' of legal reasoning. %
Finally, \emph{case retrieval} tests a model's ability to identify relevant precedent that supports specific legal claims. Such retrieval ensures that attorneys provide comprehensive citations of prior case law. 

We develop an LLM-as-a-judge using \texttt{o3-mini} instructed based on expert-written guidelines for writing brief headings and tables of contents. Although we carefully recruit human judges, their ratings often lack consistency or include justifications irrelevant to the task. Thus, to assess the reliability of the LLM judge, a qualified author provided meta-ratings of judge performance. We find that the LLM-as-judge is a more reliable evaluator than human judges. We also use it to filter out instances with low-quality headings and table of contents in the final version of \datasetname.

We benchmark \datasetname tasks against a wide variety of current LLMs and retrieval methods. GPT-4o achieves the strongest performance with  average ratings of 4.3 out of 5 for both argument summarization and guided argument completion. In contrast, human-authored headings on the full unfiltered set score lower, averaging 3.4 for summarization and 3.5 for completion, with 43\% and 22\% of examples with scores of 3 and 3.5 or lower, respectively (Fig.~\ref{fig:llm_score_freq} in Appendix shows the distribution of scores). According to the judge, common issues include lack of persuasive phrasing and weak logical integration with surrounding headings. GPT-4o's outputs, on average, require only minor edits to improve clarity. Notably, LLMs' performance generalizes to briefs published after February 2025. These results suggest LLMs are well-positioned to assist in guided argument completion and summarization. Future work could study teams of experts and LLMs assisting with these drafting tasks. 

However, model performance drops significantly on the other two tasks. \texttt{Llama-3.1-70B} identifies the correct location of a missing argument in only 18\% of cases. For case retrieval, the best-performing method retrieves the correct precedent among its top five results in just 31.5\% of instances. These findings point to the need for further advances before LLMs can reliably support legal professionals in these more complex drafting scenarios.\footnote{The dataset is available on \href{https://huggingface.co/datasets/jw4202/BriefMe}{huggingface}, and scripts collection and benchmarking are available on \href{https://github.com/utahnlp/BriefMe}{github}.}

\sectionvspace
\section{Related Work}
\sectionvspace
\label{sec:related}

Processing legal text is an active area of NLP research that has numerous tasks such as legal judgment prediction~\cite{aletras2016predicting, niklaus2021swiss, luo-etal-2017-learning, xiao2018cail2018, malik-etal-2021-ildc}, legal text classification~\cite{10.1145/3086512.3086515, papaloukas2021multi, chalkidis-etal-2019-large, chalkidis2021multieurlex}, and multiple-choice QA \cite{katz2024gpt, Zhong2020JECQA, CaseHOLD}. %
Below, we compare our work to related tasks.

Legal summarization is an established task in legal NLP with many available resources \cite{hachey2006extractive, DBLP:conf/nips/ShenLYDSD22, malik2021ildc, shukla2022legal, galgani-etal-2012-combining, kornilova-eidelman-2019-billsum, ragazzi2024lawsuit}. %
Our argument summarization task differs from this prior work by focusing on generating forward-looking arguments presented by litigants to persuade a court, rather than backward-looking judicial decisions that explain the court's reasoning. %
This distinction also holds for legal summarization works that aim to summarize other SCOTUS documents, such as CaseSumm \cite{heddaya2024casesumm}, which focuses on opinions rather than briefs.

Our argument completion task does not have a close parallel with another legal NLP task. %
Legal argument mining involves identifying argumentative components in legal texts, such as issues, evidence, and conclusions \cite{ElarabyLitman2022}, and determining how these components relate to one another.
Some approaches use richer annotation schemes to better capture the complexity of legal arguments \cite{Habernal2024}. 
However, this task is orthogonal to our argument completion task, which requires generating a missing argument. 
A broader line of work on AI in debating is related, including \href{https://en.wikipedia.org/wiki/Project_Debater}{IBM's Project Debater} and efforts to improve alignment through debate \cite{DBLP:conf/icml/KhanHVRSRGBRP24}. However, these approaches primarily focus on engaging in debates, rather than completing partial argument structures in legal briefs.

Our case retrieval can be seen as an instance of ``case law'' retrieval where relevant cases need to be retrieved given a query case, question, or some other text ~\cite{locke2018test, kano2019coliee, ma2021lecard, li2024lecardv2, hou2024clerc, xiao2019cail2019,louis2022statutory, mahari-etal-2024-lepard}. %
The task of predicting entailment relation between two cases is also related \cite{kano2019coliee}. %
The distinguishing feature of our case retrieval is that queries are from legal briefs.  
Since briefs are written by lawyers, not judges, their writing style and how they marshal precedent is closer to how most attorneys work. 

Two notable NLP corpora include briefs: \citet{henderson2022pileoflaw} scraped SCOTUS dockets and included briefs in a large pretraining corpus, but the data were not processed for or evaluated on summarization, argument completion, or retrieval tasks. \citet{DBLP:conf/nips/ShenLYDSD22} include briefs in their court filings, but the corpus is smaller, and the briefs are written for trial court motions.

\sectionvspace
\section{Dataset and Task Setup}
\sectionvspace

\input{tables/all_summary_stats}
We introduce \datasetname and formulate three tasks designed to assist lawyers in writing briefs: \textit{argument summarization}, \textit{argument completion}, and \textit{case retrieval}. Our full data extraction and cleaning pipeline is shown in Fig.~\ref{fig:pipeline_fig} in the Appendix. This was used to collect and clean our raw data to produce our three tasks. %

\sectionvspace
\subsection{Data Collection and Pre-Processing}
\sectionvspace
The key data source for \datasetname is Supreme Court of the United States Merit Cases briefs. 
All cases in the dataset were argued before the court from October 2017 to March 2024. The raw brief data in PDF format was downloaded through the \href{https://www.supremecourt.gov}{Court's website}. We choose this data for two reasons. This choice allowed for free public access since many court's dockets are only accessible through per-page charges or expensive subscription services. More importantly, we believe that the legal reasoning in the briefs for the highest court is of high quality as only the most accomplished attorneys practice in front of the Supreme Court and are highly motivated to write persuasive and technically sound briefs. %

\textbf{Automatic PDF extraction.}
SCOTUS maintains a human-navigable website for each case, but not a convenient API for automatic downloading of cases. We built a custom webscraping and text parsing pipeline using \href{https://www.selenium.dev/}{Selenium}, \hyperlink{https://pypi.org/project/PyPDF2/}{pypdf2}, regular expressions, and fuzzy matching to build the dataset.
Refer to Appendix~\ref{sec:data_details} for a detailed description of additional cleaning and parsing steps and the end-to-end extraction pipeline.

At the end of this process, we had a structured representation for our briefs with each section of text now matched to a header. See Table~\ref{tab:doc_statistics} (Appendix) for the properties of the collected brief data and Appendix~\ref{sec:matching} for details of the matching process.
We used the \hyperlink{https://github.com/freelawproject/eyecite}{eyecite}
library to identify case citations and mask citations in the text, and the courtlistener API to construct a corpus of those cited cases. See Appendix~\ref{sec:retrieval_corpus} for details on construction of the retrieval corpus.

\sectionvspace
\subsection{\datasetname Dataset}
\sectionvspace

\input{tables/dataset_comparison_new}

We now describe \datasetname and the three constituent tasks. See Fig.~\ref{fig:tasks-overview-fig} for an example and overview of these tasks, and Table~\ref{tab:all_summary_statistics} for the Train/Dev/Test split for each task. Tables~\ref{tab:summary_statistics}, \ref{tab:completio_summary}, and \ref{tab:citation_summary} in the Appendix  provide additional statistics for each task. %
We compare to other large legal text datasets with similar tasks in Table~\ref{tab:dataset_comparison}. From this, we can see that our dataset is the only one that includes all three of our tasks (or similar tasks) and the only one on briefs. It is also comparatively sized or bigger among argument summarization datasets. It is comparatively sized or bigger among case retrieval datasets as well, with the exception of LePaRD~\cite{mahari-etal-2024-lepard}, which does not include other tasks.

\sectionvspace
\subsection{Tasks}
\sectionvspace

\paragraph{Argument Summarization.}
Generating concise, accurate argument summaries for legal briefs could help with efficient navigation and understanding of complex legal arguments. Section headings in practice are just a very short summary of that part of a brief. Being able to automate this summarization process could save attorneys valuable time and provide a quick automated check that the summary of the text matches the argument they wanted to make.
This task can also be viewed as a way of evaluating models' legal reasoning. The ability to concisely summarize a legal argument from a long text is central to legal work and therefore understood and taught as a core skill attorneys must learn.

We frame the problem of generating section headings as the task of extreme summarization~\cite{narayan-etal-2018-dont},
aiming to capture the main argument of each section in a clear and concise heading. The structure of briefs lets us use the section heading as the gold-standard summary for the corresponding section. For each subsection, the input is the text under its heading, and the heading itself serves as the gold-standard summary. See the top box in Fig.~\ref{fig:tasks-overview-fig} for an example (more examples provided in Appendix~\ref{app:argument_summ_examples}).
We omit subsections containing fewer than 25 words and headings less than 3 words (e.g. just the word ``Security'').%
We find short texts often lack substantive content for meaningful summarization, and unigram and bigram
headings tend to be vague or uninformative.\footnote{Example of short text:
\textbf{Heading:} B. Oracle II’s Cramped Fair Use Analysis Conflicts With the Holdings of This Court as Well as the Ninth and First Circuits 
\newline
\textbf{Text:} The court’s Oracle II decision also conflicts with guidance from this Court and several circuit courts.
}

\textbf{Argument Completion.} Another key skill for attorneys is constructing arguments: starting with certain premises and conclusions and filling in logical gaps. %
The table of contents can be seen as a series of nested arguments. %
There may be a top-level argument and several supporting arguments that flesh out and support it. For instance:

\tcbset{quote style/.style={
    colback=gray!5,
    colframe=gray!50,
    boxrule=0.4pt,
    arc=2pt,
    boxsep=4pt,
    left=6pt,
    right=6pt,
    top=4pt,
    bottom=4pt,
    enhanced,
}}

\vspace{-.15cm}
\begin{tcolorbox}[quote style]
\vspace{-.15cm}
{\small
\begin{compactitem}
    \item I. Content questioning is a vital tool for ensuring a fair trial in high-publicity cases
\end{compactitem}
\begin{compactitem}[label=\textbullet, leftmargin=1.6em]
    \item A. Identifying biased jurors is critical to preserving the right to a fair trial in high-publicity cases.
    \item B. As ABA policy has long recognized, content questioning is a vital tool for identifying biased jurors in high-publicity trials.
    \item C. Content questioning is particularly important in the modern media landscape.
\end{compactitem}
}
\vspace{-.15cm}
\end{tcolorbox}
\vspace{-.15cm}

\normalsize

Thus, to form a task where the model has to fill in missing arguments, we can remove an argument from the table of contents, replace it with a special tag, and then ask the model to fill it back in using the rest of the table of contents as context. %
See the middle box in  Fig.~\ref{fig:tasks-overview-fig} for an example (more examples can be found in Appendix~\ref{app:argument_comp_examples}).
We can create this \textbf{guided} fill-in-the-argument task to test legal reasoning and assist attorneys when they have identified a missing argument in their briefs.\footnote{This task setup is similar to cloze-style tasks, where models choose the correct ending from few options based on given context \cite{mostafazadeh-etal-2017-lsdsem, zellers2019hellaswag}. %
However, in our case, arguments can be dropped at any position, not just at the end of a sequence. %

}

In another, more \textbf{realistic} setup, we evaluate the model's ability to suggest missing headings in a table of contents without explicit placement cues. 
Given an incomplete table, the model proposes headings to be added that could improve clarity, organization, or persuasiveness. 
In this setup, we do not indicate what is missing, so the model may suggest multiple plausible headings to insert if the argumentation in the original human-authored tables of contents was not complete or detailed enough. 
For example, preliminary experiments show that when a major heading has no supporting subheadings, the model often prioritizes adding them. 
Therefore, we evaluate models by checking whether the omitted heading appears among their recommendations,  without requiring it to be the only one. 
We break argument completion in this setup into three steps:
\begin{compactitem}
    \item Determine whether the table of contents could be improved by adding heading(s).
    \item Specify the level of the missing heading(s), major, minor, or subheading
    \item Identify the location(s) for the new headings.
    \item Generate the new heading(s).
\end{compactitem}
To create inputs for this setup, we randomly select and remove one heading from the table of contents. 
We then adjust the numbering of the remaining headings to prevent the model from inferring which heading was removed. 
This renumbering depends on the hierarchical level of the omitted heading. 
When a heading is removed, its child headings are reassigned to the parent above. If no such parent heading exists, they are promoted up one level. Furthermore, any sibling headings that follow the removed one have their numbers decremented by one to maintain a consistent sequence.
Fig.~\ref{fig:step-by-step-argcomp-ex} in Appendix shows an input example for this task.

\textbf{Case Retrieval.}
Legal decisions are based on similar cases from the past to ensure consistency and fairness \cite{ma2023incorporating}. In common law jurisdictions like the United States, prior court precedent is not just helpful guidance, but a form of binding law. Attorneys in litigation must be able to find relevant case law that supports their arguments and cite it in their briefs. The case retrieval task aims to retrieve relevant judicial opinions from a legal corpus \cite{moens2001innovative} to support legal brief arguments. 
The ability to quickly find relevant case law and marshal it in one's argument via a citation is vital to a lawyer. A model that could perform this task accurately and reliably would help legal professionals streamline their work.

For the retrieval task, the data includes a version of the section text with the case citation masked out and replaced by a unique id.
See Appendix~\ref{sec:data_details}, Fig.~\ref{fig:retrieval_fig} for a visualization of this data structure. The id points to a Citation object from the eyecite library that contains a triplet of volume number, reporter name, and page number. This triplet can be used to identify the correct documentation in the retrieval dataset, which is a JSONL file with the full text of every cited case. 
We show statistics of the case retrieval data in Table~\ref{tab:citation_summary} (Appendix). We also show the distribution of citations in Fig.~\ref{fig:citation_distribution} (Appendix), where we can see a classic long-tail distribution where most citations are cited only a few times, but some are cited quite frequently. See the bottom box in Fig.~\ref{fig:tasks-overview-fig} for an example of the task (more examples provided in Appendix~\ref{app:case_ret_examples}).
\sectionvspace
\subsection{Quality Filter}
\sectionvspace
We observe that some human-authored headings are of low quality. 
We address this by developing and using an LLM-as-judge \cite{NEURIPS2023_91f18a12}, described in \sect{sec:llm_judge}, to score and filter such headings for argument summarization and completion tasks.

Human-generated headings score an average of 3.5 out of 5 for argument summarization, according to the judge.\footnote{Score descriptions are in Appendix \ref{sec:appendix_judge_prompts}.} 
For argument completion, we score each table of contents (ToC) by averaging its constituent headings. The average score of human-authored ToCs is 3.7. 
Fig.~\ref{fig:llm_score_freq} shows the score distributions for both tasks.  
We exclude $\approx$43\% summarization samples with headings rated 3 or lower, and $\approx$22\% completion samples with ToC scores below 3.5. 
The resulting filtered dataset is the final \textbf{\datasetname} dataset, but we keep and experiment with the unfiltered version. %

\sectionvspace
\section{Experiments}
\sectionvspace
In this section, we describe the experiments conducted to develop the LLM judge and to benchmark models on the \datasetname tasks. In \sect{sec:llm_judge} we describe the LLM judge, which is a key part of our evaluation protocol. Then we describe the experiments and results for argument summarization \sect{sec:argsum}, argument completion \sect{sec:argcomp}, and case retrieval \sect{sec:caseret}. 

\sectionvspace
\subsection{LLM as Judge}
\sectionvspace
\label{sec:llm_judge}

Automatic evaluation of free-form generated text is challenging, but LLM-based rating or comparison is increasingly common \cite{li-etal-2024-leveraging-large, DBLP:journals/corr/abs-2402-01383}.
We therefore develop an LLM-as-judge for our argument summarization and completion tasks. 

We start by reviewing guidelines on writing effective briefs and headings \cite{dueDiligencePointHeading2017, persuasiveBriefDubose2020, supremeCourtStyleGuideSchweitzer2017, elegantLegalWritingMcCarl2021}.
Using these guidelines, we design instruction sets to prompt the LLM judge to assign scores from 1 (Ineffective) to 5 (Exemplary Legal Argument), along with a justification for its rating. The prompt with detailed descriptions of scoring options is in  Appendix~\ref{sec:model-prompts}.\footnote{We refined the judge instructions using \href{https://claude.ai}{Claude}.} 
We use \texttt{o3-mini-2025-01-31} as the judge.

To evaluate the judge's performance, we conduct small-scale, task-specific human studies comparing the judge's ratings with those of domain experts.\footnote{The human study was approved by our institution’s IRB.}  
We provide experts with the same instructions used for the judge. 
See more details in Appendix~\ref{sec:human_study}.  
Ideally, the judge's ratings should align with the rater consensus on most samples. 
However, we observe two issues: 
(i) limited consensus among ratings (Fig.~\ref{fig:user_variance_dist}, Appendix), and 
(ii) a suspicion that raters may use AI tools to complete the task
We find that some users' justifications are generic and unrelated to specific points in the input, and some others are irrelevant and not acceptable as valid reasoning (examples in Table~\ref{tab:user_justification}, Appendix).

Therefore, we take a different approach to evaluate the judge. 
A qualified author with a relevant background independently review all the ratings from both the expert assessors and the judge. They then assign meta-ratings on a scale from 1 (Unacceptable/Unfair) to 5 (Excellent/Fully Fair).\footnote{To ensure objectivity, we anonymize the data by shuffling the order and paraphrasing the justifications.} 
The results shown in Fig.~\ref{fig:arg-summ-metareview-heatmap} and \ref{fig:arg-comp-metareview-heatmap} (Appendix) demonstrate that the judge's scores consistently receive high meta-ratings (4s and 5s), and on average, they are higher than those for human scores. 
Specifically, in argument summarization, the judge averaged 4.6, compared to a range of meta-ratings for human assessors of 2.4 to 4.2. 
In argument completion, the judge received 4.9, while human assessors' meta-rating ranged from 3.1 to 4.3 (Table~\ref{tab:evaluation_of_raters}, Appendix). 
This suggests the judge's ratings are more reliable for evaluation.

\sectionvspace
\subsection{Argument Summarization}
\sectionvspace
\label{sec:argsum}

\textbf{Models and Controls.}
We categorize the baselines for argument summarization based on whether they (i) rely on shallow heuristics, (ii) follow an extractive approach, or (iii) use an abstractive approach, which is the intended modeling strategy. 

For \textbf{heuristic} baselines, we include one that selects a random sentence from the input text, and another, LEAD-1, that uses the first sentence of the section body as the summary. These sentences set a minimal performance bar and may serve as summaries only if there is bias in the dataset. 

For an \textbf{extractive} baseline, we use a BERT-based extractive approach  \cite{DBLP:journals/corr/abs-1906-04165} configured to extract a single representative sentence as the summary.\footnote{The lead sentence is chosen as summary in 49\% of cases.} 
Comparably high performance between this baseline and abstractive ones suggests the data may not be as abstractive as assumed. 

For \textbf{abstractive} baseline models, we benchmark several LLMs across different model families\,---\,Gemma \cite{team2024gemma}, Llama \cite{DBLP:journals/corr/abs-2407-21783}, Mistral \cite{DBLP:journals/corr/abs-2310-06825}, Qwen \cite{DBLP:journals/corr/abs-2412-15115}, and GPT-4o \cite{DBLP:journals/corr/abs-2410-21276}\,---\,in \emph{zero-shot}, \emph{few-shot}, and \emph{supervised fine-tuning} (SFT) setups.  
With the \emph{zero-shot} setup, we assess whether these models can process Supreme Court briefs without any task-specific training. 
The \emph{few-shot} setup tests whether minimal task-specific demonstrations help models adapt to legal tasks. 
Here, models receive task-specific examples: the section body and its heading as the gold output. 
We ensure diversity in sample lengths to provide balanced demonstrations. We include 3 examples per task (Table~\ref{tab:arg_summ_fewshot_ex}; Appendix).
For both zero-shot and few-shot prompting, we use the same instruction (Appendix~\ref{sec:model-prompts}).
Finally, \emph{SFT} on \datasetname's train split examines whether extensive domain-specific fine-tuning improves performance and whether fine-tuned compact models can outperform prompting a larger proprietary model.

\textbf{Evaluation Measurements.}
In addition to the LLM judge (\sect{sec:llm_judge}), we report perplexity and lexical-overlap automated metrics: BLEU~\cite{papineni2002bleu}, ROUGE~\cite{lin2004rouge}, and METEOR~\cite{banerjee2005meteor}, and  embeddings-similarity metrics: BERTScore~\cite{zhang2019bertscore} and LegalBERT~\cite{chalkidis2020legal}\footnote{Perplexity is calculated under \texttt{Llama-3.1-70B-it} model.}. Because these metrics were generally less reliable, we report results on these metrics in the Appendix.
We also report the SummaC Score \cite{laban-etal-2022-summac}, an entailment-based metric for measuring the factuality of the summaries relative to a given text.\footnote{We use the SummaC-Conv model \texttt{vitc} with granularity at the sentence level and the paragraph level.}

\input{tables/arg_summ_and_comp_combined}

\textbf{Results.}
Table~\ref{tab:compact_results} (left) shows the results of the best-performing model for each family on the test split of \datasetname and Table~\ref{tab:arg_summ_results} (Appendix~\ref{sec:additional-results}) shows the full results. 
No single model consistently outperforms the others across all metrics. 
Although some metrics suggest possible gains from in-context learning or fine-tuning, the LLM judge scores 
show that sincorporating training examples in the few-shot setup results in the reduction rate of the zero-shot score of 39\% on average, with the exception of GPT-4o, which shows a slight improvement. 
Fine-tuning has a milder effect, leading to a 6\% reduction of the zero-shot setting.

Human headings are, on average, rated 4.0 (Strong) by the judge, though this overestimates their quality as we already filtered out those rated below 4.  
In comparison, few-shot \texttt{gpt-4o-2024-08-06}'s generated headings are rated with 4.3. 
One might suspect that the filtered examples are more difficult for people and thus may be for models too, but this is not the case. \texttt{gpt-4o-2024-08-06} achieves high ratings even on the unfiltered set; see Table~\ref{tab:arg_summ_realistic} (Appendix).  
This suggests that the primary avenue for improvement in this task is refining generated headings to an exemplary level.  
A model that regularly performs above this level could offer real assistance to practicing attorneys.

Another notable difference between human- and LLM-generated headings emerges in the SummaC score, where several models surpass legal experts. 
SummaC evaluates summary faithfulness by measuring entailment between the section and the candidate heading, with higher scores indicating fewer semantic contradictions. 
This suggests that some LLMs generate fewer semantic inconsistencies as judged by SummaC, which may be conservative in flagging certain valid rephrasing as non-entailed. However, higher SummaC scores do not equate to superior overall quality or contextual fit.

\textbf{Contamination/Generalization Analysis.} We further investigate whether the LLMs' strong results on our dataset reflect authentic reasoning or simply exposure to similar examples during pretraining. 
Publicly available briefs are likely to have appeared in the models' training corpora, introducing a risk of data contamination that could allow the models to rely on memorized patterns rather than true task understanding.  
To address this, we assembled a held-out evaluation set comprised of 168 briefs published in Feb.\ 2025 or later, after each model's training cutoff date, ensuring clear temporal separation between training data and test instances.
By preserving the same task format as our primary dataset but eliminating any overlap with the models' pretraining materials, this held-out set enables a clean assessment of genuine comprehension versus potential leakage.
Table~\ref{tab:arg_summ_ood} (Appendix) reports argument summarization results on this held-out set.
Compared to \datasetname, LLM judge scores are highly consistent, with an average of only 0.7\% drop. 
The overall performance patterns closely mirror those observed with \datasetname across similar trends across zero-shot, few-shot, and fine-tuning setups. As before, \texttt{gpt-4o-2024-08-06} with few-shot examples achieves the highest score, outperforming even human-generated headings by a notable margin of 0.8 points (4.3 vs.\ 3.5).
These results suggest that the model’s success is not merely the result of memorization but rather indicates a capacity for generalization and reasoning in the argument summarization task.

\textbf{Error Analysis.} %
We review 105 argument summary generations between the top 3 scoring models. The reviewing author finds that between summaries rated 4 or 5, it is difficult to judge definitively that the lower rated summary is worse than the higher rated one, as they are both of high quality. However, they are also clearly better than those rated 3 or below because they reference specific legal doctrines or principles and state arguments concisely and assertively. This indicates that although the difference in the judge's rating between a 4 and a 5 may be small, and in some cases arbitrary, the judge is correctly separating high-quality summaries from lower-quality ones. See Appendix \ref{sec:summ_ea} for more details.

\sectionvspace
\subsection{Argument Completion}
\sectionvspace
\label{sec:argcomp}

\textbf{Models and Controls.}
We categorize the baselines for argument completion similar to argument summarization. 
For a \textbf{heuristic} baseline, we use a randomly selected section heading as the baseline output, for the same reasons as for summarization. 

For \textbf{abstractive} baselines, we adopt the  methodology used for summarization to evaluate the same LLMs on the \textbf{guided} argument completion task. 
In our few‑shot setup, each model receives 3 tables of contents (ToCs) with one heading replaced with a placeholder and specified as the target output (examples in Table~\ref{tab:arg_comp_fewshot_ex}; Appendix). 
For the \textbf{realistic} setup, where the explicit placement of the missing heading is omitted, we only use one open-source model that has a higher LLM judge score on the argument completion task, \texttt{Llama-3.1-70B}.
Model instructions are in Appendix~\ref{sec:model-prompts}. 

\textbf{Evaluation Measurements.}
In the \textbf{guided} setup with placeholders for missing headings, we use the same evaluation metrics as in the argument summarization task (except for SummaC), including the LLM judge for this task (see \sect{sec:llm_judge}). 

In the \textbf{realistic} setup, we use binary classification metrics
to evaluate whether the model detects if the ToC is missing headings. 
For heading level and heading location, we measure the frequency with which the model correctly predicts the heading level 
and its position in the table, respectively.

\textbf{Results.}
Table~\ref{tab:compact_results} (right) shows the \textbf{guided} argument completion results of the best-performing model for each family on the test split of \datasetname and  Table~\ref{tab:arg_comp_results} (Appendix \ref{sec:additional-results}) full results. 
According to LLM judge scores, using training examples in the few-shot setup yields mixed effects, ranging from a relative improvement rate of zero-shot results up to 81\% (\texttt{Llama-3.1-8B}) to a reduction rate of up to 29\% (\texttt{Gemma-2-2B}). 
A similar trend is observed with fine-tuning, where changes in LLM judge scores span from a 78\% increase to a 15\% decrease, compared to the zero-shot setup.

In the more \textbf{realistic} setup, the model perfectly detects whether a ToC misses a heading. 
In a sample of 100, it correctly predicts the heading level 86\% of the time, but succeeds in placing the heading correctly in only 18\% of cases (Table~\ref{tab:step_by_step_arg_comp_results}). 
This discrepancy indicates that while the model can effectively complete a ToC when pointed to where a heading is missing, accurately placing it within the document remains a greater challenge. 
The 18 generated headings that are correctly placed within the Table of Contents receive an average score of 2.8 (Satisfactory) from the LLM judge, with none scoring above 3. 
These results suggest that combining all steps, detecting incomplete ToC, identifying the correct structural level and location, and generating the heading poses a significant challenge. 

\textbf{Contamination/Generalization Analysis.} We repeat the contamination check for the guided argument completion task using the same holdout set. 
Results are reported in Table~\ref{tab:arg_comp_ood} (Appendix).
We observe a more pronounced (yet small) performance decline of 4.5\% in LLM judge scores compared to only 0.7\% for argument summarization, indicating that argument completion presents greater generalization challenges. 
Despite this, the top-performing models, \texttt{Qwen-2.5-32b} (zero-shot), \texttt{Llama-3.1-70B} (zero-shot), and \texttt{gpt-4o-2024-08-06} (few-shot), maintain strong performance that continues to exceed human headings, demonstrating that these LLMs retain substantial capacity for complex tasks even when evaluated on temporally separated data that eliminates potential training set overlap.

\textbf{Error Analysis.} %
We review over 147 argument completion generations between the top 3 performing models. Low-rated generations (scored 1 or 2) were nonsensical or contained little useful information. The reviewing author finds that what separates mediocre generations (scored 3) from high-quality ones (scored 4 or 5) is a lack of legal precision or clear fit with the logic of the argument, suggesting that, at the very least, the LLM judge is aligned with that author's judgment. See Appendix \ref{sec:comp_ea} for more details.

\input{tables/arg_com_stepbystep_results}

\sectionvspace
\subsection{Case Retrieval}
\sectionvspace
\label{sec:caseret}

\textbf{Models.}
We categorize the baselines for case retrieval based on whether they use (i) lexical overlap or (ii) dense retrieval based on embeddings.  
For the traditional \textbf{lexical-overlap} retrieval baseline, we use BM25~\cite{robertson1994some}, which remains a strong baseline for legal retrieval tasks~\cite{DBLP:journals/corr/abs-2105-05686}. %
We test several \textbf{dense} retrieval models: DPR \cite{karpukhin-etal-2020-dense}, ColBERT~\cite{khattab2020colbert}, SAILER~\cite{DBLP:conf/sigir/LiACDW0CT23}, and CaseEncoder~\cite{ma-etal-2023-caseencoder} in \emph{zero-shot} setups. 
We additionally fine-tune DPR and ColBERT models on \datasetname train.
If multiple citations appear within this window, we retain all unchanged, except for the target citation, which is masked, and sentences are not split or trimmed, preserving the citation’s full discourse context. We segment corpus documents into smaller parts for CaseEncoder and SAILER. Details in Appendix~\ref{sec:additiona_exp_case_ret}.

Reranking usually helps improve search quality by refining the order of initially retrieved documents based on more accurate relevance signals \cite{DBLP:journals/corr/abs-1901-04085}. 
We use a cross-encoder model, \texttt{ms-marco-MiniLM-L12-v2} \cite{reimers-gurevych-2019-sentence}, for reranking. 
Given the length of legal documents, which often exceed model input limits, we segment each document into non-overlapping 512-token chunks. 
Each chunk is individually scored by the cross-encoder, and the highest score among a document's chunks is assigned as its final relevance score. 
In cases of score ties, the original retrieval order is used to break ties.

\textbf{Evaluation Measurements.}
We test the baselines using standard information retrieval metrics:
\begin{compactitem}[noitemsep, topsep=0pt]
  \item \emph{Mean Reciprocal Rank (MRR)}: Measures the rank of the first relevant document
  \item \emph{Normalized Discounted Cumulative Gain (nDCG)}: Evaluates the ranking quality based on the position of relevant documents
  \item \emph{Recall@k}: Determines the proportion of relevant documents retrieved in the top-k results.
\end{compactitem}

\textbf{Results.}
Table~\ref{tab:case_ret_results} show the case retrieval results on the \datasetname test split. 
Off-the-shelf models struggle to achieve high performance on this domain-specific data, with ColBERT and BM25 emerging as top performers in the zero-shot setup without reranking. 
Reranking enhances performance across all models, though the magnitude of improvement varies, with DPR showing the most significant gain, a 4.5x increase on the base model (no fine-tuning), albeit from a low start, while SAILER and CaseEncoder see marginal gains.
Fine-tuning further boosts performance for both DPR and ColBERT. 
ColBERT consistently performs best, especially after fine-tuning and reranking. 
This suggests that task-specific training and reranking for neural retrievers are helpful in this domain, and traditional methods (BM25) maintain strong baselines.

Additionally, Table~\ref{tab:case_ret_results_len_based} (Appendix) compares retrieval performance across three models (BM25, DPR, and ColBERT) segmented by query length bins.
BM25 shows an increase in R@10 as the query length grows, with one exception in queries of length 90 to 119 tokens.
Such a trend does not hold for ColBERT and DPR.
Looking at R@10 and MRR@10, BM25 outperforms ColBERT for shorter queries (less than 30 tokens) and longer queries (more than 150 tokens).
Overall, BM25 and ColBERT exhibit complementary strengths across different query lengths, and DPR consistently underperforms, particularly for the longer queries with more than 150 tokens.

\input{tables/case_ret_results}

\textbf{Error Analysis.} %
We review 80 retrieval results between BM25 and ColBERT, the two best-performing models, where neither model successfully retrieved the reference case, and find that over 75\% of the time, both models retrieved at least 1 topically similar case (as judged by the author with legal expertise) to the reference case. In those cases, the models are performing an important aspect of the retrieval task (identifying the correct topic) even if they fail to find the reference case. See Appendix \ref{sec:retr_ea} for additional details.

\sectionvspace
\section{Conclusions}
\sectionvspace

We present \datasetname, a dataset designed to support the development of models that assist in drafting legal briefs.
We outline the procedure for collecting this dataset from the Web, formulate key tasks, and evaluate strong baselines in fully fine-tuned, few-shot, and zero-shot settings.
Our findings suggest that \datasetname could enable models that assist with argument summarization and completion by generating headings that surpass the quality of current human-written ones while also driving much-needed advances in case retrieval.

\sectionvspace
\section{Limitations}
\sectionvspace
One limitation of \datasetname is that the documents are only in English and primarily address U.S.\ law. 
This likely limits its utility beyond English-speaking jurisdictions. In addition, the briefs are limited to SCOTUS cases, and the Court only hears a small fraction of the overall cases in the U.S. in a given year. SCOTUS cases also skew toward certain topics, such as administrative and constitutional law, and away from prosaic topics such as common contract disputes. Further, the Court does not hear matters of pure state law, which would exclude a significant amount of law.

Another possible limitation is the difficulty of evaluating open-ended generation, especially on difficult legal topics. For our argument summarization and completion tasks, our models generate section headings open-ended as this is a useful assistive task for briefs. However, this makes evaluation difficult. %
The ideal evaluation would be to have a few legal experts rate each model response individually, but this is %
infeasible under most researchers' budgets (including ours), so we present a more tractable evaluation. %

\sectionvspace
\section{Ethics Statement}
\sectionvspace
In this work, we propose that LLMs could assist legal professionals with drafting briefs. Importantly, we do not advocate for fully automating the proposed tasks: argument summarization and completion, and case retrieval. Human oversight remains crucial: proposed summaries, completions, and retrieved cases must be critically reviewed and finalized by legal professionals. Unfortunately, over-reliance on LLMs is already a documented issue. 
For instance, an attorney famously submitted fake cases hallucinated by ChatGPT to a court. %
LLMs have also been shown to persuasively push wrong narratives \cite{salvi2025persuasiveness}. In assisting brief writing settings, this raises the risk that LLMs could encourage attorneys to write briefs that effectively mislead other people in violation of their professional duty of honesty. This could complicate the work of judges who assess presented arguments. If misled, this could lead to seriously harmful downstream consequences. 
Additionally, there are concerns about data privacy and confidentiality, as legal documents may contain sensitive information, and individuals should be careful with how they use this day, for instance to train models. %
Finally, there are also broader concerns about human creativity and expression when using LLMs in assisting various writing tasks \cite{gabriel2024ethics}. Yet, people are already using tools like ChatGPT in legal work. This makes it urgent to study AI-assisted legal brief writing. Our benchmark is a first step toward that goal.

\section{Acknowledgments}

The authors thank reviewers for their work and input, the members of the UtahNLP lab for their useful feedback, and our annotators for their contributions. We also thank Debasmita Bhattacharya for additional feedback and Lawrence Leung for help developing an earlier version of this project.

\bibliography{custom}
\clearpage
\appendix

\label{sec:appendix}
\section{Data Examples}
\label{sec:data_ex}
\subsection{Argument Summarization Data Examples}
\label{app:argument_summ_examples}
See Figures~\ref{fig:sum_data_1}, \ref{fig:sum_data_2}, and \ref{fig:sum_data_3} for examples of the inputs and references for the argument summarization task.

\subsection{Argument Completion Data Examples}
\label{app:argument_comp_examples}
See Figures~\ref{fig:com_data_1},\ref{fig:com_data_2},\ref{fig:com_data_3}, and \ref{fig:com_data_4} for example inputs and references for the argument completion task.

\subsection{Case Retrieval Data Examples}
\label{app:case_ret_examples}
See Figures~\ref{fig:cite_data_1},\ref{fig:cite_data_2}, and \ref{fig:cite_data_3} for examples of the data annotation structure for the case retrieval task.

\section{\datasetname Additional Statistics}
\label{sec:sum_stats}
This appendix contains additional summary statistics for the corpus and our three tasks. Table~\ref{tab:doc_statistics} shows overall corpus statistics including the mean number of words per brief, mean number of words per section, number of sections per brief, unique bigrams and trigrams, and automated measures of linguistic complexity. 

Table~\ref{tab:summary_statistics} shows statistics for the argument summarization task, including the number of summarization examples, mean compression ratio, coverage, and formulaicness. 
\citet{grusky2018newsroom}, who introduced the Coverage, Density, and Compression metrics, report them on Newsroom, a dataset of news articles, so a direct comparison is less meaningful. 
Yet, we observe similar Density and Coverage values, while our dataset exhibits a higher median Compression rate ($\sim$1.3x) than Newsroom. 
This is because the headings in our dataset function as summaries, which are shorter than a typical summary. Similarly, \citet{ragazzi2024lawsuit} report the formulaicness of $\sim$12.5\% for the LAWSUIT dataset. 
Our dataset has a lower value (10.4\%), indicating fewer recurring structural patterns; this suggests that a deep understanding of the input is required for the models to perform well.

We do not report informativeness, relevance, and fluency here, as the paper recommends evaluating them on a 60-sample subset with three annotators per sample. This approach would, in our case, require legal experts, but we encountered challenges recruiting reliable annotators for other tasks in this paper.
That said, \citet{grusky2018newsroom} defines informativeness as ``How well does the summary capture the key points of the article?'' and we approximate this metric by using the LLM judge score of human-authored headings in the unfiltered dataset, which averages 3.5 out of 5.

Table~\ref{tab:completio_summary} shows statistics for the argument completion task, including the number of examples, the mean masked header length, and the number of headers at each level (top, mid, and leaf).

Table~\ref{tab:citation_summary} shows statistics for the case retrieval task including the number of examples, the number of cases in the reitrieval corpus, the mean citations per brief and per section, the number of unique citations, and the number of direct quote citations (which we theorize to be easier to retrieve).

\input{tables/dataset_stats}
\input{tables/summary_task_stats}
\input{tables/completion_stats}
\input{tables/retrieval_stats}
\input{figures/citation_distr}

\section{Additional Data Set Details}
\label{sec:data_details}

\subsection{Pipeline}
In Fig.~\ref{fig:pipeline_fig} we show the steps of our data extraction pipeline.
\input{figures/pipeline}

\subsection{PDF Scraping}
\label{sec:scraping}
We constructed this dataset from legal briefs hosted on the \hyperlink{https://www.supremecourt.gov}{Supreme Court's website} in PDF form. The Court maintains a human-navigable website for each case, but not a convenient API for automatic downloading of cases. First, we compiled a list of docket numbers for cases heard by the court from 2017-2024 from \hyperlink{https://caselaw.findlaw.com/court/us-supreme-court/years}{findlaw.com}. The Court uses a consistent endpoint for each docketed case to host relevant filings, so with selenium, we were able to navigate to each case's page. Each docket page contains a table with all the relevant filings. Each row in the table has a description of the filing and a link to the PDF. See Fig.~\ref{fig:scotus_page}.
We used a regex pattern to find rows with descriptions suggesting that they were briefs and exclude certain other false positives. See Fig.~\ref{fig:code_ex_1} for relevant code.

With this approach, we scraped 4377 PDF documents. Due to some quirks of the Court's archive, this included documents that were not briefs as well as duplicate briefs. These were identified and removed in the manual cleaning process described in Appendix~\ref{sec:arg_extract}.

\subsection{Text Extraction}
\label{sec:text_extract}
We experimented with several open-source libraries to extract the text from the PDF, namely pymupdf, pdfminer, and pypdf2. None were perfect, but pypdf2 gave the best results in initial tests. Even so, the text often contained additional spaces randomly inserted within words or characters with non-standard encoding, which made the documents difficult to parse. This made additional cleaning necessary. See Fig.~\ref{fig:code_ex_1} for example cleaning code. Parsing was necessary because attempts at using LLAMA and Mistral to extract the table of contents and section headings, while initially promising, ultimately proved unreliable. Compare Figures~\ref{fig:regex_extract} and 
\ref{fig:mistral_extract} for an example of the issues with model extraction.

\subsection{Argument Extraction and Cleaning}
\label{sec:arg_extract}
We split the document into a table of contents (ToC) and main body with regex patterns that matched words that commonly bookended the table. See Fig.~\ref{fig:code_ex_2} for the regex patterns. The patterns used to divide these were not always consistent, so we reviewed the results and rewrote the regexes several times to catch as many cases as possible. This process involved a great deal of trial and error, because while convention dictates certain patterns when formatting a brief, these patterns are not always strictly followed.

We also prompted GPT-3.5-Turbo to extract the arguments from the table of contents and stored those results along with the main body text and ToC. See Fig.~\ref{fig:code_ex_3} for the GPT prompt used.

One of the authors then manually inspected each ToC in a spreadsheet editor. They compared both the regex-extracted ToC and the GPT-extracted arguments to the original ToC text. If the GPT-extracted arguments matched exactly, the author used those arguments. When GPT-3.5-Turbo did not work properly, which was frequently, they used the regex-extracted ToC after manually cleaning the text of extra whitespace characters and encoding errors. If neither method of extraction worked then the author extracted the arguments from the original text manually. Through this process they also identified additional patterns of encoding errors to fix with regular expressions and eliminated duplicate briefs.

\subsection{Matching Headers and Text}
\label{sec:matching}
Having clean, extracted arguments from each brief, we then matched each header to its corresponding section in the main body of text. We compared the header to each line of text in the main body using FuzzyWuzzy (to account for small variations such as extra spaces, newlines, or idiosyncratic encoding errors), and took the index of the best match as the start of the section. Some briefs had so many errors that they had to be thrown out, but this process ultimately yielded clean arguments matched to sections of text in the main body. 

Random sampling of the sections confirmed that the arguments had been properly extracted and matched with sections. Extraneous newlines and spaces were removed from each section of text using the eyecite clean\_text() function with the all\_whitespace parameter.

\subsection{Retrieval Corpus Construction}
\label{sec:retrieval_corpus}
We compiled a retrieval corpus of every case cited in the briefs by extracting citations with the eyecite library. We used get\_citation() to identify every citation in the text and stored only the FullTextCitation objects (as opposed to the FullLawCitation objects, which represented statutes, for example). We then looked up the case in courtlistener, an open source legal documents database with API access. Using the volume number, reporter name, and page number from the citation object allowed us to uniquely identify the case as required by the \hyperlink{https://www.courtlistener.com/help/api/rest/}{courtlistener API REST endpoint}. We downloaded these cases  directly into a separate corpus as a JSONL file.

We also used eyecite's annotate\_citations() function to replace each case citation with a unique id for the retrieval task. 

\input{figures/scotus_page}
\input{figures/retrieval}

\input{figures/bertopic}
\input{tables/top2vec}

\section{Topic Modeling}
\label{sec:topics}

In Fig.~\ref{fig:bertopic_5} and Table~\ref{tab:top2vec} we show topic modeling results on \datasetname. We used the standard implementations of BERTopic~\cite{grootendorst2022bertopic} and Top2Vec~\cite{angelov2020top2vec} and did not add any additional stopword. This topic modeling indicates the presence of certain high-profile SCOTUS cases as highly ranked topics in the corpus. The presence of ``abortion'', ``roe'', ``fetus'', and ``unborn'' suggests significant briefing for Dobbs v.\ Jackson, a landmark case about abortion decided in 2021. Terms such as ``Google'', Oracle'', ``software'', and ``copyright'' suggest that the software copyright case Google v.\ Oracle also had many briefs filed.

\section{Reliability of LLM and Human Judges}
\label{sec:human_study}

We design instructions for the LLM judge, \texttt{o3-mini}, to score brief headings as section summaries and as completions of a brief's table of contents (Appendix~\ref{sec:model-prompts}).
The goal of this small human study is to assess alignment between human and model ratings, both using these instruction sets.

\textbf{Qualifications.} We recruit participants via the crowdsourcing platform \href{https://www.prolific.com/}{Prolific}, with degrees in administration and law, who work in legal roles, are fluent in English, and are located in the U.S. 
We screen them with a qualification exam.
For this exam, we create a pool of 30 questions, each presenting users with a Table of Contents from a brief with one missing heading and two heading options to fill in the gap: one is the existing heading from the same brief, and the other is a random heading from a different brief. 
Users are asked to select the heading that best completes the table. 
We adjust the formatting of the headings (e.g., capitalization) to prevent any cues that might reveal the correct choice. Each participant solves 7 samples and qualifies for the main studies only if they solve at least 6 questions correctly. 65\% of the participants passed the exam.

\textbf{Initial Annotation Study.} 
We ask annotators to score the headings provided for a given context on a scale from 1 to 5.
We focus on the argument summarization task and recruit 17 qualified annotators.
Each participant receives seven headings to score. 
Annotators are provided with clear instructions (See Fig.~\ref{fig:arg-summ-inst-old}) on how to score the headings, but are not informed that some of the generations originate from AI models. 
An example of the task is illustrated in Fig.~\ref{fig:arg-summ-ex-old}.

Table~\ref{tab:old_arg_summ_score_user_vs_o3mini} compares human ratings with those of the LLM judge.

\textbf{Annotation Study with Enhanced Rubric and Controls.} 
We replicate the study with two major changes. 
First, we revise the scoring instructions in accordance with established guidelines for writing effective legal brief headings. 
Second, we introduce an additional control: disabling copy-paste functionality.
Given the relative difficulty of the argument completion task, we also provide the conclusion section of the brief as supplementary context. 
To avoid revealing the missing heading by this addition, we select briefs whose conclusions are approximately one paragraph in length.
Also, we adopt a more fine-grained scoring scale for the argument completion task by introducing intermediate points between each level on the original 1-to-5 scale (Fig.~\ref{fig:arg-comp-ex}).

For both the argument summarization and argument completion tasks, we hire three qualified annotators. 

Figures~\ref{fig:arg-summ-inst} and ~\ref{fig:arg-comp-inst} show the instructions provided to the users, and Fig.~\ref{fig:arg-comp-ex} shows an example of the argument completion task. Presentation of information in argument summarization is unchanged compared to the initial annotation study (Fig.~\ref{fig:arg-summ-ex-old}).
Results are discussed in \S\ref{sec:llm_judge}.

\input{figures/step-by-step-arg-comp-ex}

\input{tables/arg_summ_fewshot}

\input{tables/arg_comp_fewshot}

\input{tables/arg_summ_results}

\input{tables/arg_comp_results}

\input{tables/arg_summ_realistic}
\input{tables/arg_comp_realistic}

\input{tables/arg_summ_ood_results}

\input{tables/arg_comp_ood_results}
\input{tables/case_ret_result_len}
\input{figures/score_freq}

\input{figures/argsumm_inst-old}

\input{figures/argsumm_example_study-old}

\input{figures/ArgSumm-Inst-2}

\input{figures/ArgComp-Inst-2}

\input{figures/ArgComp-Conc-Ex}

Both the exam and the main studies started with informed consent.
On average, annotators are compensated at a rate of \$13.7/hr.

\input{tables/evaluation_of_raters}

\input{tables/old_human_exp_compare_with_o3mini}

\input{figures/user_variance_dist}

\input{figures/human_study_user_score_and_metareview-argsumm}

\input{figures/human_study_user_score_and_metareview-argcomp}

\input{tables/user_justification}

\section{Prompting Details}
\label{sec:model-prompts}

This section provides the prompts used in our experiments, both for the argument summarization and completion tasks, and for evaluating the generated headings using an LLM-as-judge to ensure transparency and reproducibility.

\subsection{LLM Prompts for Summarization and Completion Tasks}
Here, we provide the instructions used to prompt LLMs to generate headings in all setups.

\paragraph{Argument Summarization:}

\begin{tcolorbox}[breakable]
{\small \label{prompt:arg-summ-gen}
You are a highly skilled lawyer. Draft a concise, persuasive, and legally precise heading for the provided section text, ensuring it adheres to the following principles: 
The heading must clearly identify the legal issue, state the conclusion, and possibly hint at the key reasoning, all while maintaining an affirmative tone (preferably) and avoiding ambiguity or misleading information. 
It should be written as a \textbf{complete sentence}, spanning one to three lines, and must reflect the section’s main argument with confidence and accuracy. 
Be careful not to order the court to take any actions. 
Avoid generic or vague language, and ensure the heading is easy to understand, informative, memorable, and directly relevant to the section’s content. 
Ignore stylistic issues in the section text. Use ordinary capitalization in the heading.
}
\end{tcolorbox}

\paragraph{Argument Completion:}

\begin{tcolorbox}[breakable]
{\small \label{prompt:arg-comp-gen}
You are a highly skilled lawyer. Your task is to draft a heading that fills a specific gap in the Table of Contents for a Supreme Court brief. Begin by carefully reviewing the surrounding headings to understand the logical flow and structure of the arguments. Identify whether the missing entry is a major heading, minor heading, or subheading, and ensure your proposed heading aligns with the tone, style, and legal reasoning of adjacent headings. Your heading should ideally be a complete sentence in a positive tone that clearly articulates a legal conclusion, possibly supported by specific reasoning, and written with confidence and conviction. Avoid vague or abstract language, and ensure the heading integrates seamlessly into the Table of Contents, maintaining the persuasive and coherent structure of the brief. Use ordinary capitalization in the heading.
}
\end{tcolorbox}

\paragraph{Step-by-step Argument Completion:}

\begin{tcolorbox}[breakable]
{\small \label{prompt:arg-comp-step-by-step}
\# **Guidelines for Assessing Missing Headings in a Supreme Court Brief's Table of Contents (ToC)**

You are a professional lawyer. You are analyzing a Table of Contents (ToC) in a Supreme Court brief that your skilled coworker has asked you to review. Your coworker is unsure about the flow of the arguments in the ToC. 

Your goal is to decide whether adding headings would improve clarity, organization, or persuasiveness. 

Many Tables of Contents can be improved by adding more headings, but this should be done strategically.

\#\# The Heading Hierarchy

The ToC is usually structured into three levels:

1. **Major Headings**

   - Present independent grounds for relief.

   - Directly correspond to the questions presented to the court.

   - Must progress logically without redundancy.

2. **Minor Headings**

   - Develop and support the arguments of their respective major headings.

   - Include specific legal reasoning, claim elements, or analytical factors.

   - Should comprehensively build the case for the parent major heading.

3. **Subheadings**

   - Provide further detail under minor headings.

   - Offer additional support and context for the arguments made at the minor level.

\#\# Assessment Framework

You might decide to add one or more headings at these levels: major, minor, or subheading.

Here's how you can approach determining if adding headings is needed or not, and in case it is needed, where to add it:

\textbf{***For Major Headings:***}

Assess Completeness of Core Issues:

Review whether the current major headings collectively capture all independent grounds for relief and key legal questions. 

Look for any critical argument or legal issue that is discussed within the minor headings and subheadings under it that might benefit from being highlighted as its own major category.

Examine Logical Progression:

Ensure that the headings advance the narrative clearly and without overlap. If an important theme is buried within several minor headings, consider whether elevating it would better guide the reader. If a group of consequent major headings does not seem parallel to another major heading, consider downgrading them to a minor heading and add an umbrella major heading.

Audience Navigation:

Consider the reader’s perspective: would new major headings help locate pivotal arguments or clarify the relief sequence? Reflect on whether the added headings would improve overall persuasiveness and coherence.

\textbf{***For Minor Headings:***}

Refine Argument Structure:

Look at each major heading’s supporting arguments under it. Determine if any nuanced line of reasoning, legal claim element, or analytical factor is underdeveloped or amalgamated to obscure its importance.

Isolate Key Reasoning:

Analyze the subheadings. Consider if separating a specific argument or legal rationale into its own minor heading could enhance clarity without fragmenting the discussion. Balance this need against the risk of over-complicating the structure.

The same principles used to determine if a Major heading is necessary based on the headings it encapsulates (Assess Completeness of Core Issues, Examine Logical Progression, and Audience Navigation) also apply in this case to determine if a Minor heading is necessary.

\textbf{***For Subheading:***}

Enhance Detail and Context:

Identify whether any minor heading contains complex or layered details that could benefit from further subdivision. 

A subheading might be helpful if it isolates a crucial fact, legal precedent, or analytic point that adds depth without cluttering the main argument.

Maintain Readability:

Evaluate if adding subheadings would help clarify the argument flow by breaking down intricate details or if it might simply over-segment a clear line of reasoning.

Avoid Unnecessary Complexity:

Consider whether the current minor headings already convey sufficient detail. 

The goal is to enhance persuasiveness by spotlighting important nuances, not to overwhelm the reader with excessive subdivision.

\textbf{\#\# Confirmation}

Ask yourself:

1. Would adding headings (at any level) improve the ToC's effectiveness? 

- If yes → Some headings are missing, and you need to report the level of the heading as well.

- If no → The ToC is complete as is

2. (In case the answer to the previous question is yes) Where could I add headings (at any level) that improve the ToC's effectiveness?

To confirm your choice for each heading you decide to add, check the following:

- Does it highlight a missing but important argument?

- Does it improve the logical flow or reader navigation?

- Does it clarify an argument without unnecessary fragmentation?

3. Now that the locations are identified, the next step is generating a recommendation for the missing headings. Consider the following points when generating each of the new headings:

- What purpose does this new heading serve?

- How does this heading contribute to the overall structure and readability of the ToC?

- How to ensure alignment with the surrounding content and maintain logical progression?

- How to make it specific and descriptive enough to guide the reader effectively?

\textbf{\#\# Response Format}

Provide your assessment in JSON format:

\verb|```|json

\{

    "binary\_verdict": "<your final determination>",

    "missing\_headings": [

        \{

            "heading\_level": "<major heading / minor heading / subheading>",

            "heading\_after": "<heading immidiately after the missing heading>",

            "heading\_before": "<heading immidiately before the missing heading>",

            "explanations": "<detailed reasoning for your evaluation>",

            "new\_heading": "<new suggested heading to be added>"

        \},

        \{

            "heading\_level": "<major heading / minor heading / subheading>",

            "heading\_after": "<heading immidiately after the missing heading>",

            "heading\_before": "<heading immidiately before the missing heading>",

            "explanations": "<detailed reasoning for your evaluation>",

            "new\_heading": "<new suggested heading to be added>"

        \},

        \{

            "heading\_level": "<major heading / minor heading / subheading>",

            "heading\_after": "<heading immidiately after the missing heading>",

            "heading\_before": "<heading immidiately before the missing heading>",

            "explanations": "<detailed reasoning for your evaluation>",

            "new\_heading": "<new suggested heading to be added>"

        \},

        ...

    ]

\}

\verb|```|

The binary\_verdict must be either:

- "No, the ToC is complete."

- "Yes, the ToC is missing heading"

If the answer is No, put "null" for heading\_level, heading\_before, and heading\_after.

If the answer is Yes, you need to record all the details about each one of the missing headings: 

- You should determine the level of the heading that, once added, can enhance the ToC. In this case, the heading\_level can be one of <major heading / minor heading / subheading>.

- Next, you should record the heading that comes immediately before the presumed missing heading and the heading that comes immediately after the presumed missing heading.

}
\end{tcolorbox}

\subsection{LLM-as-a-Judge}
\label{sec:appendix_judge_prompts}

Below are the prompts we use to prompt the LLM judge, \texttt{o3-mini}, to score headings in the two legal summarization tasks.

\paragraph{Argument summarization task:}

\begin{tcolorbox}[breakable]
{\small \label{prompt:arg-summ-judge}
\# Instructions for Evaluating Supreme Court Brief Section Headings

\textbf{\#\# Objective}

Headings in a legal brief parse out the exact legal issue and should offer the reader a persuasive overview of the argument provided in the section text. Given a section text, your task is to evaluate the proposed section heading in Supreme Court briefs on a scale of 1-5 based on their effectiveness as tools of legal advocacy. These headings appear in documents written by attorneys to present legal arguments to the Supreme Court, including Justices and Court attorneys, who are sophisticated readers familiar with legal terminology.

\textbf{\#\# Key Principles}

Clarity and Relevance:

The heading must organize the argument and directly relate to the paragraph it introduces. It should do more than offer generic topical abstractive terms. Even if a reader only skims it, they should immediately grasp what the section will cover.

Informative Content:

A strong heading identifies the applicable rule, states the writer’s conclusion, and hints at the key facts and reasoning. It should outline the argument without ordering the court to take action.

Affirmative Tone:

Whenever possible, a heading should use a positive, assertive tone. Affirmative statements tend to be more convincing than those phrased negatively. Also, headings written in all-caps are very hard to follow.

Precision:

While not every detail can be included, aim to craft a heading that is both memorable and precise—using powerful wording and focusing on the facts that matter. Avoid superfluous adjectives and adverbs that only serve to exaggerate. Headings should be about the case before the court, not abstract pronouncements about the law.

Length:

Ideally, most headings should be written as complete sentences, typically spanning one to four lines. The quality of the heading is determined by its clarity and impact, not its length.

Minimizing Ambiguity:

The heading should eliminate any potential for misunderstanding. It must clearly set the stage for the arguments, position, conclusion, and reasons detailed in the text below.

Confidence and Credibility:

Headings should be written with conviction and confidence; hesitation can undermine writer’s credibility and weaken the argument. Every heading should be meticulously accurate and honest, fostering trust with the reader.

Wisdom Over Cleverness:

Effective legal writing reflects deep understanding and sound reasoning rather than mere clever phrasing. A wise heading is both persuasive and just, clearly explaining why the proposed argument is the correct course of action.

\textbf{\#\# Illustrative Examples}

\#\#\# Here are some examples of bad headings. Consider why each fails to inform or persuade the reader:

“ISSUE 5: Whether a genuine issue of material fact on Appellant's contract claim precluded the grant of summary judgment.”

→ This heading poses a question when it should convey a conclusion. 

“THE INVALIDITY JUDGMENT SHOULD BE REVERSED”

→ This heading conveys a conclusion but does not explain why.

“DUPONT'S INTERNAL DOCUMENTS”

→ The heading is vague and uninformative. It does not provide a clear direction for the argument.

“DEFENDANT FAILS TO SATISFY THE RULE 12(b)(6) STANDARD”

→ This heading is not informative. It introduces the issue but does not explicitly provide reasoning to support the conclusion.

Preliminary inquiries of a wounded citizen concerning the perpetrator and circumstances of the shooting are nontestimonial because “made under circumstances objectively indicating that the primary purpose of the interrogation is to enable police assistance to meet an ongoing emergency,” that emergency including not only aid to a wounded victim, but also the prompt identification and apprehension of an apparently violent and dangerous individual.

→ The heading is clear, legally grounded, and persuasive. However, it is too long and complex. 

\#\#\# Here are some examples of good headings:

“The cases Equity discusses — none of which involve an adhesion contract — are irrelevant.”

→ This heading is strong, clear, persuasive, and informative.

“The district court erred in certifying the settlement class because Rule 23(a)(4)’s adequacy requirements were not satisfied; the zero-recovery subclass required separate representation.”

→ The heading clearly states a legal conclusion and provides a specific reason. It effectively structures the argument by tying the court’s error to a specific legal principle and consequence.

“Charges imposed only upon breach are not ‘options for alternative performance’ because breach is not performance.”

“The securities convictions should be reversed because the prosecution failed to offer sufficient evidence of a material GAAP violation.”

“The District Court’s erroneous jury instruction requires a new trial on the healthcare fraud counts.”

“TransWeb did not prove that plasma-fluorinated polymeric material was in public use before the critical date.”

{\color{purple}
\textbf{\#\# Score Scale}

\textbf{5 - Exceptional}
}

Provides a masterful, persuasive overview of the precise legal issue discussed in the section text

Uses confident language that compels agreement without exaggeration

Exceptionally clear and well-organized, easy to read and understand

Fully identifies a legal conclusion and provides the specific reason(s) with persuasive language

Sets a benchmark for persuasive advocacy and structural coherence

{\color{purple}
\textbf{4 - Strong}
}

Clearly articulates legal issue with specific, relevant facts and reasons

Maintains proper terminology while remaining accessible

Advances argument rather than merely summarizing content

Minor refinements could further enhance clarity or coherence

{\color{purple}
\textbf{3 - Adequate}
}

Identifies legal issue but lacks optimal persuasive phrasing

Contains relevant information, though it may feel generic, without standout clarity or impact

Meets the minimum requirements in tone and structure but lacks a compelling, persuasive edge

{\color{purple}
\textbf{2 - Weak}
}

Lacks clarity in identifying key legal elements and does not strongly support the argument

Uses language that is overly abstract, vague, or hesitant that doesn't preview specific argument

Functions as content label with minimal persuasive value

Requires significant revisions in detail and structure to guide the reader effectively

{\color{purple}
\textbf{1 - Ineffective}
}

Misidentifies legal issue or misrepresents section content

Contains confusing or misleading language

Uses improper terminology and undermines argument credibility

Fails to communicate the legal issue or provide a persuasive argument clearly

Topical or abstract discussions and hesitant language

Needs a complete overhaul to fulfill the role of a guiding heading in the brief

\textbf{\#\# Evaluation Process}

Step 1: Review the Section Text

Read the entire section carefully to understand its legal issue and argument.

Identify key facts, legal principles, and the overall persuasive message.

Note any specific elements that must be reflected in the heading.

Step 2: Examine the Proposed Heading: 

Read the heading in isolation to gauge its immediate clarity.

Check that the language is precise and employs proper legal terminology.

Try asking:

Does it just summarize content, or does it show how the section advances the argument?

Would a reader understand the section's strategic purpose from the heading alone?

Does the heading preview the section's analytical contribution?

Are any cited authorities or facts actually central to the section's analysis?

Does it help the court understand why this section matters to their decision?

Step 3: Assess Relevance, Accuracy, Clarity, and Persuasiveness

Determine if the heading accurately reflects the content of the section.

Ensure that it previews the argument and legal issues detailed in the text.

Verify that the heading does not include extraneous or misleading information.

Check for clear persuasive phrasing that enables quick comprehension.

Assess whether the heading uses an appropriate tone to assert the position.

Determine if it highlights key facts or legal conclusions effectively.

\textbf{\#\# Output Format}

Structure your entire output in JSON format as follows:

\verb|```|json

\{

    "Overall Analysis and Comments": "comments",

    "Strength": "strengths",

    "Weakness": "Areas for Improvement",

    "Final Verdict": "verdict"

\}

\verb|```|

Where 'verdict' must be one of Exceptional, Strong, Adequate, Weak, or Ineffective.

Remember: The key is distinguishing between text that functions as a structural heading versus text that reads like it was extracted from the brief's body.

}
\end{tcolorbox}

\paragraph{Argument completion task:}

\begin{tcolorbox}[breakable]
{\small \label{prompt:arg-comp-judge}
\# Instructions for Evaluating Missing Headings in Supreme Court Brief Tables of Contents

\textbf{\#\# Overview}

Table of Contents, which is a list of all headings, is often the reader’s introduction to the arguments. The organization of headings lends clarity and structure to the arguments. Headings parse out the exact legal issue and should offer the reader a complete and persuasive overview of the arguments. The Table of Contents should clearly convey what is being argued and why the position is correct. It must be easy to follow and understand and should summarize the argument, but no individual heading needs to do so. 

A **major** heading articulates a complete and independent ground for relief and will correspond to the question presented or the issue before the court. While each stands independent, the writer should arrange them in a logical order without repetition. 

A major heading can be divided into **minor** headings when multiple legal arguments support its conclusion. Minor headings develop the main contention by providing specific reasons, representing claim elements or factors in a “totality of the circumstances” analysis.

A minor heading can be divided into subheadings. The same suggestions for developing minor headings relative to the major heading apply here.

\textbf{\#\# Objective}

Evaluate proposed headings for missing entries in Supreme Court brief Tables of Contents on a scale of 1-5. Given the overview above, the evaluation should focus on how well the proposed text functions as a heading within the Table of Contents structure.

\textbf{\#\# Key Principles}

Headings should be informative. Headings should not merely be topical or abstract discussions of the law. Rather, headings should identify the applicable rule, convey the writer’s conclusion on the issue, and relate legally significant facts and reasoning. However, they should not order the court to do anything. Headings should be about the case before the court, not abstract pronouncements about the law. 

When possible, headings should be written in the affirmative. Such statements are more convincing than those written in the negative. A heading should make a positive point for the position. An effective heading should not be just a word or phrase. 

Headings may be the writer’s only chance to inform a busy judge of the writer’s arguments. So, if the reader only reads the Table of Contents, they should understand the writer's argument.

Random sentences from the brief, even if legally accurate and stylistically consistent, often make ineffective headings.

The proposed heading must show a clear relationship to its surrounding headings, including a major heading, minor headings, or subheadings.

Headings written in all-caps are very hard to follow.

The length of headings should range from one to four lines of type; however, the length of the heading is not correlated with its quality. Most headings should be complete sentences.

Headings should be written with conviction and confidence. If the writer lacks conviction, the Court will, too. Judges expect advocacy, and hesitation suggests a weak or nonexistent position.

Headings must be meticulously accurate and scrupulously honest in everything, written to cultivate credibility.

Effective legal writing reflects deep understanding and sound reasoning rather than mere clever phrasing. A wise heading is persuasive and just, explaining why the proposed argument is the correct course of action.

\textbf{\#\# Illustrative Examples}

\#\#\# Here are some examples of bad headings. Consider why each fails to inform or persuade the reader:

“ISSUE 5: Whether a genuine issue of material fact on Appellant's contract claim precluded the grant of summary judgment.”

→ This heading poses a question when it should convey a conclusion. 

“THE INVALIDITY JUDGMENT SHOULD BE REVERSED”

→ This heading conveys a conclusion but does not explain why.

“DUPONT'S INTERNAL DOCUMENTS”

→ The heading is vague and uninformative. It does not provide a clear direction for the argument.

“DEFENDANT FAILS TO SATISFY THE RULE 12(b)(6) STANDARD”

→ This heading is not informative. It introduces the issue but does not explicitly provide reasoning to support the conclusion.

Preliminary inquiries of a wounded citizen concerning the perpetrator and circumstances of the shooting are nontestimonial because “made under circumstances objectively indicating that the primary purpose of the interrogation is to enable police assistance to meet an ongoing emergency,” that emergency including not only aid to a wounded victim, but also the prompt identification and apprehension of an apparently violent and dangerous individual.

→ The heading is clear, legally grounded, and persuasive. However, it is too long and complex. 

\#\#\# Here are some examples of good headings:

“The cases Equity discusses — none of which involve an adhesion contract — are irrelevant.”

→ This heading is strong, clear, persuasive, and informative.

“The district court erred in certifying the settlement class because Rule 23(a)(4)’s adequacy requirements were not satisfied; the zero-recovery subclass required separate representation.”

→ The heading clearly states a legal conclusion and provides a specific reason. It effectively structures the argument by tying the court’s error to a specific legal principle and consequence.

“Charges imposed only upon breach are not ‘options for alternative performance’ because breach is not performance.”

“The securities convictions should be reversed because the prosecution failed to offer sufficient evidence of a material GAAP violation.”

“The District Court’s erroneous jury instruction requires a new trial on the healthcare fraud counts.”

“TransWeb did not prove that plasma-fluorinated polymeric material was in public use before the critical date.”

{\color{purple}
\textbf{\#\# Scoring Scale}

\textbf{Score 5: Exemplary}
}

- Exceptionally clear and well-organized, easy to read and understand

- Fully identifies a legal conclusion and provides specific reason(s) with persuasive language

- Seamlessly fills the gap in the Table of Contents, integrating perfectly with surrounding headings

- Sets a benchmark for persuasive advocacy and structural coherence

{\color{purple}
\textbf{Score 4: Strong}
}

- Clearly identifies the legal issue and supports the argument with confident language

- Fills the missing entry effectively, aligning well with adjacent headings in the Table of Contents

- Minor refinements could further enhance clarity or integration

{\color{purple}
\textbf{Score 3: Satisfactory}
}

- Fulfills the basic function of filling the missing entry, though it may feel generic or less integrated with the surrounding headings

- Meets the minimum requirements in tone and structure but lacks a compelling, persuasive edge

- Contributes to the overall structure, albeit without standout clarity or impact

{\color{purple}
\textbf{Score 2: Weak}
}

- Lacks clarity in identifying key legal elements and does not strongly support the argument

- Attempts to fill in the missing entry, but the connection to the overall Table of Contents is weak or ambiguous

- Uses language that is overly abstract or hesitant

- Poor integration with adjacent headings

- Requires significant revisions in detail and structure to guide the reader effectively

{\color{purple}
\textbf{Score 1: Ineffective}

}

- Fails to communicate the legal issue or provide a persuasive argument clearly

- Does not function well as a missing entry, disrupting the flow and coherence of the Table of Contents

- Vague, uninformative, misleading, or disorganized, lacking any effective integration with the overall structure

- Topical or abstract discussions and hesitant language

- Needs a complete overhaul to fulfill the role of a guiding heading in the brief

\textbf{\#\# Evaluation Process}

Step 1: Analyze the Original Table of Contents

- Review the complete Table of Contents to understand the overall argument structure

- Note the logical flow between major headings, minor headings, and subheadings

- Identify whether the missing entry is a major heading, minor heading, or subheading

- Attend to the headings that come before and after the missing entry

- Identify the legal issues, rules, or arguments presented in these surrounding headings

- Note the writing style, tone, and sentence structure used in adjacent headings

Step 2: Assess the Proposed Heading

- Evaluate whether the proposed heading makes a clear, informative legal argument, free of vague or abstract terms

- Ensure it uses clear, persuasive, confident language rather than questions or hesitant statements

- Determine how well the proposed heading bridges the gap between surrounding entries

- Assess whether it maintains the logical progression of the argument

- Check if the proposed heading creates redundancy, repeats other headings, or introduces unrelated content

\#\# Step 3: Apply the Scoring Scale

- Assign a score from 1-5 based on the established criteria

- Consider how well the heading would guide a reader through the argument

- Determine whether the heading would effectively communicate the argument if read in isolation

\#\# Step 4: Provide Specific Feedback

- Identify specific strengths of the proposed heading

- Note any weaknesses or areas for improvement

- Suggest modifications that would elevate the heading's effectiveness

\textbf{\#\# Output Format}

Structure your entire output in JSON format as follows:

\verb|```|json

\{

    "Overall Analysis and Comments": "comments",

    "Strength": "strengths",

    "Weakness": "Areas for Improvement",

    "Final Verdict": "verdict"

\}

```

Where 'verdict' must be one of Exemplary, Strong, Satisfactory, Weak, or Ineffective.

Remember: The key is distinguishing between text that functions as a suitable heading versus text that reads like it was carelessly extracted from the brief.

}
\end{tcolorbox}

\section{Additional Results and Details of the \datasetname Tasks}
\label{sec:additional-results}
This section contains additional experimental details and results for the three \datasetname tasks.

\paragraph{Argument Summarization:}

\begin{compactitem}
    
\item Table~\ref{tab:arg_summ_results} shows the performance of the full set of models tested on the \datasetname test set for argument summarization.

\item Table~\ref{tab:arg_summ_realistic} reports the results for the summarization task on the unfiltered test set.

\item Table~\ref{tab:arg_summ_ood} presents the performance of models on the held-out test set for argument summarization.
\end{compactitem}

\paragraph{Argument Completion:}

\begin{compactitem}
\item Table~\ref{tab:arg_comp_results} shows the results of the full set of models evaluated on the test split of \datasetname.
\item Table~\ref{tab:arg_comp_realistic} reports the results for the argument completion task on an unfiltered test set.
\item Table~\ref{tab:arg_comp_ood} presents the model performances on the held-out set, aiming to test for data contamination.
\end{compactitem}

\paragraph{Case Retrieval:}
\label{sec:additiona_exp_case_ret}

\begin{compactitem}
\item Table~\ref{tab:case_ret_results_len_based} compares the retrieval performance of three models (BM25, DPR, and ColBERT) across different query lengths, showing recall metrics (R@1 through R@100), mean reciprocal rank (MRR\@10), and normalized discounted cumulative gain (nDCG\@10) for queries binned by token count from very short ($leq$ 29 tokens) to very long ($\geq$ 150 tokens), with the percentage distribution of queries in each length category.
\end{compactitem}

\paragraph{Legal Case Segmentation:}

Both the SAILER and CaseEncoder models require documents in the retrieval corpus to be divided into distinct sections.
SAILER expects each legal case to be segmented into five parts: Procedure, Fact, Reasoning, Decision, and Tail.
CaseEncoder, on the other hand, uses three segments: Fact, Holding, and Decision.

The SAILER paper mentions the use of regular expressions for segmentation, but the code is not provided. 
We attempted a similar approach using regular expressions, but it resulted in poor model performance.

As an alternative, we segmented the case documents into individual sentences and then used a large language model (Llama-3.1-70B) to identify the start and end of each section. 
The definitions for these segments are adopted from the SAILER paper. Below, we include the instructions provided to the LLM for segmenting the cases.

\tcbset{breakable style/.style={
    colback=gray!6,
    colframe=gray!150,
    boxrule=1.3pt,
    arc=2pt,
    boxsep=4pt,
    left=6pt,
    right=6pt,
    top=4pt,
    bottom=4pt,
    enhanced,
    breakable,
    before skip=10pt, %
    after skip=10pt 
}}

\begin{tcolorbox}[breakable]
{\small \label{prompt:case-segment}
You are a legal professional. Your job is to segment a legal document into potentially five parts:

1. Procedure: The Procedure section introduces the parties’ information and procedural posture. E.g., 'Plaintiff', 'Defendant', 'On appeal', 'procedural posture'.

2. Fact: The Fact section is a description of the parties’ arguments, evidence, and basic events. 

3. Reasoning: The Reasoning component is the process where the court selects the rules and applies them to the facts. In Reasoning, the judge explains the reasons for the application of the rules. In other words, the events that are relevant to the application of the rules, i.e., the key legal elements, are repeatedly mentioned in this section. 

4. Decision: The Decision section is the specific response given by the court to the legal dispute based on the key facts of the case. The Reasoning section and the Fact section are the basis of the court’s decision. 

5. Tail: The Tails section introduces the basic information about the court, the judge, etc.

The legal document is broken down into smaller units (sentences). Your task is to carefully identify the units that mark the beginning of each segment. Keep in mind that the segments must follow this specific sequence: Procedure, Fact, Reasoning, Decision, and Tail. Therefore, the selected starting units should align with this prescribed order.

Structure your entire output in JSON format as follows:

\verb|```|json

\{

"Start of Procedure": "\#UnitID: Unit text",

"Start of Fact": "\#UnitID: Unit text",

"Start of Reasoning": "\#UnitID: Unit text",

"Start of Decision": "\#UnitID: Unit text",

"Start of Tail": "\#UnitID: Unit text"

\}
\verb|```|
}
\end{tcolorbox}

The same segmentation is then mapped to the structure required by CaseEncoder as follows:
\begin{compactitem}
    \item Procedure and Fact $\,\to\,$ Fact
    \item Reasoning  $\,\to\,$ Holding
    \item Decision and Tail  $\,\to\,$ Decision
\end{compactitem}

Although using LLM-detected segments provides slightly better performance for both models, the results remain unsatisfactory. We hypothesize that more accurate and consistent segmentation is a crucial first step toward improving retrieval quality. Therefore, we exclude both SAILER and CaseEncoder from subsequent experiments involving fine-tuning.

\section{Error Analysis}
\label{sec:error_analysis}

This section describes the error analysis performed by the author with legal expertise. Refer to Table \ref{tab:error_analysis} for a summary of the number of generations reviewed for the summarization and completion tasks. Refer to Table \ref{tab:error_analysis_retr} for a summary of the examples reviewed for the retrieval task.

\input{tables/error_analysis}
\subsection{Summarization Task Error Analysis}
\label{sec:summ_ea}

The author with legal expertise reviewed 30 summaries generated by Llama-3.1-70b (zero-shot) that were scored by the judge with values strictly below 3. They observed the following trends. 

There were 17 of 30 analyzed generations rated a 1 by the judge. 11/17 were incomplete sentences (e.g. ``inal activity after his first offense is therefore an entirely irrelevant inquiry.''). 2/17 were snippets of punctuation (e.g. ``aa.''). 4/17 were citations without context (e.g. ``States, 556 U.S. 568, 570 (2009).'').

There were 13/30 generations rated a 2 by the judge were also often incomplete sentences or citations, but were more likely to relate to the subject matter of the input text. In some cases, the generation related to the input text in such a way that it might serve as a valid summarization, but would contain additional information that was incorrect or not found in the input. Take the following generation as an example:

\begin{tcolorbox}[breakable]
{\small \label{prompt:case-segment}
The circuit courts' disparate interpretations of the scope of liability under Section 10(b) and Rule 10b-5(b) warrant resolution by this Court, as a misstatement that does not meet the threshold for misstatement liability can be a critical 
}
\end{tcolorbox}

The first part of the generation correctly captures the argument of the text, which is that a circuit split over the interpretation of these laws requires Supreme Court review. However, the text does not address misstatement liability. So that part of the generation is extraneous and incorrect.

We hypothesized that the model is repeating memorized information from its parameters from the same or similar text ingested during pretraining, but the results from a held-out set of briefs that were released after the training cutoff date suggest this is not the case.

The reviewing author also examined the summaries of three of the top models (GPT-4o zero-shot, GPT-4o few-shot, and Llama-3.1-70b zero-shot) on 25 section text examples, where the judge scored the summaries a mix of 3, 4, and 5 between the three models. So the author reviewed 75 such summaries in total. This was done to compare and understand the difference between a 4 and 5. For three of these examples, the summaries rated 5 and 4 were nearly identical. Compare the following summaries generated by GPT-4o Zero-shot and Few-shot, and rated a 5 and 4 respectively:

GPT-4o Zero-shot (rated 5): 

\begin{tcolorbox}[breakable]
{\small \label{prompt:case-segment}
the ``usual course'' prong is unconstitutionally vague, lacking clear standards and inviting arbitrary enforcement
}
\end{tcolorbox}

GPT-4o Few-shot (rated 4): 

\begin{tcolorbox}[breakable]
{\small \label{prompt:case-segment}
The ``usual course'' prong is unconstitutionally vague, lacking clear standards and inviting arbitrary enforcement.
}
\end{tcolorbox}

In terms of content, the summaries are identical, and in fact the lower-rated generation has slightly better punctuation.

In the remaining 22 examples, the reviewing author found it difficult to definitively judge that the lower-rated summary was worse than the higher-rated ones. The author judged them both to be of high quality. However, the 75 summaries rated 4 or 5 were clearly better than those rated 3 or below. They referenced specific legal doctrines or principles and stated an argument in a concise and assertive manner. This indicates that although the difference in the LLM judge’s rating of 4 and 5 is small and in some cases arbitrary, the judge is correctly separating high-quality summaries rated 4 or 5 from those of lower quality. Similar dynamics may play out among attorneys, where individual differences in style or opinion can influence one’s assessment of quality.

\subsection{Completion Task Error Analysis}
\label{sec:comp_ea}
The author with legal expertise performed error analysis on argument completion generations from the three top-scoring models from the test set: GPT-4o Zero-shot, GPT-4o Few-shot, and Qwen-2.5-14b (zero-shot). Each model was prompted to complete the argument for 500 examples from the argument completion test set. Only one generation was rated a 1 by the LLM judge. That generation by Qwen-2.5-14b was nonsensical text: 

\begin{tcolorbox}[breakable]
{\small \label{prompt:case-segment}
A. Section 1981\textbackslash z \textbackslash u539f \textbackslash u544a \textbackslash u5fc5 \textbackslash u987b \textbackslash u8bc1 \textbackslash u660e \textbackslash u4f46 \textbackslash u56e0 \textbackslash u56e0 \textbackslash u679c \textbackslash u5173 \textbackslash u7cfb.
}
\end{tcolorbox}

Qwen-2.5-14b had two generations rated a 2 by the LLM judge. One was a very close restatement of another header from the input, and the other verbatim copy. The author reviewed 15 Qwen generations, rated a 3 by the LLM judge. Of those, 7 contained multiple headings in one generation, even though the task was to generate only the missing heading. 1 generation contained extraneous text responding to the prompt but not relevant to the task. The remaining 7 generations were reasonably well-formatted and sounded fluent, but lacked precision or quite fit the logic of the rest of the arguments.

It is more difficult to discern a clear pattern for GPT-4o sero-shot generations. There were 25 headers rated 3 or below (2 rated a 2, and 23 rated a 3). Text length could influence the score, as 2 of the Zero-shot generations rated with a 3 rating were only 3 tokens long (e.g. ``Invasion of Privacy''). In general, headers rated 2 or 3 used less specific legal references. Compare the following two headers generated by GPT-4o zero-shot and few-shot and scored 3 and 4, respectively:

\begin{tcolorbox}[breakable]
{\small \label{prompt:case-segment}
II. section 404 must be interpreted in a manner that respects the principles of fairness and justice inherent in the criminal justice system    
}
\end{tcolorbox}

\begin{tcolorbox}[breakable]
{\small \label{prompt:case-segment}
Section 404 Should Be Interpreted to Promote Fairness and Consistency in Sentencing Adjustments    
}
\end{tcolorbox}

Both headers address the same topic, but the few-shot header references sentencing adjustments, rather than just the criminal justice system generally. 

There were 14 Few-shot headers rated 2 or 3 (1 rated 2, and 13 rated 3), and a similar trend is seen there. Headers that were too short to state an argument were scored poorly (2 out of 14), and less specific headers were rated worse than more specific ones. This is consistent with our prompt to the judge. 

The author also reviewed headers from 3 different models for 30 argument completion examples (90 in total) where the judge gave a mix of scores that included at least one 5. Those models were GPT-4o zero-shot, GPT-4o few-shot, and Qwen-2.5-14b zero-shot. Similar to results from the summarization task, the difference between a 4 and 5 was generally difficult to discern. In two examples, the generations were exactly the same in content except for minor formatting differences, yet were given different scores.

\textbf{Example 1}: \par
GPT-4o Few-shot (rated 4):

\begin{tcolorbox}[breakable]
{\small \label{prompt:case-segment}
The Plan and Its Districts Violate the Equal Protection Clause
}
\end{tcolorbox}

QWEN-14b Zero-shot (rated 5): 

\begin{tcolorbox}[breakable]
{\small \label{prompt:case-segment}
The Plan and Its Districts Violate the Equal Protection Clause
}
\end{tcolorbox}

\textbf{Example 2:} \par
GPT-4o Zero-shot (rated 3): 

\begin{tcolorbox}[breakable]
{\small \label{prompt:case-segment}
petitioners have a significant legal interest in the subject matter of this litigation
}
\end{tcolorbox}

GPT-4o Few-shot (rated 4): 

\begin{tcolorbox}[breakable]
{\small \label{prompt:case-segment}
Petitioners Have A Significant Legal Interest In The Subject Matter Of This Case
}
\end{tcolorbox}

\subsection{Retrieval Task Error Analysis}
\label{sec:retr_ea}

\input{tables/error_analysis_retrieval}

The author with legal expertise reviewed 40 examples of the retrieval task for which no model was able to retrieve the gold reference among the top 5 results. They reviewed the input text and gold reference case (the full case that was cited in the input text), along with the top 5 retrieval results for BM25 and ColBERT, our two best-scoring models. They recorded the topic of the reference case (e.g., First Amendment law, or civil procedure) and compared it to the topic of each of the retrieved cases. 

Even where the models could not retrieve the gold reference case, they were able to retrieve at least one case of the correct topic an overwhelming majority of the time. Further, in many cases, 15/40 for BM25 and 17/40 for ColBERT, the all of the top 5 retrieved cases were the correct topic. For example, the reference case concerned a trademark issue, and the retrieved cases were also about trademark law. For two of 40 the examples, both models retrieved cases that, according to the author's judgment as an attorney, could plausibly have been cited for the proposition in the input text even though there were not same as the reference case. This points to a difficulty in evaluating this task, as frequently there may be multiple cases that plausibly support a proposition, even if they are not all cited in a brief. 

In three of the examples where neither model was able to retrieve any relevant case in the top 5, the input text was very short or otherwise lacked much context. For example, the following input text is only 167 characters long, compared to the 537 mean character input length. Given this short input and lack of context, both models failed to retrieve any case that matched the topic of the reference:

\begin{tcolorbox}[breakable]
{\small \label{prompt:case-segment}
    Pet'rs Br. 26 n.4. [[[CITATION REQUIRED]]], is therefore consistent with our argument; there was no state-imposed free, first-come, first-served rule there. Pet. App.
}
\end{tcolorbox}

In one other example where the models failed, the models retrieved cases based on information in the input that was not directly relevant to the proposition. In the following instance, the BM25 retrieved cases about sovereign immunity, and ColBERT retrieved cases about immigration. The author believes this may be due to the mention of embassies and the Vienna Convention in the input text, even though the reference text concerns bankruptcy proceedings, and is in fact cited to support an argument about statutory construction.

\begin{tcolorbox}[breakable]
{\small \label{prompt:case-segment}
    29- 32. Because the plain text of Section 1608(a)(3) resolves the question presented, "that is where the inquiry should end." Puerto [[[CITATION REQUIRED]]]. To the extent the Court wishes to consider it, however, the legislative history sheds little light on the question in this case. Petitioner and the government cite the House Judiciary Committee's report, which suggests that "[s]ervice on an embassy by mail [is] precluded under this bill" in order to ``avoid questions of inconsistency with [Article 22(1)] of the Vienna Convention.''
}
\end{tcolorbox}

\section{Hardware and Implementation Details}
We use the HuggingFace transformers, HuggingFace TRL, and PyTorch libraries to fine-tune the models. 
We format input text based on the unique special tokens of each model. 
On average, it takes around 9 hours for each model to be fine-tuned. 
We run all the fine-tuning experiments on a shared cluster of nodes using SLURM, requesting one node with an NVIDIA Ampere A100 80GB GPU and 245GB of CPU memory. 

Our fine-tuning details are as follows. We use the following models from the HuggingFace model hub: 
Mistral-7B-Instruct-v0.2, Gemma-2-2b-it, Gemma-2-9b-it, Meta-Llama-3.1-8B-Instruct,  Qwen-2.5-7B-Instruct, and Qwe-n2.5-14B-Instruct. 
These models are fine-tuned for one epoch with a batch size of 4 (except for Qwen-2.5-14B, for which we set a batch size of 2), a learning rate of 2e-5, and a cosine learning rate scheduler, checkpointing every 250 steps. 
The best checkpoint is selected based on the loss on the development set. 
We set a limit of 5K tokens for the input text and 50 tokens for the generated headings. 
The 5K limit comfortably accommodates all inputs from our summarization and completion tasks and the 50-token generation cap is twice the length of an average human-created heading.
We use the parameter-efficient Low-Rank Adaptation (LoRA) method \cite{DBLP:conf/iclr/HuSWALWWC22}. 
For LoRA, the hyperparameters are configured as follows: \texttt{rank = 8}, \texttt{lora alpha = rank * 2}, \texttt{lora dropout = 0.05}.

\input{figures/code_ex_1}
\input{figures/regex_extract}
\input{figures/mistral_extract}
\input{figures/code_ex_2}
\input{figures/code_ex_3}

\input{figures/sum_data_ex1}
\input{figures/sum_data_ex2}
\input{figures/sum_data_ex3}

\input{figures/com_data_ex1}
\input{figures/com_data_ex2}
\input{figures/com_data_ex3}
\input{figures/com_data_ex4}

\input{figures/cite_data_ex1}
\input{figures/cite_data_ex2}
\input{figures/cite_data_ex3}

\end{document}

%% file: figures/tasks_sample_fig.tex
\begin{figure*}
    \centering
    \includegraphics[width=0.99\linewidth]{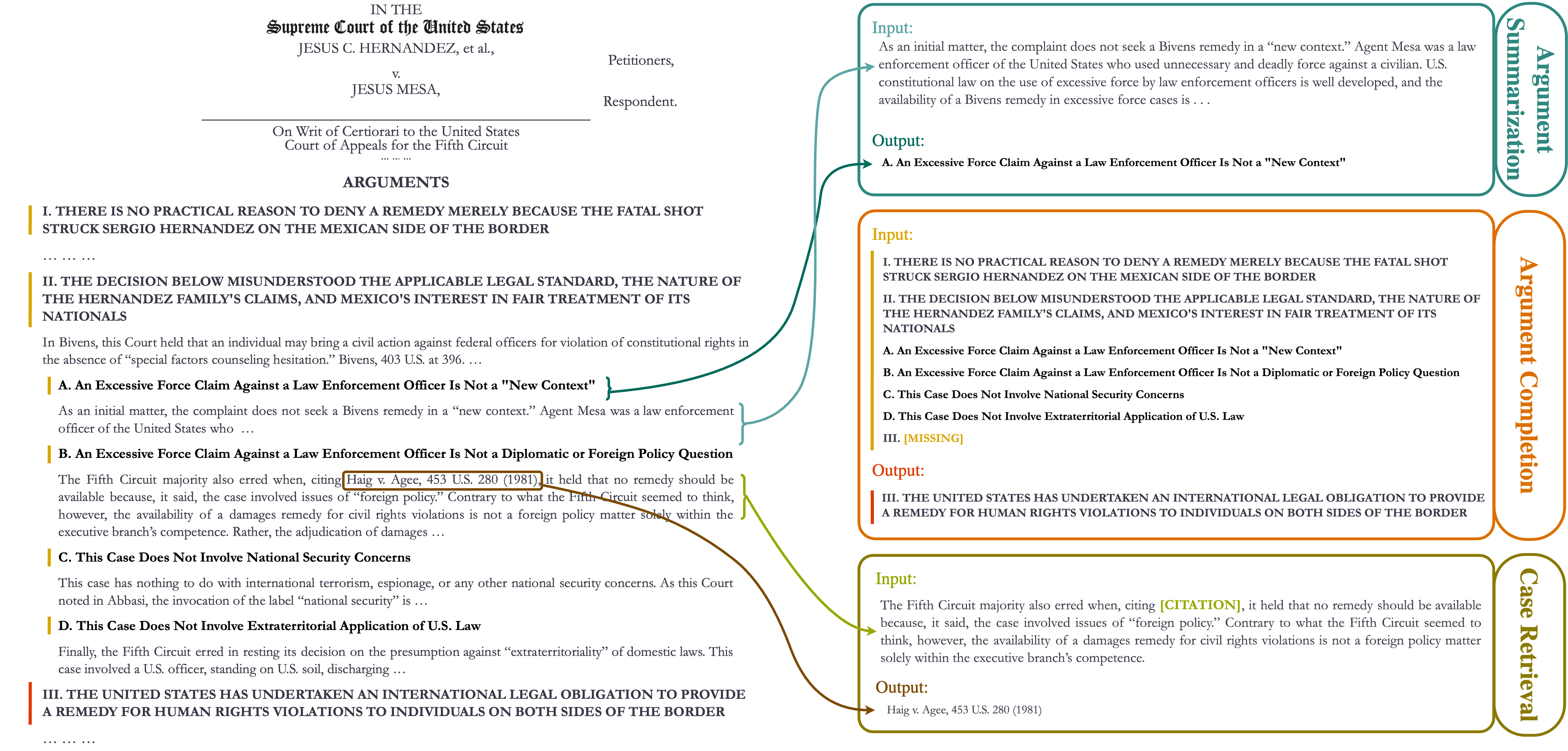}
    \vspace{-.45cm}
    \caption{We illustrate the typical structure of briefs and highlight specific parts that serve as input and desired output for the argument summarization, argument completion, and case retrieval tasks.}
    \vspace{-.45cm}
    \label{fig:tasks-overview-fig}
\end{figure*}

%% file: tables/all_summary_stats.tex
\begin{table}[t]
    \centering
    \small
    \begin{tabular}{p{2.8cm}rrr}
        \toprule
        \textbf{Statistic} & \textbf{Summ.} & \textbf{Comp.} & \textbf{Retr.} \\
        \midrule
        Train Examples & 18,642 & 3,377 & 72,868 \\
        Test Examples & 2,345 & 1,275 & 9,109 \\
        Dev. Examples & 2,345 & 1,253 & 9,109 \\
        Total Examples & 23,332 & 5,905 & 91,086 \\
        Cases in Retr. Corpus & -- & -- & 24,525 \\
        \bottomrule
    \end{tabular}
    \vspace{-.25cm}
    \caption{Train/Test/Dev splits for each task prior to quality filter. Additional statistics can be found in Appendix~\ref{sec:sum_stats}.}
    \vspace{-.55cm}
    \label{tab:all_summary_statistics}
\end{table}

%% file: tables/dataset_comparison_new.tex
\begin{table}[t]
\centering
\footnotesize
\resizebox{\columnwidth}{!}{
\begin{tabular}{p{1.8cm}p{1.2cm}p{1.2cm}|p{1.2cm}p{1.2cm}p{1.2cm}}
\toprule
\textbf{Dataset} & \textbf{Num Docs} & \textbf{Doc Type} & \textbf{Arg Summ} & \textbf{Arg Comp} & \textbf{Case Retr} \\
\midrule
BillSum & 22.2K & Statutes & 22.2K & \texttimes & \texttimes \\ 
Multi-LexSum & 9.2K & Case Files & 9.2K & \texttimes & \texttimes \\ 
CaseSumm & 25.6K & Cases & 25.6K & \texttimes & \texttimes \\ 
IN/UK-Ext/Abs & 8K & Cases & 8K & \texttimes & \texttimes \\ 
\midrule
BSARD & 22.3K & Statutory Articles & \texttimes & \texttimes & 1.1K \\ 
CLERC & 25.5M & Cases & 6K & \texttimes & 105K \\ 
Caselaw & 4M & Cases & \texttimes & \texttimes & 2.6K \\ 
LePaRD & 1.7M & Cases & \texttimes & \texttimes & 3.5M \\
\midrule
\textbf{\datasetname} & 3.7K & \textbf{Briefs} & 23.3K & 5.9K & 91K \\
\bottomrule
\end{tabular}
}
\vspace{-.25cm}
\caption{Comparison of \datasetname with other datasets with comparable tasks. Sources: BillSum~\cite{kornilova-eidelman-2019-billsum}, Multi-LexSum~\cite{DBLP:conf/nips/ShenLYDSD22}, CaseSumm~\cite{heddaya2024casesumm}, IN/UK-Ext/Abs~\cite{galgani-etal-2012-combining},
BSARD~\cite{louis2022statutory}, CLERC~\cite{hou2024clerc}, Caselaw~\cite{locke2018test}, LePaRD~\cite{mahari-etal-2024-lepard}.
We omit works with <1k total annotations. }
\vspace{-.45cm}
\label{tab:dataset_comparison}
\end{table}

%% file: tables/arg_summ_and_comp_combined.tex
\begin{table}[t]\centering
\scriptsize %
\setlength{\tabcolsep}{3pt} %
\renewcommand{\arraystretch}{0.95} %
\begin{tabular}{ll|rrrr|rr}\toprule
\textbf{Setup} & \textbf{Model} 
& \rotatebox{90}{\textbf{Perp.}} 
& \rotatebox{90}{\textbf{SC-par}} 
& \rotatebox{90}{\textbf{SC-sent}} 
& \rotatebox{90}{\textbf{o3-miniS}} 
& \rotatebox{90}{\textbf{Perp.}} 
& \rotatebox{90}{\textbf{o3-miniS}} \\
& & \multicolumn{4}{c|}{\textbf{Summarization}} & \multicolumn{2}{c}{\textbf{Completion}} \\
\midrule
\textbf{Gold} & Human & 42.6 & 38.0 & 35.5 & 4.0 & 46.5 & 3.9 \\
\midrule
\textbf{Heuristics} & Random & 31.8 & 35.2 & 90.1 & 2.1 & 47.8 & 1.4 \\
& Lead-1 & 18.4 & 35.4 & 90.7 & 2.1 & - & - \\
\midrule
\textbf{Extractive} & BERTExSumm & 32.4 & 96.8 & 92.3 & 2.4 & - & - \\
\midrule
\multirow{5}{*}{\shortstack[l]{\textbf{Abstractive}\\\emph{Zero-Shot}}}
& GPT-4o & 47.0 & 34.5 & 29.3 & 4.2 & 28.4 & \textbf{4.3} \\
& Mistral-7b & 28.3 & 42.1 & 34.5 & 4.0 & 25.0 & 3.2 \\
& Gemma-2-9b & 45.9 & 39.2 & 35.2 & 4.1 & 38.0 & 4.2 \\
& Llama-3.1-70b & \textbf{15.6} & 34.0 & 28.8 & 4.2 & 18.4 & \textbf{4.3} \\
& Qwen-2.5-32b & 29.0 & 34.9 & 30.4 & 4.1 & 28.9 & \textbf{4.3} \\
\midrule
\multirow{5}{*}{\shortstack[l]{\textbf{Abstractive}\\\emph{Few-Shot}}}
& GPT-4o & 34.9 & 34.8 & 29.9 & \textbf{4.3} & 26.7 & \textbf{4.3} \\
& Mistral-7b & 16.8 & 40.6 & 38.2 & 2.0 & 26.6 & 3.4 \\
& Gemma-2-9b & 35.7 & 42.5 & \textbf{39.2} & 3.2 & 38.0 & 3.6 \\
& Llama-3.1-70b & 37.1 & 36.2 & 32.9 & 2.6 & 42.0 & 4.1 \\
& Qwen-2.5-72b & 23.8 & 37.9 & 39.0 & 2.5 & 24.7 & \textbf{4.3} \\
\midrule
\multirow{4}{*}{\shortstack[l]{\textbf{Abstractive}\\\emph{Fine-Tuned}}}
& Mistral-7b & 21.2 & 35.5 & 31.5 & 3.8 & 20.5 & 3.0 \\
& Gemma-2-9b & 39.0 & 37.6 & 38.6 & 3.7 & 48.6 & 3.6 \\
& Llama-3.1-8b & 17.7 & 34.4 & 30.2 & 4.0 & 14.9 & 3.8 \\
& Qwen-2.5-7b/14b & 20.8 & 36.0 & 33.2 & 3.9 & 22.9 & 3.7 \\
\bottomrule
\end{tabular}
\vspace{-0.2cm}
\caption{Performance on the \textbf{argument summarization} and \textbf{argument completion} tasks. For both tasks we report perplexity (Perp.) and o3-miniS for our o3-mini LLM judge’s average rating (1-5). For summarization we also report SummaC Score (SC-par/SC-sent). We show full results in the Appendix.}
\vspace{-0.55cm}
\label{tab:compact_results}
\end{table}

%% file: tables/arg_com_stepbystep_results.tex
\begin{table}[t]\centering
\resizebox{0.47\textwidth}{!}{%
\begin{tabular}{lr|lr|lr}\toprule
\multicolumn{2}{l|}{\textbf{Binary}}  & \textbf{Heading Level} & & \textbf{Heading Location} & \\\midrule
\textbf{P} &100/100 &\textbf{Major Heading} &29/35 &\textbf{Major Heading} &6/35 \\
\textbf{R} &100/100 &\textbf{Minor Heading} &46/49 &\textbf{Minor Heading} &11/49 \\
\textbf{F1} &100/100 &\textbf{Subheading} &11/16 &\textbf{Subheading} &1/16 \\
\cmidrule(lr){3-6}
& & \textbf{Avg.} & 86/100 & \textbf{Avg.} & 18/100 \\
\bottomrule
\end{tabular}
}
\vspace{-.25cm}
\caption{Results of the realistic %
argument completion. 
\vspace{-.45cm}
}\label{tab:step_by_step_arg_comp_results}
\end{table}

%% file: tables/case_ret_results.tex
\renewcommand{\arraystretch}{0.95} %
\setlength{\tabcolsep}{3pt} %
\begin{table}[t]
\centering \small
\resizebox{\columnwidth}{!}{%
\begin{tabular}{ll|rrrrr|r|r}\toprule
\textbf{Setup} &\textbf{Model} &\textbf{R@1} &\textbf{R@5} &\textbf{10} &\textbf{50} &\textbf{00} &\textbf{MRR@10} &\textbf{nDCG@10} \\\midrule
\multirow{5}{*}{\textbf{Zero-Shot}} &BM25 &7.6 &19.6 &27.4 &46.5 &55.2 &12.9 &16.3 \\
&DPR &1.1 &4.2 &5.8 &15.1 &19.6 &2.3 &3.1 \\
&ColBERT &10.7 &21.2 &28.7 &43.2 &51.4 &15.3 &18.4 \\
&SAILER &2.2 &7.3 &11.6 &24.5 &30.3 &4.5 &6.2 \\
&CaseEncoder &2.3 &6.7 &9.8 &21.6 &27.0 &4.2 &5.6 \\
\midrule
\multirow{5}{*}{\makecell[l]{\textbf{Zero-Shot} \\ \textbf{+ Rerank}}}
&BM25 &10.0 &23.4 &33.2 &48.6 &55.2 &16.0 &20.0 \\
&DPR &4.9 &11.6 &14.3 &19.4 &19.6 &7.8 &9.3 \\
&ColBERT &11.0 &23.3 &29.8 &45.6 &51.4 &16.3 &19.5 \\
&SAILER &3.5 &8.9 &12.3 &25.2 &30.3 &6.6 &7.4 \\
&CaseEncoder &2.7 &7.6 &12.6 &22.9 &27.0 &4.4 &6.9 \\
\midrule
\multirow{2}{*}{\textbf{SFT}} &DPR &3.2 &6.4 &11.4 &23.6 &30.0 &4.9 &6.4 \\
&ColBERT &11.8 &26.4 &34.2 &51.6 &\textbf{58.5} &18.1 &21.9 \\
\midrule
\multirow{2}{*}{\textbf{SFT + Rerank}} 
&DPR &4.2 & 7.6 & 13.8 & 24.7 & 30.0 & 6.1 & 7.9 \\
&\textbf{ColBERT} & \textbf{13.7} & \textbf{31.4} & \textbf{36.3} & \textbf{52.6} & \textbf{58.5} & \textbf{18.6} &\textbf{24.7} \\
\bottomrule
\end{tabular}
}
\vspace{-.25cm}
\caption{ Retrieval results on the test set our data evaluated with Recall@1, 5, 10, 50, and 100, MRR@10, and
nDCG@10. MRR is Mean Reciprocal Rank, and nDCG normalized Discounted Cumulative Gain.}
\vspace{-.45cm}
\label{tab:case_ret_results}
\end{table}

%% file: tables/dataset_stats.tex
\begin{table}[t]
    \centering
    \begin{tabular}{lr}
        \toprule
        \textbf{Statistic} & \textbf{Value} \\
        \midrule
        Date Range & 2017--2024 \\
        Num Cases & 360 \\
        Num Briefs & 3753 \\
        \midrule
        Mean Num Words per Brief & 8956 \\
        Mean Num Sentences per Section & 52 \\
        Mean Num Words per Section & 1024 \\
        \midrule
        Mean Num Sections per Brief & 7 \\
        Min Sections per Brief & 1 \\
        Max Sections per Brief & 46 \\
        \midrule
        Num Unique Bigrams & 3,508,269 \\
        Num Unique Trigrams & 10,684,655 \\
        Num Unique Non Stop-Words & 147,647 \\
        \midrule
        Mean Flesch-Kincaid & 9 \\
        Mean Coleman-Liau & 9 \\
        Mean SMOG & 12 \\
        Mean ARI & 15 \\
        \bottomrule
    \end{tabular}
    \vspace{-.25cm}
    \caption{Corpus Statistics. Flesch-Kincaid, Coleman-Liau, SMOG, and ARI are automated measures of linguistic complexity \cite{ManorLi2019}. Each corresponds to the number of years of education required to understand the text. We used NLTK for stopword removal.}
    \label{tab:doc_statistics}
    \vspace{-.55cm}
\end{table}

%% file: tables/summary_task_stats.tex
\begin{table}[t]
    \centering
    \begin{tabular}{lr}
        \toprule
        \textbf{Statistic} & \textbf{Value} \\
        \midrule
        Train Summarization Examples & 18642\\
        Test Summarization Examples& 2345\\
        Dev Summarization Examples& 2345\\
        Total Summarization Examples& 23332\\
        \midrule
        Mean Words per Section & 840\\
        Min Words per Section & 25\\
        Max Words per Section & 35,713\\
        \midrule
        Mean Words per Header & 15\\
        Min Words per Header & 3\\
        Max Words per Header & 105\\
        \midrule
        Mean Coverage & 0.80\\
        Mean Density & 2.60\\
        Mean Compression & 59.90\\
        Median Formulaicness & 10.36\\
        \bottomrule
    \end{tabular}
    \caption{Argument Summarization Data Statistics. Mean Compression is the ratio of the number of words in the header to that in the section text. Coverage measures the percentage of words in the header that are part of an extractive fragment in the section. Density is the average length of the extractive fragment from the section to which each word in the header belongs. \cite{grusky2018newsroom} introduced Coverage and Density for summarization tasks. \cite{ragazzi2024lawsuit} introduced formulaicness, a measure of recurring structural patterns. 
}
    \label{tab:summary_statistics}
\end{table}

%% file: tables/completion_stats.tex
\begin{table}[t]
    \centering
    \begin{tabular}{lr}
        \toprule
        \textbf{Statistic} & \textbf{Value} \\
        \midrule
        Train Completion Examples& 3377\\
        Test Completion Examples& 1275\\
        Dev Completion Examples& 1253\\
        Total Completion Examples& 5905\\
        \midrule
        Mean Masked Header Length&15 \\
        Num Masked Top Level Headers&2751 \\
        Num Masked Mid Level Headers&228 \\
        Num Masked Leaf Level Headers&2926 \\
        \bottomrule
    \end{tabular}
    \caption{Argument Completion Data Statistics. Each section represents an argument that may be masked for the argument completion task. In the training set, only a single argument is masked per brief. In the test and dev set, a brief may appear multiple times with different arguments masked, so the total is greater than the number of briefs in the corpus.}
    \label{tab:completio_summary}
\end{table}

%% file: tables/retrieval_stats.tex
\begin{table}[t]
    \centering
    \begin{tabular}{lr}
        \toprule
        \textbf{Statistic} & \textbf{Value} \\
        \midrule
        Train Citation Examples & 72868\\
        Test Citation Examples& 9109\\
        Dev Citation Examples& 9109\\
        Total Citation Examples& 91086\\
        Num Cases in Retrieval Corpus & 24525\\
        \midrule
        Mean Citations per Brief & 25 \\
        Mean Citations per Section & 4 \\
        \midrule
        Num of Unique Citations & 33,034 \\
        Num of Direct Quote Citations & 16,697 \\
        Mean Direct Quotes per Section & 0.65 \\
        \bottomrule
    \end{tabular}
    \caption{Case Retrieval Data Statistics. The retrieval corpus is the collection of full-text judicial opinions. Multiple citations can point to a single case, so the number of opinions is less than the number of citation examples. Direct quotes, which could make retrieval trivial, do not dominate the corpus.}
    \label{tab:citation_summary}
\end{table}

%% file: figures/citation_distr.tex
\begin{figure}
    \centering
    \includegraphics[width=0.99\linewidth]{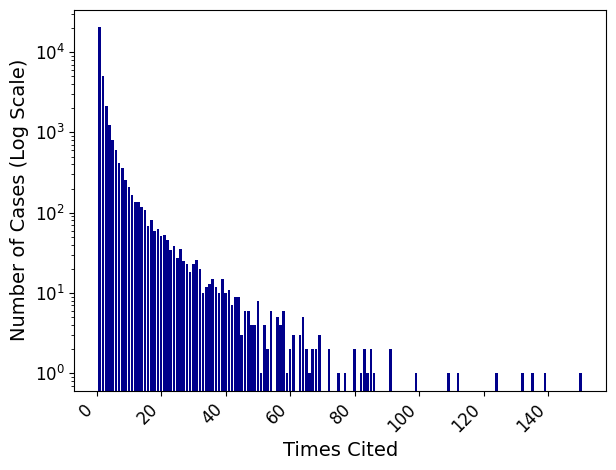}
    \vspace{-.85cm}
    \caption{Distribution of case citations.
    Shows a long-tail of citations where most cases are cited very infrequently and a few are cited many times. }
    \vspace{-.5cm}

    \label{fig:citation_distribution}
\end{figure}

%% file: figures/pipeline.tex
\begin{figure*}[t]
    \centering
    \includegraphics[width=1.00\linewidth]{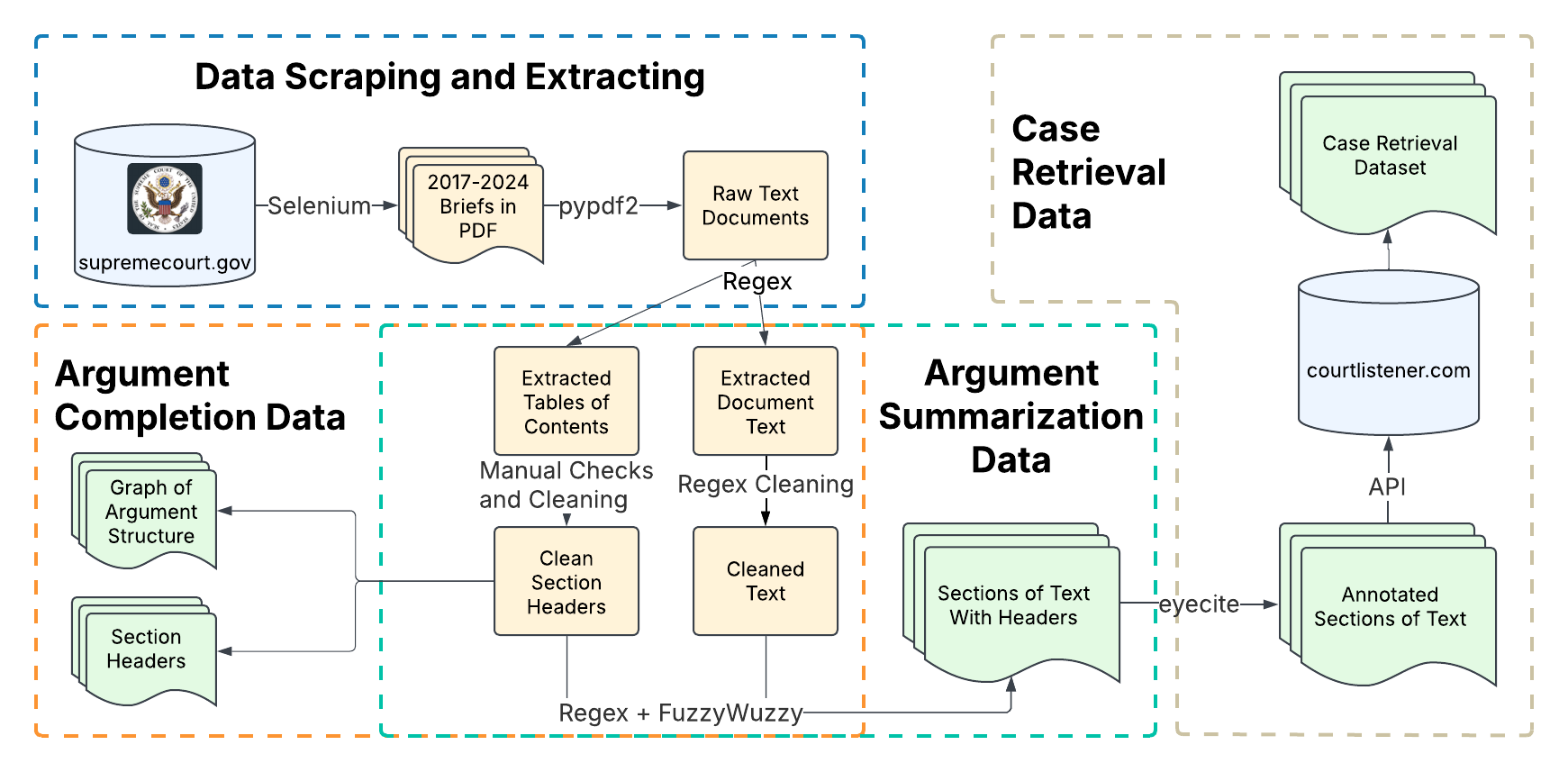}
    \vspace{-.55cm}
    \caption{An illustration of the extraction and cleaning pipeline for \datasetname. 
    }
    \label{fig:pipeline_fig}
    \vspace{-.15cm}
\end{figure*}

%% file: figures/scotus_page.tex
\begin{figure*}[!h]
    \centering
    \includegraphics[width=0.99\linewidth]{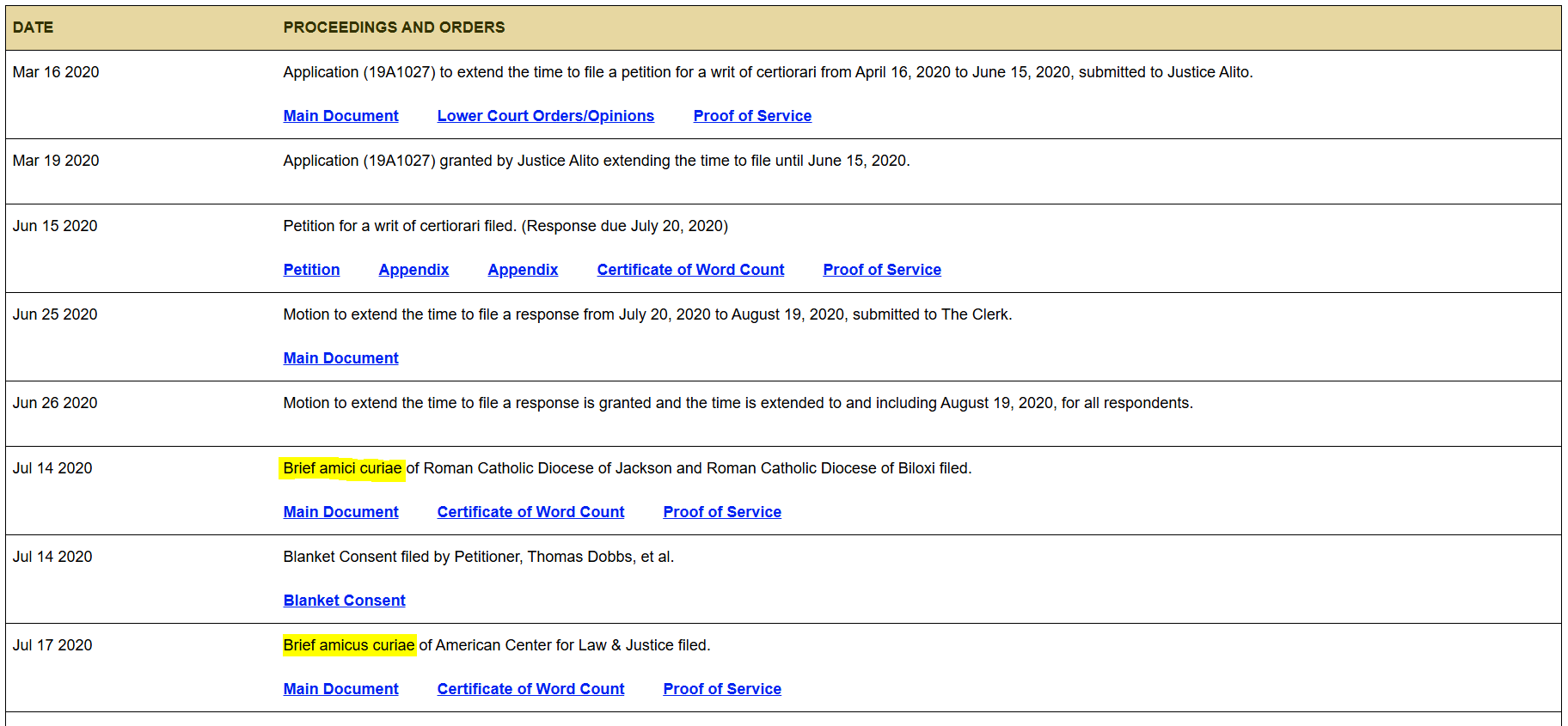}
    \caption{An example of the table that organizes case documents in \url{www.supremecourt.gov}.}
    \label{fig:scotus_page}
\end{figure*}

%% file: figures/retrieval.tex
\begin{figure*}
    \centering
    \includegraphics[width=0.9\linewidth]{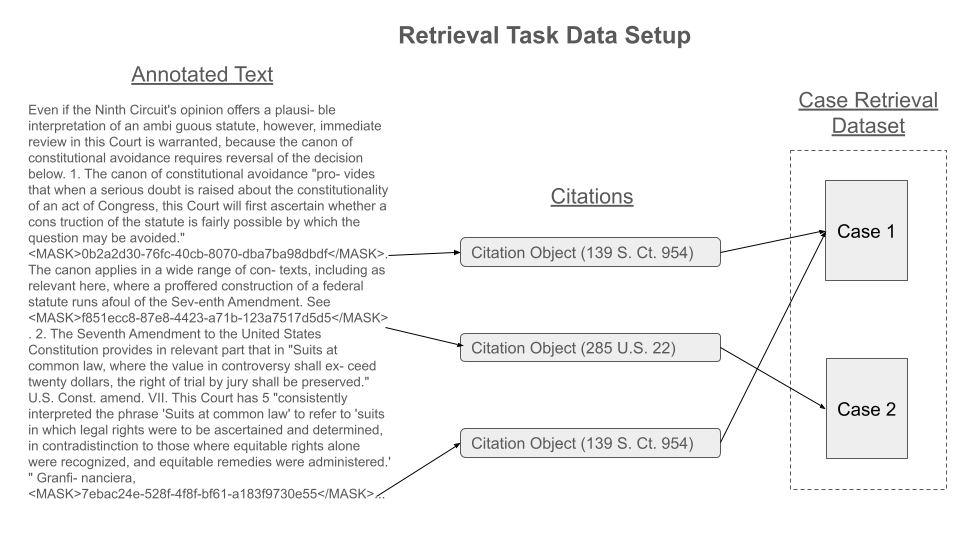}
    \caption{This figure illustrates the structure and flow of the case retrieval data. A case that is cited multiple times will have a unique mask token ID, but the different citation objects will all point to the same case.}
    \label{fig:retrieval_fig}
\end{figure*}

%% file: figures/bertopic.tex
\begin{figure*}[t]
    \centering
    \includegraphics[width=0.99\linewidth]{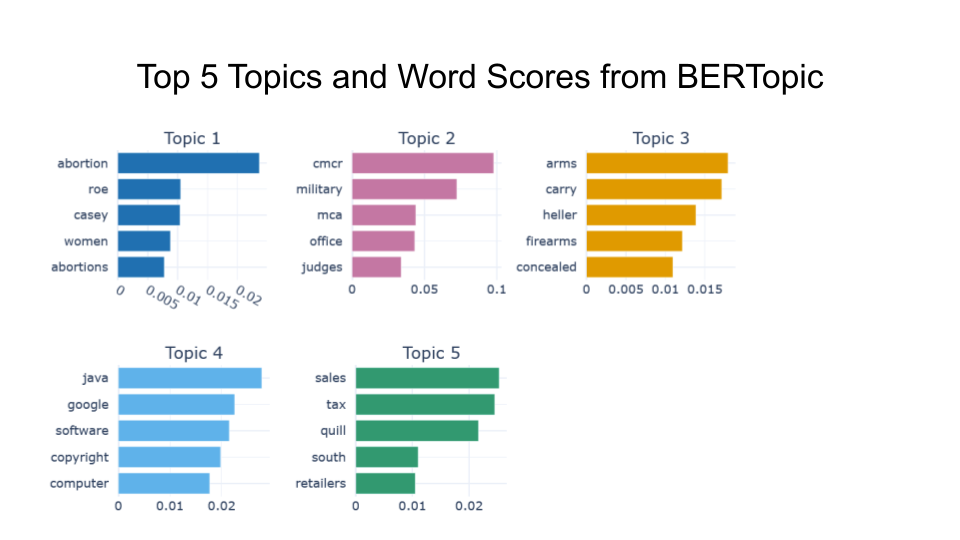}
    \caption{This figure shows the top 5 topic groupings and their word scores calculated with BERTopic. \cite{grootendorst2022bertopic}.}
    \label{fig:bertopic_5}
\end{figure*}

%% file: tables/top2vec.tex
\begin{table*}[ht]
    \centering
        \begin{tabular}{c c c c c c c c c c c}
        \specialrule{1.2pt}{0pt}{0pt}
        \textbf{Topic Rank} & \textbf{Word 1} & \textbf{Word 2} & \textbf{Word 3} & \textbf{Word 4} & \textbf{Word 5} \\
        \midrule
        1 & merits & case & but & bielski & exhaustion \\
        2 & oracle & copying & copyrighted & copied & java \\
        3 & gundy & nondelegation & delegation & wayman & delegated \\
        4 & orden & bladensburg & commandments & allegheny & memorial \\
        5 & fetus & viability & fetal & womb & unborn \\
        \specialrule{1.2pt}{0pt}{0pt}
    \end{tabular}
    
    \caption{Topic Modeling Results from Top2Vec  \cite{angelov2020top2vec}.}
    \label{tab:top2vec}
\end{table*}

%% file: figures/step-by-step-arg-comp-ex.tex
\begin{figure*}[t]
    \centering
    \includegraphics[width=0.99\linewidth]{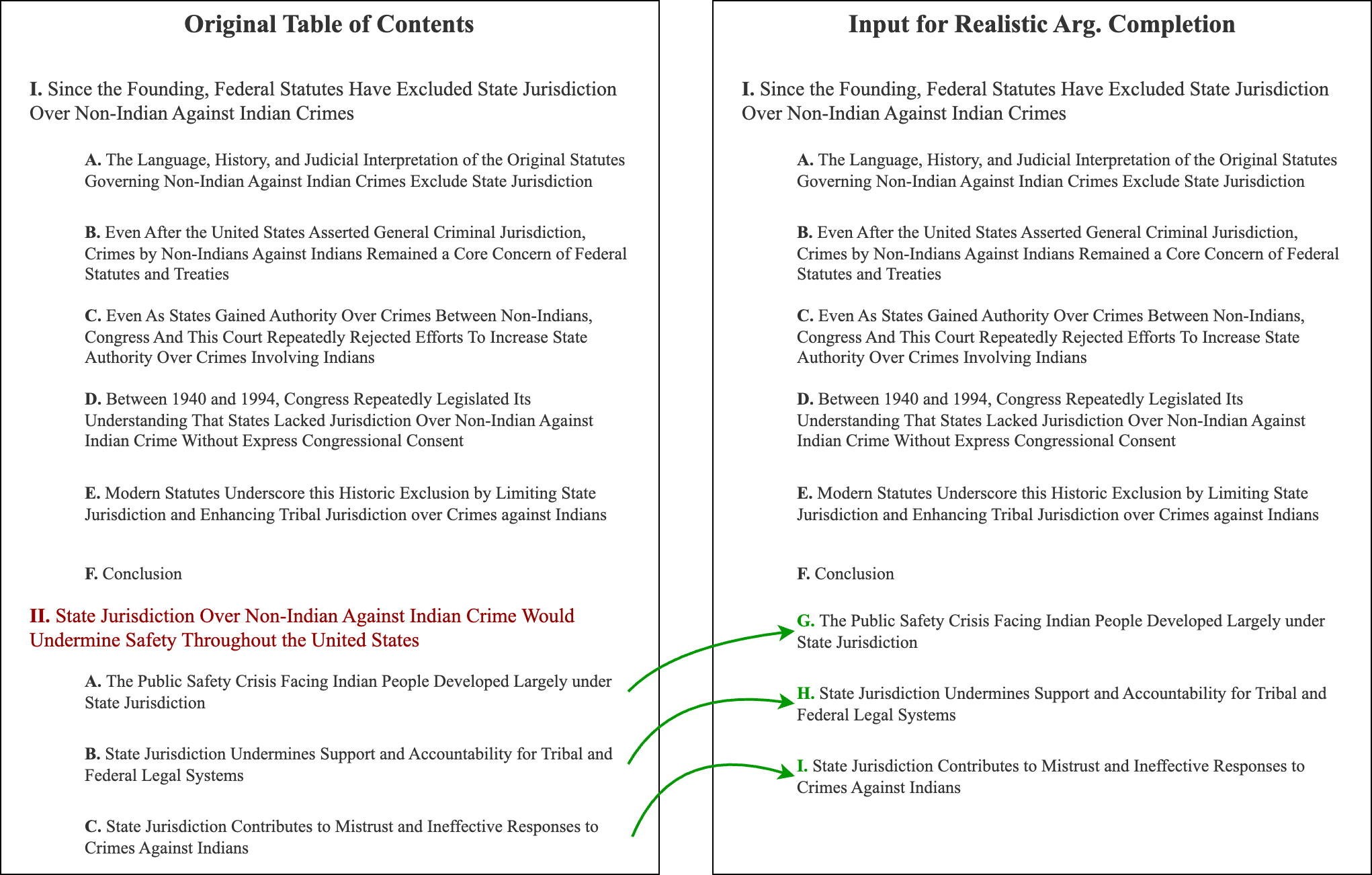}
    \caption{Step-by-step Argument Completion Task Data Example}
    \label{fig:step-by-step-argcomp-ex}
\end{figure*}

%% file: tables/arg_summ_fewshot.tex
\begin{table*}[ht]
    \centering
    \small
    \begin{tabular}{p{0.95\textwidth}}  %
        \toprule
        \textcolor{red}{\textbf{Example 1:}} \\
        \textcolor{blue}{Section Text:} \\
        The court of appeals affirmed two restitution awards (one of them for nearly \$1.27 billion) even though the purported authority for that remedy, \S13(b) of the FTC Act, says merely that a court may issue a ``temporary restraining order,'' a ``preliminary injunction,'' or a ``permanent injunction.'' 15 U.S.C. \S 53(b). \\
        ``Injunction'' does not mean ``restitution.'' ``Apples,'' after all, does not mean ``oranges.'' Nor does ``injunction'' mean ``equitable relief (including, at times, restitution).'' That would be like saying that ``apples'' means ``fruit (including, at times, oranges).'' Nor, finally, can it be said that some aspect of the FTC Act's structure reveals Congress's subtle intent to use ``injunction'' to mean ``injunction, but maybe restitution too.'' Section 13(b) is plainly designed to be ``a simple stop-gap measure,'' 910 F.3d at 431 (O'Scannlain, J., specially concurring), one that enables the FTC to enjoin a practice while it uses other statutory authority to prosecute an offender. \\
        The panel was bound by Ninth-Circuit precedent to conclude that ``injunction,'' as used in \S13(b), can mean ``restitution.'' Like most other circuits, the Ninth Circuit has decided that Porter, 328 U.S. 395, requires this twisted interpretation. Porter concludes that Congress's use of ``injunction'' in a different statute ``invoked the court's . . . inherent equitable powers.'' FTC v. Commerce Planet, Inc., 815 F.3d 593, 598 (9th Cir. 2016). Although Porter means by this that the word ``injunction'' triggers the equity jurisdiction that originated in the Court of Chancery, there are distinct shades, in Porter and other mid-twentieth century Supreme Court cases, of another kind of ``equity.'' These cases engage in a form of judicial lawmaking that harkens back to the ancient—and defunct—concept of the equity of the statute. \\
        \textcolor{blue}{Heading:} \\
        The Lower Courts' Expansion Of section 13(b) Is, In Effect, An Exercise Of The Equity Of The Statute \\
        \midrule
        \textcolor{red}{\textbf{Example 2:}} \\
        \textcolor{blue}{Section Text:} \\
        In 1991, the European Union adopted a Software Directive, which reflects a policy judgment that copyright should not prevent competition in the software industry.3 Council of Ministers Directive 91/250/EEC of 14 May 1991 on the Legal Protection of Computer Programs, 1991 O.J. (L 122). In particular, Article 6 of the Software Directive permits reverse engineering ``indispensable to obtain the information necessary to achieve . . . interoperability.''4 Further, Article 9(1) renders unenforceable contractual prohibitions on such reverse engineering. The Software Directive has been implemented by all EU member states, as well as Norway, Russia, Serbia, Switzerland, and Turkey. Global API Conflict at 619. \\
        The Software Directive did not directly address the protectability of software interfaces. However, in 2012, the EU’s highest court ruled in SAS Institute v. World Programming, (C-406/10) [2012] 3 CMLR 4 (Eng.), \S 40, that the Software Directive ``must be interpreted as meaning that neither the functionality of a computer program nor the programming language and the format of data files used in a computer program in order to exploit its functions constitute a form of expression of that program and, as such, are not protected by copyright. . . .'' This affirmed World Programming’s ability to create ``middleware'' that interoperated with SAS Institute’s software. The Court of Justice of the European Union (``CJEU'') observed that ``the main advantage of protecting computer programs by copyright'' as opposed, presumably, to patents, ``is that such protection covers only the individual expression of the work and thus leaves other authors the desired latitude to create similar or even identical programs,'' id. at \S 41, provided that they refrain from copying protected expression. In other words, the CJEU reached precisely the same conclusion as the district court below, and the opposite of the Federal Circuit’s 2014 decision. \\
        \textcolor{blue}{Heading:} \\
        European Union Law Encourages Competition in the Software Industry \\
        \midrule
        \textcolor{red}{\textbf{Example 3:}} \\
        \textcolor{blue}{Section Text:} \\
        The Computer \& Communications Industry Association (``CCIA'') represents more than 20 companies of all sizes providing high technology products and services, including computer hardware and software, electronic commerce, telecommunications, and Internet products and services—companies that collectively generate more than \$540 billion in annual revenues.2 CCIA members have a large stake in the rules of software copyright: effective intellectual property protection encourages developers to create new applications, but the improper extension of copyright law to functional elements discourages innovation and inhibits competition in the industry. \\
        Over the past 30 years, and largely as a result of American jurisprudence and leadership, a global consensus has emerged on the appropriate scope of copyright protection for software. Legislatures and courts around the world have exercised great care to prevent overly restrictive rules that would impede the creation of new computer programs that can run on existing operating systems, or the creation of new operating systems that can be used by programmers with their existing skill-set. The two decisions in this case of the U.S. Court of Appeals for the Federal Circuit run directly contrary to this global consensus, and thus threaten uniquely to disadvantage American innovation. For this reason, Google’s Petition should be granted. \\
        \textcolor{blue}{Heading:} \\
        INTEREST OF AMICUS CURIAE \\
        \bottomrule
    \end{tabular}
    \caption{Three examples used for few-shot prompting in the argument summarization task.}
    \label{tab:arg_summ_fewshot_ex}
\end{table*}

%% file: tables/arg_comp_fewshot.tex
\begin{table*}[ht]
    \centering
    \small
    \begin{tabular}{p{0.95\textwidth}}  %
        \toprule
        \textcolor{red}{\textbf{Example 1:}} \\
        \textcolor{blue}{Heading List:} \\
        I. EXTENDING ATS LIABILITY TO U.S.-BASED COMPANIES BECAUSE OF THIRD-PARTY OVERSEAS CONDUCT THREATENS THE VITAL ROLE CORPORATIONS PLAY ABROAD \\
        A. The Coca-Cola Company's Efforts Demonstrate The Beneficial Impact U.S.-Based Corporations Can Have Abroad \\
        B. Imposing ATS Liability For Corporate Oversight Would Deter Proactive Efforts \\
        II. NEITHER INTERNATIONAL LAW PRINCIPLES, NOR DOMESTIC SEPARATION-OF-POWER PRINCIPLES, PERMIT COURTS TO RECOGNIZE CORPORATE LIABILITY UNDER THE ATS \\
        A. The Lack Of Consensus For Extending International Law Status To Corporations Stems In Part From Concerns That Doing So Will Compromise The Sovereignty Of Nations \\
        1. ``Subjects'' of International Law Typically Possess Powers As Well as Obligations \\
        2. Recognizing Corporations As ``Subjects'' Of International Law Is Perceived To Compromise State Sovereignty \\
        \textbf{3. [MISSING]} \\
        B. The Lack Of International Law Consensus Regarding Corporate Liability Forecloses Such Liability Under The ATS \\
        C. To The Extent That Corporate Liability Poses A Domestic Law Question, It Is One That Congress Must Answer \\
        III. DOMESTIC CORPORATE OVERSIGHT OF OVERSEAS ACTIVITIES CANNOT OVERCOME THE EXTRATERRITORIALITY BAR \\
        \textcolor{blue}{Missing Heading:} \\
        Acceptance Of Corporations As International Law Subjects Does Not Follow From The Fact That Some International Law Norms Have Been Deemed To Bind Individuals \\
        \midrule
        \textcolor{red}{\textbf{Example 2:}} \\
        \textcolor{blue}{Heading List:} \\
        \textbf{I. [MISSING]} \\
        A. The government's textual arguments are meritless. \\
        1. Text of section 1182(a)(2). \\
        2. Surrounding provisions in section 1182(a). \\
        3. Structure of the INA. \\
        B. The two-part structure of the stop-time rule demonstrates that Petitioner is correct. \\
        C. The purpose and history of the stop-time rule reinforce that Petitioner's position is correct. \\
        II. Alternatively, If An Alien Is Capable Of Being Charged With Inadmissibility, Then The Offense ``Renders The Alien Inadmissible.'' \\
        \textcolor{blue}{Missing Heading:} \\
        An Offense ``Renders The Alien Inadmissible'' If The Immigration Judge Finds That It Renders The Alien Inadmissible. \\
        \midrule
        \textcolor{red}{\textbf{Example 3:}} \\
        \textcolor{blue}{Heading List:} \\
        I. AIDING-AND-ABETTING LIABILITY UNDER section 2333(d)(2) IS NOT LIMITED TO INSTANCES IN WHICH A PARTICULAR ACT OF ASSISTANCE IS CLOSELY CONNECTED TO THE PARTICULAR ACT OF INTERNATIONAL TERRORISM THAT INJURED THE PLAINTIFF \\
        A. Introduction \\
        B. The Halberstam Standard Encompasses Aiding and Abetting By Assisting A Wrongful Enterprise \\
        C. The Text of section 2333(d)(2) Applies To Assistance To A Terrorist Enterprise \\
        D. The Statutory Context Supports Interpreting section 2333(d)(2) To Apply To Assisting A Terrorist Enterprise \\
        E. The Other Arguments of Defendants and The United States Are Not Persuasive \\
        F. Neither The Defendants Nor The United States Propose A Plausible Standard for Determining Which Types of Assistance To A Terrorist Enterprise Would And Would Not Be Covered by section 2333(d)(2) \\
        1. The Proposed Twitter/Facebook Standard \\
        2. The Proposed Government Standard \\
        II. THE COURT SHOULD NOT ADOPT ANY OF THE NEW LEGAL RULES AND REQUIREMENTS PROPOSED BY DEFENDANTS OR THE UNITED STATES \\
        1. Knowledge of Accounts or Postings Connected To The Reina Attack \\
        2. Knowledge of Accounts or Postings Used for Particular Attacks \\
        3. Knowledge of Substantial Assistance \\
        4. Intent Requirement \\
        5. Special Standard for Remoteness \\
        \textbf{6. [MISSING]} \\
        7. Special Standard for Widely Available, Ordinary Services \\
        8. Requirement of Direct Knowledge \\
        III. THE COMPLAINT PLAUSIBLY ALLEGED THE DEFENDANTS KNOWINGLY ASSISTED ISIS'S TERRORIST ACTIVITIES \\
        A. The Allegations of The Complaint \\
        B. Defendants' Asserted Lack of More Specific Knowledge \\
        \textcolor{blue}{Missing Heading:} \\
        Special Standard for Routine Services \\
        \bottomrule
    \end{tabular}
    \caption{Three examples used for few-shot prompting in the argument completion task.}
    \label{tab:arg_comp_fewshot_ex}
\end{table*}

%% file: tables/arg_summ_results.tex
\begin{table*}[t]\centering
\resizebox{\textwidth}{!}{%
\begin{tabular}{llrrrr|rr|rrr|rr|r}\toprule
\textbf{Setup} & \textbf{Model} & \textbf{R1} & \textbf{RL} & \textbf{Bleu} & \textbf{Meteor} & \textbf{BS} & \textbf{LegalBS} & \textbf{Perp.} & \makecell{\textbf{Avg. Tkn}\\\textbf{Common}} & \makecell{\textbf{Avg. Norm}\\\textbf{Levenshtein}} & \textbf{SC-par} & \textbf{SC-sent} & \textbf{o3-miniS} \\\midrule
\textbf{Gold} & Human & - & - & - & - & - & - & 42.6 & - & - & 38.0 & 35.5 & 4.0 \\
\midrule
\multirow{3}{*}{\textbf{Baseline}} & Random & 17.7 & 14.0 & 1.4 & 15.9 & 46.5 & 64.4 & 31.8 & 35.7 & 78.1 & 35.2 & 90.1 & 2.1 \\
 & Lead-1 & 23.7 & 18.3 & 2.3 & 28.1 & 52.0 & 68.1 & 18.4 & 70.6 & 79.5 & 35.4 & 90.7 & 2.1 \\
 & BERTExSumm & 27.6 & 22.2 & 3.6 & 26.3 & 53.3 & 68.9 & 32.4 & 44.2 & 74.5 & 96.8 & 92.3 & 2.4 \\
\midrule
\multirow{9}{*}{\textbf{Zero-shot}} & GPT-4o & 34.9 & 27.4 & 3.7 & 33.3 & 59.5 & 71.9 & 47.0 & 37.4 & 73.8 & 34.5 & 29.3 & 4.2 \\
 & Mistral-7b & 30.3 & 23.0 & 3.0 & 27.6 & 54.6 & 69.7 & 28.3 & 47.4 & 75.0 & 42.1 & 34.5 & 4.0 \\
 & Gemma-2-2b & 34.3 & 28.0 & 3.9 & 25.8 & 54.1 & 69.4 & 51.2 & 31.5 & 72.7 & 41.6 & 35.6 & 3.9\\
 & Gemma-2-9b & 34.1 & 28.0 & 5.0 & 27.7 & 58.2 & 70.8 & 45.9 & 31.4 & 72.3 & 39.2 & 35.2 & 4.1 \\
 & Llama-3.1-8b & 32.7 & 25.3 & 3.8 & 32.4 & 57.4 & 71.3 & 19.1 & 49.6 & 74.6 & 34.8 & 30.4 & 4.1 \\
 & Llama-3.1-70b & 32.6 & 24.8 & 4.1 & \textbf{35.6} & 57.9 & 71.8 & \textbf{15.6} & 51.4 & 75.3 & 34.0 & 28.8 & 4.2 \\
 & Qwen-2.5-7b & 33.5 & 25.8 & 4.0 & 34.2 & 58.4 & 71.8 & 24.4 & 48.3 & 74.5 & 34.5 & 29.0 & 4.1 \\
 & Qwen-2.5-14b & 34.1 & 26.5 & 4.3 & 33.1 & 58.5 & 72.0 & 24.6 & 42.4 & 73.5 & 34.0 & 30.1 & 4.1 \\
 & Qwen-2.5-32b & 33.9 & 26.5 & 4.0 & 33.4 & 58.5 & 71.7 & 29.0 & 44.9 & 73.8 & 34.9 & 30.4 & 4.1 \\
 & Qwen-2.5-72b & 34.3 & 26.9 & 3.3 & 32.5 & 58.6 & 71.5 & 41.2 & 38.2 & 73.7 & 34.7 & 29.8 & 3.9 \\
\midrule
\multirow{9}{*}{\textbf{Few-shot}} & GPT-4o & 36.2 & 28.5 & 5.3 & 33.9 & \textbf{60.0} & \textbf{72.6} & 34.9 & 35.5 & 72.4 & 34.8 & 29.9 & \textbf{4.3} \\
 & Mistral-7b & 21.6 & 16.7 & 1.6 & 20.2 & 48.2 & 65.0 & 16.8 & 49.9 & 77.6 & 40.6 & 38.2 & 2.0 \\
 & Gemma-2-2b & 23.2 & 18.0 & 1.8 & 21.2 & 48.6 & 66.1 & 27.8 & 49.6 & 77.6 & \textbf{46.6} & 38.2 & 2.4 \\
 & Gemma-2-9b & 27.4 & 21.5 & 2.4 & 24.8 & 53.0 & 67.7 & 35.7 & 45.9 & 75.5 & 42.5 & \textbf{39.2} & 3.2 \\
 & Llama-3.1-8b & 23.1 & 18.1 & 2.2 & 20.4 & 49.6 & 66.4 & 37.8 & 40.7 & 77.7 & 37.6 & 35.9 & 2.6 \\
 & Llama-3.1-70b & 21.9 & 17.2 & 2.8 & 19.8 & 48.1 & 65.8 & 37.1 & 37.1 & 78.6 & 36.2 & 32.9 & 2.6 \\
 & Qwen-2.5-7b & 23.1 & 17.7 & 1.8 & 21.5 & 50.1 & 66.0 & 25.5 & 50.2 & 77.5 & 38.3 & 36.4 & 2.4 \\
 & Qwen-2.5-14b & 23.1 & 17.9 & 1.9 & 21.9 & 49.9 & 65.9 & 25.0 & 50.2 & 77.2 & 38.0 & 38.1 & 2.4 \\
 & Qwen-2.5-32b & 22.7 & 17.8 & 1.7 & 20.8 & 49.7 & 65.8 & 25.1 & 50.1 & 77.4 & 38.6 & 37.9 & 2.4 \\
 & Qwen-2.5-72b & 24.3 & 18.8 & 2.3 & 22.4 & 50.5 & 66.4 & 23.8 & 50.3 & 76.6 & 37.9 & 39.0 & 2.5 \\
\midrule
\multirow{5}{*}{\textbf{Fine-Tuning}} & Mistral-7b & \textbf{36.3} & 29.5 & 5.0 & 33.1 & 59.1 & 71.8 & 21.2 & 41.2 & 72.2 & 35.5 & 31.5 & 3.8 \\
 & Gemma-2-9b & 36.2 & \textbf{30.5} & \textbf{6.6} & 31.0 & 58.8 & 71.2 & 39.0 & 31.2 & 70.5 & 37.6 & 38.6 & 3.7\\
 & Llama-3.1-8b & 33.2 & 26.2 & 3.8 & 31.8 & 57.7 & 71.3 & 17.7 & 46.1 & 73.1 & 34.4 & 30.2 & 4.0 \\
 & Qwen-2.5-7b & 33.7 & 27.4 & 4.4 & 32.8 & 58.1 & 71.1 & 20.8 & 45.3 & 73.1 & 36.0 & 33.2 & 3.9 \\
 & Qwen-2.5-14b & 32.9 & 26.5 & 3.0 & 31.4 & 57.5 & 70.9 & 22.9 & 45.5 & 73.4 & 37.7 & 35.2 & 3.8 \\

\bottomrule
\end{tabular}
}
\caption{Performance of different models for the \textbf{argument summarization} task on the test split of \textsc{BriefMe} under automatic metrics and transformer-based embedding similarity. R1 stands for Rouge-1, 
BS for BERTScore, LegalBS for LegalBERTScore, Perp. for median perplexity, SC-par for median paragraph-level SummaC score, SC-sent for median sentence-level SummaC score, and o3-miniS for \texttt{o3-mini}'s average rating. The scale for \texttt{o3-mini}'s scores is: Ineffective (1), Weak (2), Adequate (3), Strong (4), and Exemplary (5). } \label{tab:arg_summ_results}
\end{table*}

%% file: tables/arg_comp_results.tex
\begin{table*}[t]\centering
\resizebox{\textwidth}{!}{%
\begin{tabular}{llrrrr|rr|rrr|r}\toprule
\textbf{Setup} &\textbf{Model} &\textbf{R1} &\textbf{RL} &\textbf{Bleu} &\textbf{Meteor} &\textbf{BS} &\textbf{LegalBS} &\textbf{Perp.} &\makecell{\textbf{Avg. Tkn}\\\textbf{Common}} &\makecell{\textbf{Avg. Norm}\\\textbf{Levenshtein}} &\textbf{o3-miniS} \\\midrule
\textbf{Gold} &Human & -& -& -& -& -& -& 46.5& -& -& 3.9 \\
\midrule
\textbf{Baseline} &Random &21.2& 18.0& 6.3& 18.6& 54.4& 67.0& 47.8& 23.4& 74.2& 1.4 \\
\midrule
\multirow{9}{*}{\textbf{Zero-shot}} &GPT-4o &29.3& 25.4& 8.3& 26.2& 57.5& 69.0& 28.4& 31.2& 70.9& \textbf{4.3}\\
&Mistral-7b &22.2& 17.3& 2.7& 21.5& 53.6& 67.7& 25.0& 47.5& 77.8& 3.2\\
&Gemma-2-2b &20.4& 16.7& 1.9& 19.0& 49.2& 66.1& 31.7& 50.9& 79.7& 2.8\\
&Gemma-2-9b &23.0& 18.2& 3.2& 23.2& 49.3& 66.7& 38.0& 57.3& 77.5& 4.2\\
&Llama-3.1-8b &18.9& 14.8& 2.2& 22.9& 49.3& 66.5& 23.3& 59.5& 80.5& 2.2 \\
&Llama-3.1-70b &23.9& 19.8& 4.3& 25.1& 52.9& 68.0& 18.4& 52.9& 77.4& \textbf{4.3}\\
&Qwen-2.5-7b &26.6& 22.1& 5.6& 27.4& 56.8& 69.2& 19.6& 49.8& 75.5& 3.5 \\
&Qwen-2.5-14b &25.5& 21.8& 5.2& 26.7& 55.1& 68.7& 23.9& 49.9& 74.9& \textbf{4.3}\\
&Qwen-2.5-32b &28.0& 23.9& 6.1& 25.7& 56.6& 68.8& 28.9& 38.3& 72.5& \textbf{4.3}\\
&Qwen-2.5-72b &31.3& 27.2& 5.8& 26.1& 58.6& 69.7& 39.1& 29.8& 72.5& 4.0 \\
\midrule
\multirow{9}{*}{\textbf{Few-shot}} &GPT-4o &25.5& 21.6& 4.7& 18.5& 53.1& 66.9& 26.7& 28.7& 73.6& \textbf{4.3}\\
&Mistral-7b &22.5& 18.2& 3.6& 20.8& 53.3& 67.5& 26.6& 40.0& 76.1& 3.4\\
&Gemma-2-2b &12.3& 10.3& 1.1& 9.6& 43.4& 62.0& 52.3& 34.5& 82.5& 2.0\\
&Gemma-2-9b &24.2& 20.5& 3.9& 19.2& 50.5& 66.4& 38.0& 33.7& 76.0& 3.6\\
&Llama-3.1-8b &24.5& 19.9& 4.0& 20.4& 51.6& 66.7& 39.3& 37.5& 74.8& 3.9\\
&Llama-3.1-70b &27.9& 23.3& 6.7& 23.0& 53.8& 67.8& 42.0& 31.4& 73.5& 4.1\\
&Qwen-2.5-7b &24.7& 20.7& 4.1& 19.3& 52.5& 66.8& 63.9& 33.4& 75.2& 4.0\\
&Qwen-2.5-14b &23.2& 19.4& 3.8& 19.2& 51.8& 67.1& 43.2& 39.4& 75.6& 4.1\\
&Qwen-2.5-32b &28.0& 23.8& 6.7& 23.2& 55.1& 68.3& 27.2& 34.5& 72.2& \textbf{4.3}\\
&Qwen-2.5-72b &29.1& 25.0& 6.7& 22.3& 55.1& 68.4& 24.7& 29.9& 71.6& \textbf{4.3}\\
\midrule
\multirow{5}{*}{\textbf{Fine-Tuning}} &Mistral-7b &25.5& 21.2& 6.2& 24.7& 53.3& 68.0& 20.5& 41.8& 75.4& 3.0 \\
&Gemma-2-9b &\textbf{33.1}& \textbf{29.6}& \textbf{13.1}& \textbf{31.7}& \textbf{59.9}& \textbf{70.7}& 48.6& 30.5& 68.3& 3.6\\
&Llama-3.1-8b &24.3& 19.7& 4.2& 22.7& 52.5& 67.4& 14.9& 47.3& 77.5& 3.8\\
&Qwen-2.5-7b &23.3& 20.2& 5.2& 26.5& 53.3& 67.9& \textbf{13.0}& 59.2& 76.1& 3.5\\
&Qwen-2.5-14b &23.2& 19.8& 5.2& 20.2& 51.1& 66.5& 22.9& 44.8& 75.1& 3.7\\
\bottomrule
\end{tabular}
}
\caption{Performance of different models for the \textbf{argument completion} task on the test split of \textsc{BriefMe} under automatic metrics and transformer-based embedding similarity. R1 stands for Rouge-1, R2 for Rouge-2, BS for BERTScore, LegalBS for LegalBERTScore, Perp. for perplexity, and o3-miniS for o3-mini's average rating. The scale for \texttt{o3-mini}'s scores is: Ineffective (1), Weak (2), Satisfactory (3), Strong (4), and Exemplary (5). } \label{tab:arg_comp_results}

\end{table*}

%% file: tables/arg_summ_realistic.tex
\begin{table*}[t]\centering
\resizebox{\textwidth}{!}{%
\begin{tabular}{llrrrr|rr|rrr|rr|r}\toprule
\textbf{Setup} & \textbf{Model} & \textbf{R1} & \textbf{RL} & \textbf{Bleu} & \textbf{Meteor} & \textbf{BS} & \textbf{LegalBS} & \textbf{Perp.} & \makecell{\textbf{Avg. Tkn}\\\textbf{Common}} & \makecell{\textbf{Avg. Norm}\\\textbf{Levenshtein}} & \textbf{SC-par} & \textbf{SC-sent} & \textbf{o3-miniS} \\\midrule
\textbf{Gold} & Human & - & - & - & - & - & - & 52.7 & - & - & 38.6 & 33.2 & 3.4 \\
\midrule
\multirow{3}{*}{\textbf{Baseline}} & Random & 17.1 & 13.7 & 1.4 & 15.6 & 45.3 & 63.9 & 33.0 & 37.7 & 79.0 & 34.8 & 89.9 & 2.1 \\
 & Lead-1 & 21.6 & 16.8 & 2.2 & 26.4 & 50.5 & 67.3 & 18.7 & 68.7 & 80.9 & 34.8 & 90.5 & 2.0\\
 & BERTExSumm & 25.5 & 20.5 & 3.2 & 25.7 & 51.9 & 68.1 & 34.1 & 45.3 & 75.9 & 96.3 & 91.5 & 2.3 \\
\midrule
\multirow{9}{*}{\textbf{Zero-shot}} & GPT-4o & 30.0 & 23.9 & 3.3 & 29.4 & 56.3 & 70.0 & 45.7 & 36.9 & 75.7 & 34.0 & 28.3 & 4.2\\
 & Mistral-7b & 26.9 & 20.6 & 3.2 & 25.6 & 52.2 & 68.4 & 30.9 & 47.5 & 76.9 & 38.5 & 32.0 & 4.0\\
 & Gemma-2-2b & 30.0 & 25.0 & 3.9 & 24.1 & 52.1 & 68.1 & 51.0 & 31.8 & 74.7 & 38.4 & 33.6 & 3.9\\
 & Gemma-2-9b & 30.2 & 25.4 & \textbf{5.2} & 26.4 & 56.1 & 69.6 & 45.8 & 31.4 & 73.6 & 37.4 & 33.3 & 4.0 \\
 & Llama-3.1-8b & 28.6 & 22.4 & 3.6 & 29.1 & 54.7 & 69.9 & 19.7 & 49.8 & 76.8 & 34.4 & 29.7 & 4.1 \\
 & Llama-3.1-70b & 28.7 & 22.3 & 3.6 & \textbf{31.9} & 55.3 & 70.4 & \textbf{15.8} & 51.7 & 77.3 & 33.4 & 27.3 & 4.1 \\
 & Qwen-2.5-7b & 27.8 & 22.2 & 3.7 & 29.7 & 55.2 & 70.0 & 23.6 & 49.7 & 77.1 & 33.3 & 27.6 & 4.0 \\
 & Qwen-2.5-14b & 29.5 & 23.3 & 4.2 & 30.4 & 55.8 & 70.2 & 23.8 & 42.4 & 75.7 & 33.9 & 29.7 & 4.0 \\
 & Qwen-2.5-32b & 29.0 & 22.8 & 3.4 & 29.7 & 55.5 & 70.0 & 29.6 & 43.9 & 76.4 & 34.4 & 29.2 & 4.0\\
 & Qwen-2.5-72b & 29.9 & 23.8 & 3.2 & 29.9 & 56.1 & 70.0 & 40.4 & 37.8 & 75.6 & 33.4 & 28.5 & 3.9 \\
\midrule
\multirow{9}{*}{\textbf{Few-shot}} & GPT-4o & 31.0 & 24.7 & 4.5 & 30.4 & \textbf{56.5} & \textbf{70.6} & 35.6 & 35.8 & 74.8 & 33.8 & 28.5 & \textbf{4.3}\\
 & Mistral-7b & 18.9 & 14.8 & 1.3 & 18.1 & 46.9 & 64.3 & 18.6 & 49.7 & 79.3 & 38.8 & 35.0 & 2.1 \\
 & Gemma-2-2b & 20.9 & 16.4 & 1.6 & 19.7 & 46.9 & 65.5 & 26.3 & 50.0 & 78.9 & \textbf{44.2} & 35.8 & 2.5 \\
 & Gemma-2-9b & 25.5 & 20.3 & 2.4 & 24.5 & 51.9 & 67.2 & 37.1 & 44.9 & 76.6 & 39.9 & 35.9 & 3.3 \\
 & Llama-3.1-8b & 21.2 & 16.8 & 2.3 & 19.8 & 48.6 & 65.8 & 36.7 & 40.8 & 78.5 & 35.6 & 31.8 & 2.6 \\
 & Llama-3.1-70b & 20.0 & 16.3 & 2.5 & 19.0 & 47.4 & 65.4 & 35.2 & 36.2 & 79.6 & 33.4 & 31.4 & 2.6\\
 & Qwen-2.5-7b & 20.6 & 16.1 & 2.0 & 20.1 & 48.5 & 65.4 & 27.7 & 49.1 & 79.0 & 37.2 & 33.7 & 2.5\\
 & Qwen-2.5-14b & 20.9 & 16.2 & 1.8 & 20.6 & 48.6 & 65.6 & 29.6 & 48.5 & 78.6 & 36.4 & 34.4 & 2.5 \\
 & Qwen-2.5-32b & 21.2 & 16.5 & 2.0 & 21.1 & 48.8 & 65.6 & 26.7 & 49.8 & 78.6 & 35.9 & 34.2 & 2.5\\
 & Qwen-2.5-72b & 21.5 & 17.0 & 1.9 & 21.4 & 49.1 & 65.8 & 24.7 & 48.7 & 78.7 & 36.4 & 35.6 & 2.6 \\
\midrule
\multirow{5}{*}{\textbf{Fine-Tuning}} & Mistral-7b & 31.2 & 25.4 & 4.0 & 29.9 & 55.5 & 69.8 & 21.3 & 41.1 & 74.8 & 34.4 & 31.2 & 3.8\\
 & Gemma-2-9b & \textbf{31.9} & \textbf{27.3} & \textbf{5.2} & 27.6 & 55.9 & 69.4 & 46.4 & 29.5 & 71.5 & 37.4 & \textbf{39.7} & 3.6\\
 & Llama-3.1-8b & 29.7 & 24.0 & 4.0 & 29.8 & 55.2 & 69.9 & 18.2 & 45.5 & 74.8 & 34.2 & 29.9 & 4.0 \\
 & Qwen-2.5-7b & 27.9 & 22.9 & 3.0 & 27.9 & 54.3 & 68.8 & 20.5 & 44.9 & 75.0 & 35.2 & 31.6 & 3.8 \\
 & Qwen-2.5-14b & 28.0 & 23.2 & 2.9 & 28.6 & 54.3 & 69.1 & 21.7 & 46.2 & 75.5 & 37.0 & 36.0 & 3.7\\

\bottomrule
\end{tabular}
}
\caption{Performance of different models for the \textbf{argument summarization} task using the \textbf{raw, unfiltered dataset} under automatic metrics and transformer-based embedding similarity. R1 stands for Rouge-1, 
BS for BERTScore, LegalBS for LegalBERTScore, Perp. for median perplexity, SC-par for median paragraph-level SummaC score, SC-sent for median sentence-level SummaC score, and o3-miniS for \texttt{o3-mini}'s average rating. The scale for \texttt{o3-mini}'s scores is: Ineffective (1), Weak (2), Adequate (3), Strong (4), and Exemplary (5).} \label{tab:arg_summ_realistic}
\end{table*}

%% file: tables/arg_comp_realistic.tex
\begin{table*}[t]\centering
\resizebox{\textwidth}{!}{%
\begin{tabular}{llrrrr|rr|rrr|r}\toprule
\textbf{Setup} &\textbf{Model} &\textbf{R1} &\textbf{RL} &\textbf{Bleu} &\textbf{Meteor} &\textbf{BS} &\textbf{LegalBS} &\textbf{Perp.} &\makecell{\textbf{Avg. Tkn}\\\textbf{Common}} &\makecell{\textbf{Avg. Norm}\\\textbf{Levenshtein}} &\textbf{o3-miniS} \\\midrule
\textbf{Gold} &Human & -& -& -& -& -& -& 54.7& -& -& 3.5 \\
\midrule
\textbf{Baseline} &Random &20.6& 17.2& 4.2& 16.4& 48.3& 65.0& 52.0& 22.6& 75.4& 1.3 \\
\midrule
\multirow{9}{*}{\textbf{Zero-shot}} &GPT-4o& 23.4& 19.8& 4.3& 19.9& 51.5& 65.9& 25.8& 30.7& 73.7& 4.2\\
&Mistral-7b &19.9& 16.3& 1.9& 18.2& 48.2& 65.4& 28.3& 46.7& 78.4& 3.3\\
&Gemma-2-2b &18.8& 15.7& 0.9& 15.7& 44.0& 64.0& 35.7& 47.9& 80.7& 2.9\\
&Gemma-2-9b &19.9& 16.0& 1.9& 19.2& 45.5& 65.2& 35.8& 56.7& 78.5& 4.2\\
&Llama-3.1-8b &16.3& 12.9& 1.3& 18.8& 46.4& 65.0& 23.3& \textbf{58.6}& 81.1& 2.3 \\
&Llama-3.1-70b &19.8& 15.9& 2.4& 20.9& 48.5& 65.7& 18.5& 52.1& 78.8& \textbf{4.3} \\
&Qwen-2.5-7b &22.3& 17.9& 2.6& 20.9& 50.0& 65.9& 21.8& 48.2& 77.3& 3.4 \\
&Qwen-2.5-14b &20.3&16.2& 2.3& 20.2& 49.7& 66.1& 23.4& 50.6& 77.6& 4.2\\
&Qwen-2.5-32b &24.1& 20.2& 3.8& 20.8& 51.9& 66.5& 31.4& 36.2& 74.7& 4.2 \\
&Qwen-2.5-72b &24.9& 21.0& 3.3& 21.2& 52.0& 66.7& 40.4& 29.3& 74.4& 4.0 \\
\midrule
\multirow{9}{*}{\textbf{Few-shot}} &GPT-4o &25.1& 21.7& 4.1& 20.6& 53.6& 66.9& 28.3& 27.4& 72.9& 4.1\\
&Mistral-7b &20.9& 17.2& 2.2& 18.1& 48.8& 65.7& 26.9& 39.6& 77.1& 3.5\\
&Gemma-2-2b &12.1& 10.1& 0.8& 9.3& 41.2& 61.3& 47.7& 34.3& 82.5& 2.1\\
&Gemma-2-9b &22.2& 18.8& 2.5& 18.0& 47.8& 65.1& 41.7& 32.8& 76.2& 3.5\\
&Llama-3.1-8b &23.3& 19.1& 3.2& 19.7& 51.0& 66.3& 42.4& 36.5& 74.3& 3.8\\
&Llama-3.1-70b &26.3& 22.1& 5.0& 22.3& 53.1& 67.2& 40.1& 30.2& 72.7& 4.0\\
&Qwen-2.5-7b &23.3& 19.5& 3.4& 19.5& 51.4& 65.9& 68.3& 32.7& 74.6& 3.9\\
&Qwen-2.5-14b &21.9& 18.5& 2.9& 19.9& 51.0& 66.5& 41.3& 40.0& 75.5& 4.0 \\
&Qwen-2.5-32b &26.0& 22.0& 5.1& 21.9& 53.3& 67.3& 29.2& 32.7& 72.4& 4.2\\
&Qwen-2.5-72b &\textbf{28.5}& \textbf{24.7}& 6.0& \textbf{23.5}& \textbf{55.1}& \textbf{68.2}& 27.8& 28.7& 70.7& 4.2\\
\midrule
\multirow{5}{*}{\textbf{Fine-Tuning}} &Mistral-7b &23.0& 19.2& 3.6& 21.4& 50.2& 66.6& 22.0& 41.6& 75.9& 3.1 \\
&Gemma-2-9b &27.0& 23.0& \textbf{6.3}& 22.5& 52.0& 67.1& 50.9& 29.3& 72.3& 3.5\\
&Llama-3.1-8b &23.1& 18.5& 2.8& 21.2& 50.7& 66.8& 15.6& 45.5& 76.6& 3.8\\
&Qwen-2.5-7b &19.3& 16.3& 2.8& 21.1& 47.8& 65.3& \textbf{12.4}& 58.6& 77.6& 3.3\\
&Qwen-2.5-14b &21.6& 18.4& 3.5& 19.6& 50.0& 66.0& 25.9& 43.5& 74.4& 3.6\\
\bottomrule
\end{tabular}
}
\caption{Performance of different models for the \textbf{argument completion} task with the \textbf{raw, unfiltered dataset} under automatic metrics and transformer-based embedding similarity. R1 stands for Rouge-1, R2 for Rouge-2, BS for BERTScore, LegalBS for LegalBERTScore, Perp. for perplexity, o3-miniS for o3-mini's average rating. The scale for \texttt{o3-mini}'s scores is: Ineffective (1), Weak (2), Satisfactory (3), Strong (4), and Exemplary (5).}\label{tab:arg_comp_realistic}

\end{table*}

%% file: tables/arg_summ_ood_results.tex
\begin{table*}[!htp]\centering
\resizebox{\textwidth}{!}{%
\begin{tabular}{llrrrr|rr|rrr|rr|r}\toprule
\textbf{Setup} &\textbf{Model} &\textbf{R1} &\textbf{RL} &\textbf{Bleu} &\textbf{Meteor} &\textbf{BS} &\textbf{LegalBS} &\textbf{Perp.} &\textbf{Avg. \#Tkn} &\makecell{\textbf{Avg. Norm}\\\textbf{Levenshtein}} &\textbf{SummaC-Par} &\textbf{SummaC-Sent} &\textbf{O3-miniS} \\\midrule
\textbf{Gold} &Human &- &- &- &- &- &- &64.8 &- &- &22.2 &35.2 &3.5 \\
\midrule
\textbf{Baseline} &BERTExSumm &25.8 &21.5 &3.4 &24.1 &51.4 &67.5 &32.4 &38.4 &74.2 &\textbf{87} &\textbf{92.5} &2.3 \\
\midrule
\multirow{10}{*}{\textbf{Zero-shot}} &GPT-4o &\textbf{31.7} &24.1 &4.0 &27.9 &\textbf{56.3} &69.9 &40.6 &36.5 &75.8 &22.2 &29.2 &4.1 \\
&Mistral-7B &26.0 &20.1 &2.4 &21.2 &52.2 &67.7 &31.1 &47.6 &77.1 &22.2 &34.8 &3.9 \\
&Gemma-2-2B &29.4 &24.1 &3.4 &20.1 &52.6 &67.6 &51.6 &31.6 &74.1 &22.2 &38.6 &3.9 \\
&Gemma-2-9B &28.8 &23.9 &4.3 &22.0 &54.6 &68.3 &42.4 &30.5 &74.1 &22.0 &37.2 &4 \\
&Llama-3.1-8B &27.7 &21.8 &3.9 &26.6 &53.8 &68.9 &20.0 &49.3 &77.2 &22.1 &30.8 &4.1 \\
&Llama-3.1-70B &27.6 &21.0 &3.6 &\textbf{29.1} &53.7 &69.5 &\textbf{16.2} &\textbf{53.2} &77.4 &22.1 &29.2 &4.2 \\
&Qwen-2.5-7B &28.6 &22.5 &3.8 &28.5 &54.6 &69.4 &24.7 &50.4 &76.2 &22.2 &29.4 &3.9 \\
&Qwen-2.5-14B &29.4 &22.7 &4.5 &28.1 &54.4 &69.1 &24.2 &42.9 &75.4 &22.2 &31.2 &4.1 \\
&Qwen-2.5-32B &29.3 &22.4 &3.1 &27.9 &54.9 &69.3 &28.5 &44.8 &76.2 &22.1 &30.7 &4 \\
&Qwen-2.5-72B &29.8 &23.7 &3.5 &27.0 &55.1 &69.1 &39.5 &38.5 &75.0 &22.2 &30.1 &3.9 \\
\midrule
\multirow{10}{*}{\textbf{Few-shot}} &GPT-4o &31.4 &24.4 &5.5 &\textbf{29.4} &55.8 &\textbf{70.2} &32.7 &36.3 &74.1 &22.1 &30.2 &\textbf{4.3} \\
&Mistral-7B &19.1 &14.2 &1.6 &17.8 &47.8 &64.5 &18.5 &49.6 &78.8 &21.4 &41.2 &2 \\
&Gemma-2-2B &21.2 &17.1 &2.2 &19.8 &48.6 &65.8 &28.4 &49.9 &77.6 &21.2 &38.4 &2.6 \\
&Gemma-2-9B &23.7 &19.0 &2.6 &20.9 &51.0 &66.2 &38.2 &46.5 &76.3 &21.8 &42.2 &3.3 \\
&Llama-3.1-8B &20.8 &16.2 &2.5 &18.4 &47.5 &64.7 &32.0 &42.7 &78.3 &21.4 &33.7 &2.5 \\
&Llama-3.1-70B &19.2 &15.2 &2.4 &\textbf{17.9} &46.6 &64.3 &33.5 &37.8 &\textbf{79.3} &21.4 &34.0 &2.5 \\
&Qwen-2.5-7B &19.9 &15.6 &1.8 &18.3 &48.5 &64.9 &27.4 &50.3 &78.3 &21.4 &39.1 &2.4 \\
&Qwen-2.5-14B &20.4 &15.8 &2.0 &19.4 &48.3 &65.1 &26.2 &51.4 &77.7 &21.4 &38.4 &2.4 \\
&Qwen-2.5-32B &19.7 &15.3 &1.8 &18.2 &48.2 &64.8 &23.3 &51.0 &78.9 &21.4 &41.9 &2.4 \\
&Qwen-2.5-72B &20.9 &16.5 &1.5 &18.6 &48.6 &65.0 &25.2 &49.7 &77.7 &21.4 &40.7 &2.5 \\
\midrule
\multirow{5}{*}{\textbf{Fine-tuned}} &Mistral-7B &29.9 &24.2 &3.2 &25.3 &54.0 &68.8 &22.8 &40.7 &75.0 &22.2 &31.2 &3.6 \\
&Gemma-2-9B &\textbf{31.7} &\textbf{27.1} &\textbf{5.6} &24.4 &55.2 &68.6 &45.4 &30.3 &71.2 &22.1 &40.0 &3.7 \\
&Llama-3.1-8B &27.5 &21.2 &3.0 &25.4 &53.0 &68.4 &19.9 &43.9 &75.7 &22.2 &32.0 &4 \\
&Qwen-2.5-7B &28.0 &22.9 &2.4 &24.8 &53.9 &68.1 &21.9 &45.0 &75.7 &22.2 &32.3 &3.8 \\
&Qwen-2.5-14B &27.2 &22.0 &3.0 &24.6 &53.6 &68.2 &24.0 &45.2 &76.6 &22.2 &37.4 &3.8 \\
\bottomrule
\end{tabular}
}
\caption{Performance of different models for the \textbf{argument summarization} task on the \textbf{held-out set} under automatic metrics and transformer-based embedding similarity. R1 stands for Rouge-1, 
BS for BERTScore, LegalBS for LegalBERTScore, Perp. for median perplexity, SC-par for median paragraph-level SummaC score, SC-sent for median sentence-level SummaC score, and o3-miniS for \texttt{o3-mini}'s average rating. The scale for \texttt{o3-mini}'s scores is: Ineffective (1), Weak (2), Adequate (3), Strong (4), and Exemplary (5).} \label{tab:arg_summ_ood}
\end{table*}

%% file: tables/arg_comp_ood_results.tex
\begin{table*}[!htp]\centering
\resizebox{\textwidth}{!}{%
\begin{tabular}{llrrrr|rr|rrr|r}\toprule
& &\textbf{R1} &\textbf{RL} &\textbf{Bleu} &\textbf{Meteor} &\textbf{BS} &\textbf{LegalBS} &\textbf{Perp.} &\textbf{Avg. \#Tkn} &\makecell{\textbf{Avg. Norm}\\\textbf{Levenshtein}} &\textbf{O3-miniS} \\\midrule
\textbf{Gold} &Human &- &- &- &- &- &- &68.8 &27.0 &- &3.7 \\
\midrule
\multirow{10}{*}{\textbf{Zero-shot}} &GPT-4o &29.8 &25.4 &7.5 &26.1 &58.0 &69.1 &23.7 &29.2 &70.8 &4.1 \\
&Mistral-7B &21.9 &16.6 &2.6 &20.1 &53.1 &67.3 &28.0 &48.0 &77.1 &2.9 \\
&Gemma-2-2B &19.5 &15.4 &1.4 &16.3 &48.3 &65.7 &34.0 &45.7 &79.7 &2.5 \\
&Gemma-2-9B &22.4 &17.4 &2.6 &22.9 &49.2 &66.2 &34.1 &55.9 &77.5 &4 \\
&Llama-3.1-8B &18.4 &13.9 &2.3 &22.0 &49.4 &66.0 &24.0 &57.6 &79.6 &2.3 \\
&Llama-3.1-70B &22.8 &18.4 &3.7 &23.5 &52.0 &67.5 &19.0 &51.8 &77.6 &\textbf{4.2} \\
&Qwen-2.5-7B &27.4 &21.8 &6.0 &28.1 &57.2 &69.0 &21.3 &48.0 &75.0 &3.3 \\
&Qwen-2.5-14B &23.9 &20.0 &3.6 &24.1 &54.0 &67.8 &23.7 &48.3 &75.3 &4 \\
&Qwen-2.5-32B &27.4 &22.7 &5.7 &25.4 &56.4 &68.8 &29.3 &37.3 &72.8 &\textbf{4.2} \\
&Qwen-2.5-72B &\textbf{31.4} &26.5 &6.2 &28.1 &\textbf{58.7} &\textbf{70} &38.5 &30.0 &71.3 &4 \\
\midrule
\multirow{10}{*}{\textbf{Few-shot}} &GPT-4o &26.2 &21.7 &3.1 &17.8 &53.8 &67.1 &25.5 &27.7 &74.1 &\textbf{4.2} \\
&Mistral-7B &23.2 &18.3 &2.3 &19.4 &53.2 &67.6 &28.0 &39.4 &76.0 &3.3 \\
&Gemma-2-2B &13.5 &11.2 &0.9 &10.7 &43.7 &62.3 &52.5 &36.2 &\textbf{82.4} &1.9 \\
&Gemma-2-9B &22.0 &17.7 &2.0 &18.0 &49.5 &65.9 &41.8 &37.9 &76.6 &3.1 \\
&Llama-3.1-8B &24.2 &18.8 &4.0 &20.5 &51.7 &66.7 &35.1 &35.1 &74.4 &3.8 \\
&Llama-3.1-70B &27.8 &23.3 &5.6 &22.1 &54.2 &67.8 &38.0 &30.8 &72.4 &4 \\
&Qwen-2.5-7B &25.4 &20.7 &3.0 &19.1 &52.5 &66.6 &68.1 &33.8 &74.8 &3.9 \\
&Qwen-2.5-14B &23.3 &19.0 &3.6 &20.4 &51.8 &66.6 &40.5 &39.0 &75.1 &4 \\
&Qwen-2.5-32B &27.6 &23.2 &5.4 &22.9 &54.7 &68.1 &25.4 &32.5 &72.6 &4.1 \\
&Qwen-2.5-72B &28.1 &23.5 &4.3 &21.0 &54.3 &67.8 &27.3 &28.5 &72.0 &4.1 \\
\midrule
\multirow{5}{*}{\textbf{Fine-tuned}} &Mistral-7B &26.8 &22.3 &5.9 &24.1 &54.2 &68.5 &22.8 &40.2 &73.5 &2.8 \\
&Gemma-2-9B &\textbf{31.4} &\textbf{27.1} &\textbf{10.9} &\textbf{28.9} &57.7 &69.5 &52.9 &27.9 &68.7 &3.3 \\
&Llama-3.1-8B &24.4 &19.4 &4.4 &21.5 &52.3 &67.3 &16.9 &40.1 &74.8 &3.7 \\
&Qwen-2.5-7B &23.5 &19.4 &4.1 &25.0 &53.1 &67.4 &\textbf{13.5} &\textbf{57.7} &76.4 &3.3 \\
&Qwen-2.5-14B &22.6 &18.8 &4.4 &19.0 &51.0 &66.1 &26.3 &39.5 &73.9 &3.6 \\
\bottomrule
\end{tabular}
}
\caption{Performance of different models for the \textbf{argument completion} task on the \textbf{held-out set} under automatic metrics and transformer-based embedding similarity. R1 stands for Rouge-1, R2 for Rouge-2, BS for BERTScore, LegalBS for LegalBERTScore, Perp. for perplexity, and o3-miniS for o3-mini's average rating. The scale for \texttt{o3-mini}'s scores is: Ineffective (1), Weak (2), Satisfactory (3), Strong (4), and Exemplary (5). This table presents the performance of the top-performing model from each model family based on the o3-miniS evaluation. For a complete list of all tested models, see the Appendix. } \label{tab:arg_comp_ood}

\end{table*}

%% file: tables/case_ret_result_len.tex
\begin{table*}[!htp]\centering
\resizebox{0.9\textwidth}{!}{
\begin{tabular}{lll|rrrrr|rr}\toprule
\textbf{Model} &\textbf{Query Len. Bin (\#tokens)} &\textbf{\%Queries} &\textbf{R@1} &\textbf{R@5} &\textbf{R@10} &\textbf{R@50} &\textbf{R@100} &\textbf{MRR@10} &\textbf{nDCG@10} \\\midrule
\multirow{6}{*}{\textbf{BM25}} &$\leq$ 29 &1.8 &12.5 &25.0 &25.0 &25.0 &25.0 &18.8 &20.4 \\
&30-59 &12.9 &6.9 &13.8 &20.7 &50.0 &56.9 &10.0 &12.4 \\
&60-89 &40.5 &7.7 &21.4 &28.6 &48.4 &57.1 &13.4 &17.0 \\
&90-119 &29 &4.6 &17.7 &23.8 &41.5 &50.8 &10.3 &13.5 \\
&120-149 &12.7 &10.5 &22.8 &33.3 &49.1 &61.4 &16.5 &20.4 \\
&$\geq$ 150 &3.1 &21.4 &21.4 &50.0 &57.1 &57.1 &24.6 &30.1 \\
\midrule
\multirow{6}{*}{\textbf{DPR}} &$\leq$ 29 &1.8 &0.0 &12.5 &12.5 &12.5 &12.5 &2.5 &4.8 \\
&30-59 &12.9 &1.7 &6.9 &12.1 &24.1 &25.9 &4.3 &6.1 \\
&60-89 &40.5 &0.5 &2.2 &4.4 &14.8 &19.8 &1.4 &2.1 \\
&90-119 &29 &1.5 &6.2 &6.2 &13.1 &19.2 &3.0 &3.8 \\
&120-149 &12.7 &1.8 &3.5 &3.5 &14.0 &17.5 &2.3 &2.6 \\
&$\geq$ 150 &3.1 &0.0 &0.0 &0.0 &7.1 &7.1 &0.0 &0.0 \\
\midrule
\multirow{6}{*}{\textbf{ColBERT}} &$\leq$ 29 &1.8 &0.0 &25.0 &25.0 &25.0 &25.0 &6.7 &11.1 \\
&30-59 &12.9 &13.6 &20.3 &28.8 &42.4 &52.5 &16.4 &19.2 \\
&60-89 &40.5 &10.4 &22.5 &29.7 &45.6 &53.8 &15.4 &18.7 \\
&90-119 &29 &7.8 &16.3 &24 &40.3 &48.8 &11.9 &14.8 \\
&120-149 &12.7 &15.8 &28.1 &38.6 &49.1 &54.4 &21.9 &25.8 \\
&$\geq$ 150 &3.1 &14.3 &21.4 &21.4 &28.6 &42.9 &17.9 &18.8 \\
\bottomrule
\end{tabular}
}
\caption{The average query length and retrieval performance metrics (R@1 to R@100, MRR@10, and nDCG@10) across different query length bins and methods. 
}\label{tab:case_ret_results_len_based}
\end{table*}

%% file: figures/score_freq.tex
\begin{figure*}[htp]
    \centering
    \begin{subfigure}[b]{0.49\textwidth}
        \centering
        \includegraphics[width=\textwidth]{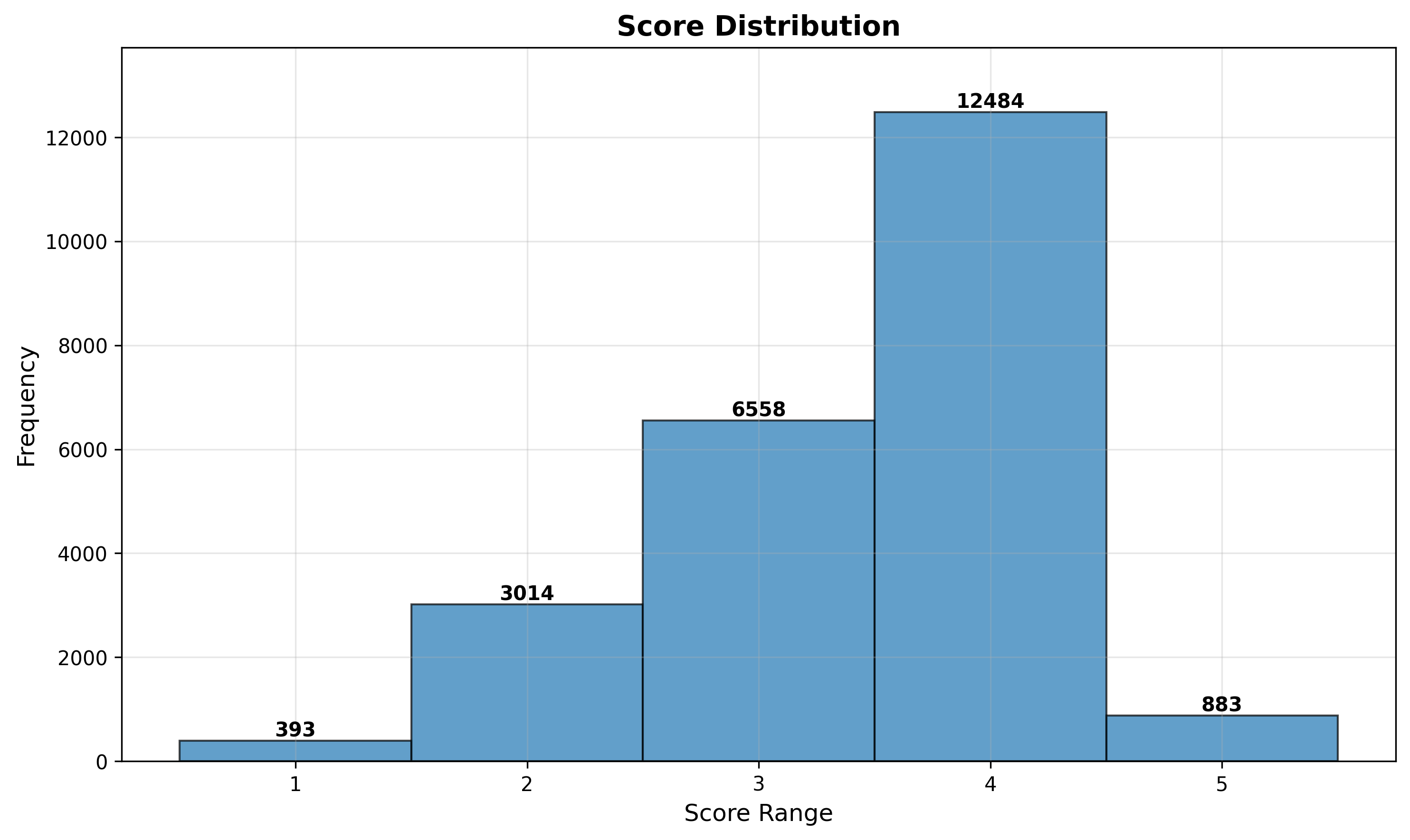}  
        \caption{}
        \label{fig:arg_summ_score_freq}
    \end{subfigure}
    \hfill
    \begin{subfigure}[b]{0.49\textwidth}
        \centering
        \includegraphics[width=\textwidth]{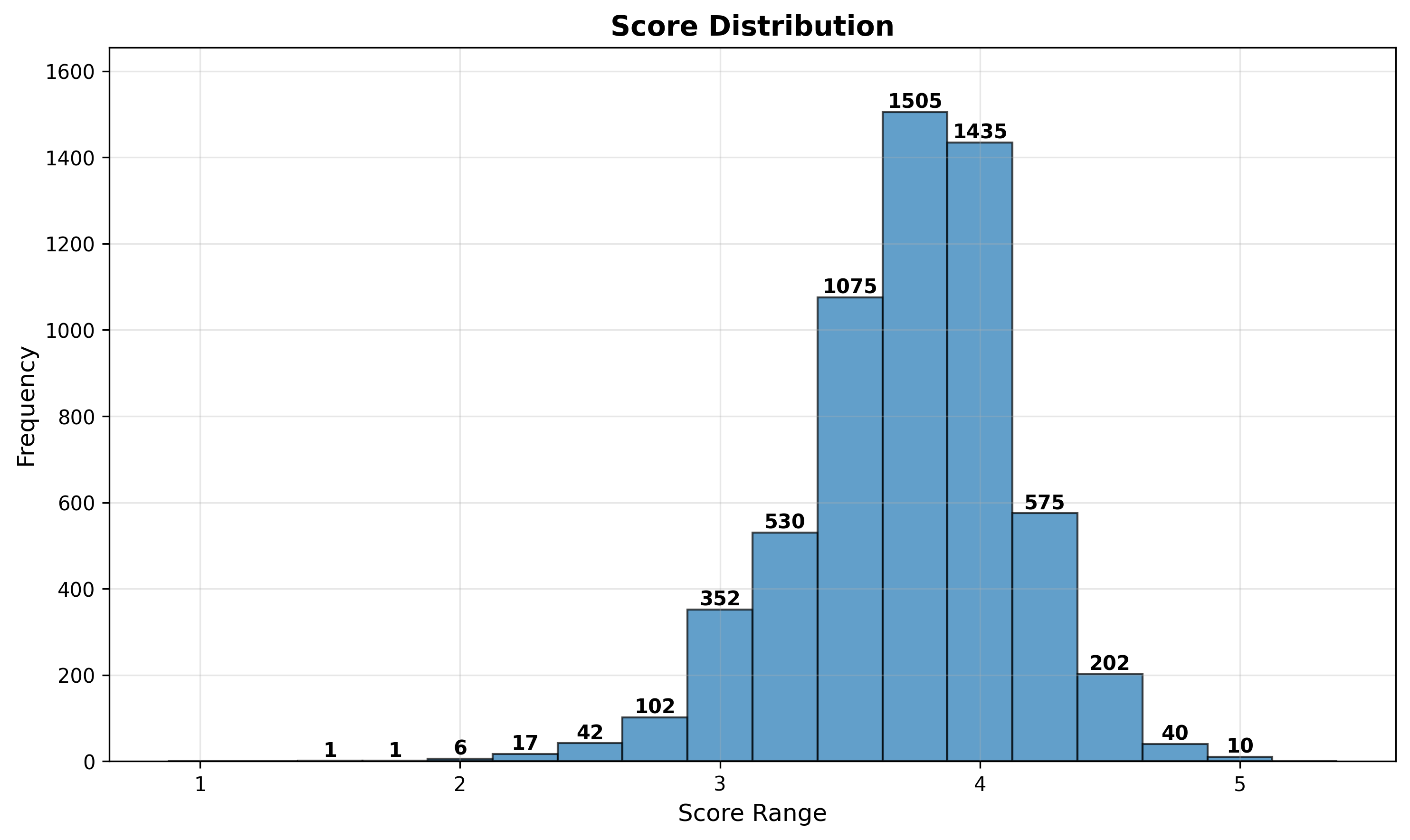} 
        \caption{}
        \label{fig:arg_comp_score_freq}
    \end{subfigure}
    \caption{(a) Argument Summarization Task: Distribution of LLM-as-judge ratings for how well the human-authored headings summarize the corresponding section text. (b) Argument Completion Task: Distribution of average LLM-as-judge ratings for how well each human-authored heading in the Table of Contents fits with the rest of the headings. }
    \label{fig:llm_score_freq}
\end{figure*}

%% file: figures/argsumm_inst-old.tex
\begin{figure}[t]
    \centering
    \includegraphics[width=0.98\linewidth]{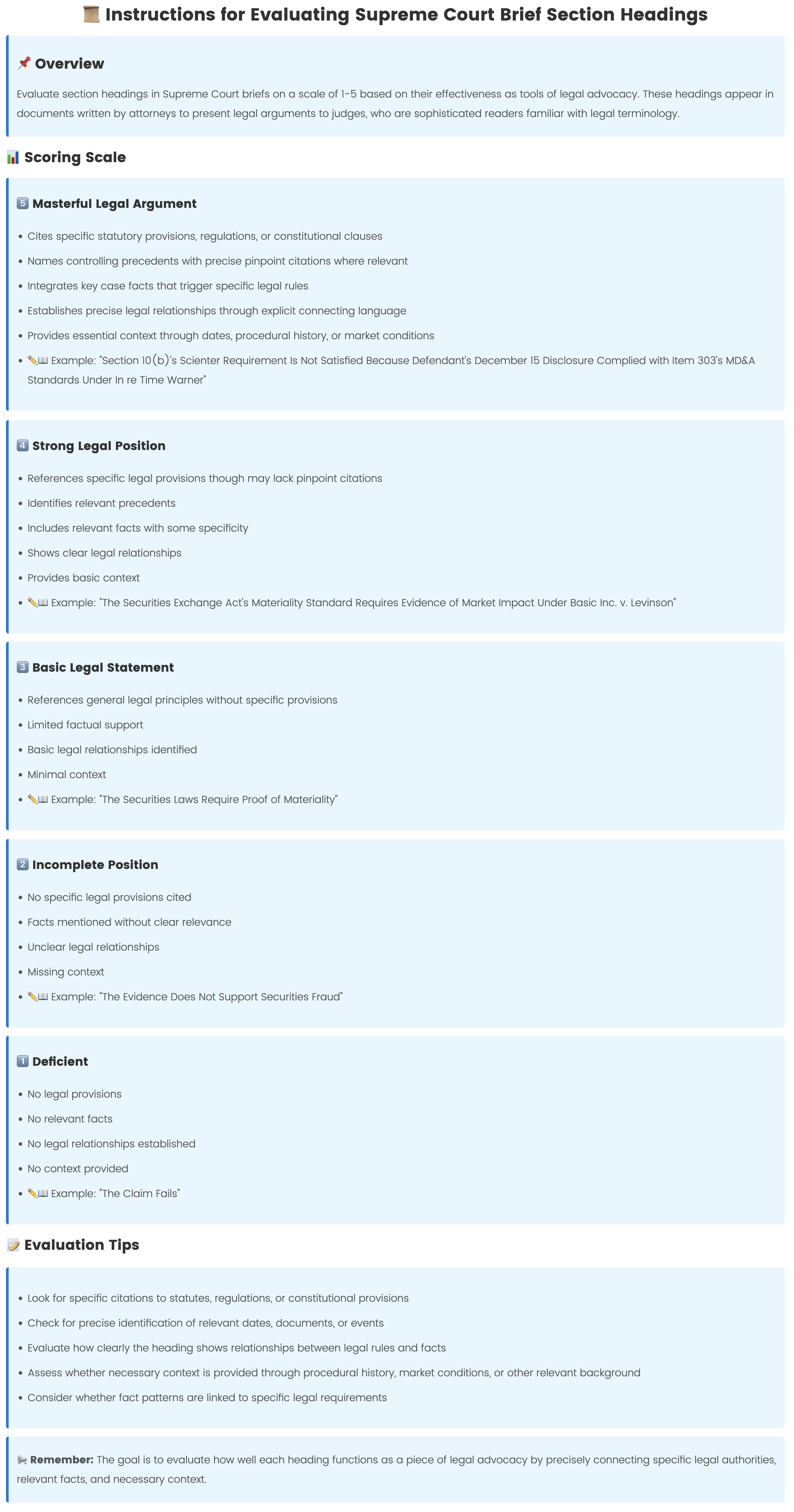}
    \caption{Guidelines provided to annotators for the argument summarization task in the initial annotation study.}
    \label{fig:arg-summ-inst-old}
\end{figure}

%% file: figures/argsumm_example_study-old.tex
\begin{figure*}
    \centering
   \includegraphics[width=0.99\linewidth]{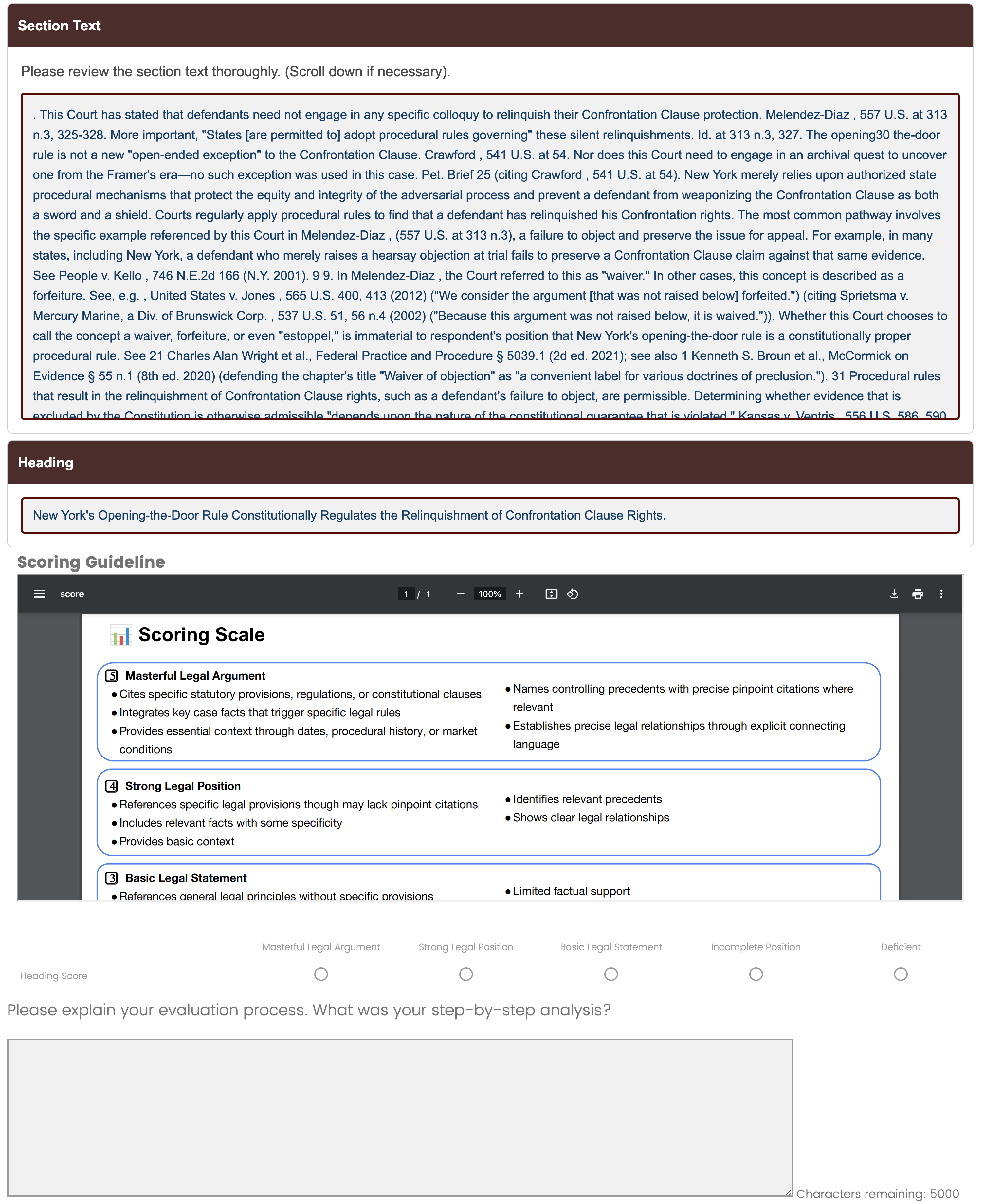}
    \caption{Example of argument summarization annotation task in the initial annotation study.}
    \label{fig:arg-summ-ex-old}
\end{figure*}

%% file: figures/ArgSumm-Inst-2.tex
\begin{figure}
    \centering
    \includegraphics[width=0.79\linewidth]{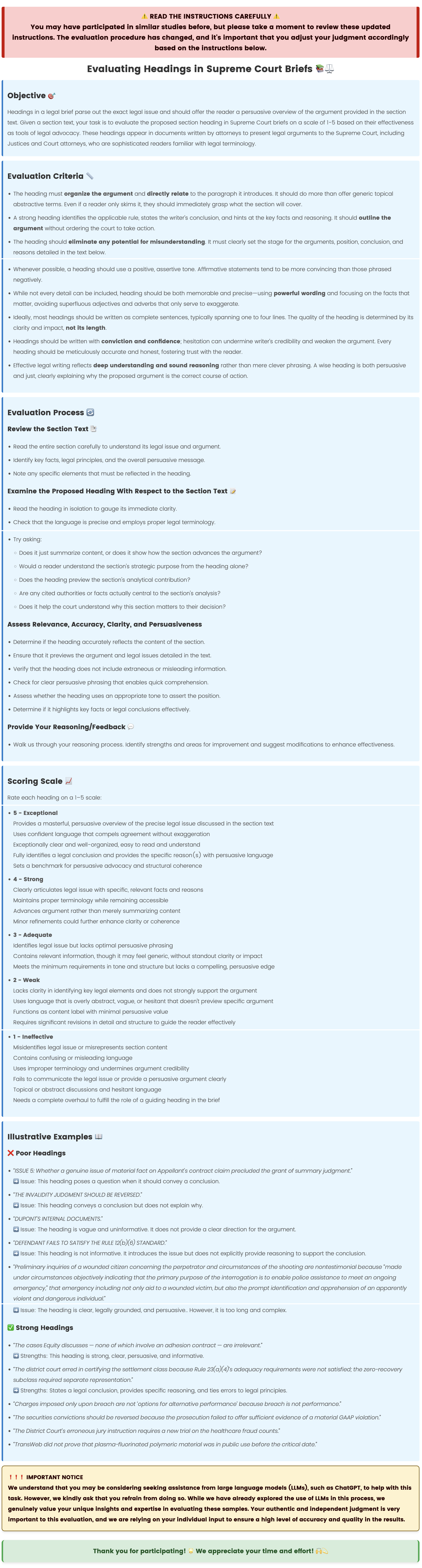}
    \caption{Guidelines provided to annotators for the argument summarization task in the enhanced annotation study.}
    \label{fig:arg-summ-inst}
\end{figure}

%% file: figures/ArgComp-Inst-2.tex
\begin{figure}
    \centering
    \includegraphics[width=0.79\linewidth]{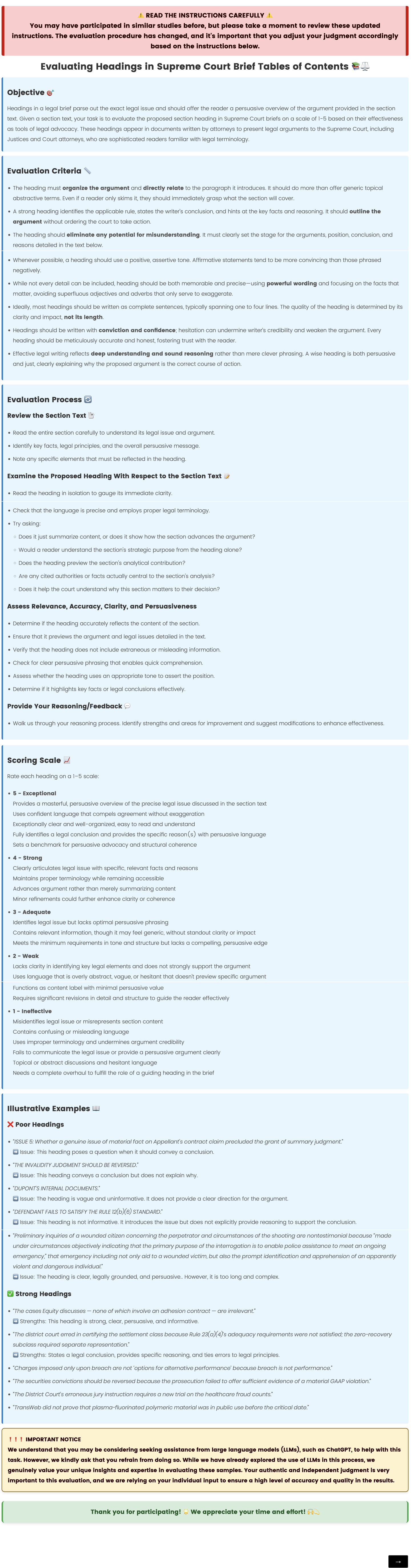}
    \caption{Guidelines provided to annotators for the argument completion task in the enhanced annotation study.}
    \label{fig:arg-comp-inst}
\end{figure}

%% file: figures/ArgComp-Conc-Ex.tex
\begin{figure*}
    \centering
    \includegraphics[width=0.79\linewidth]{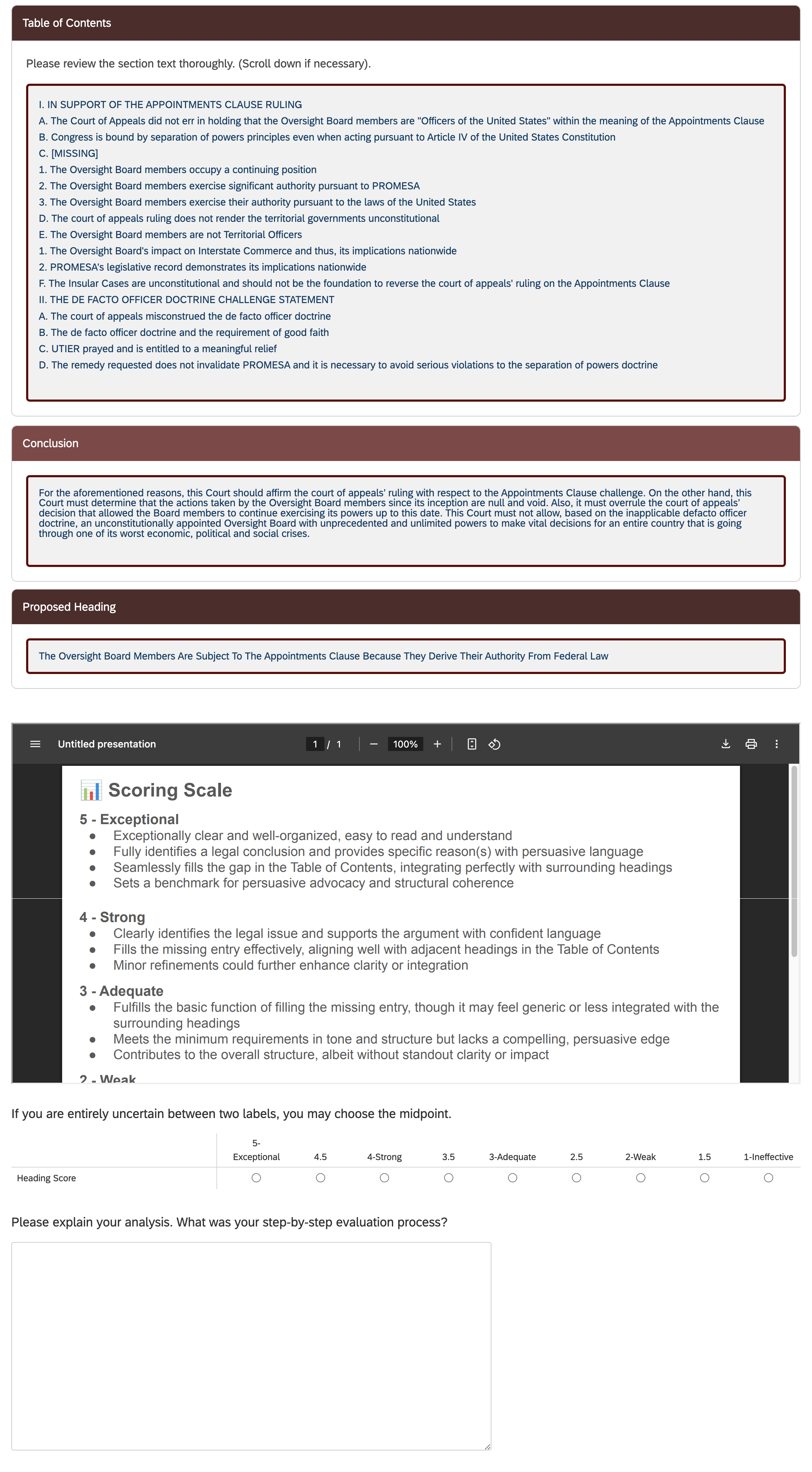}
    \caption{Example of argument completion annotation task in the enhanced annotation study.}
    \label{fig:arg-comp-ex}
\end{figure*}

%% file: tables/evaluation_of_raters.tex
\begin{table*}[t]\centering
\resizebox{0.58\textwidth}{!}{%
\begin{tabular}{l|rr}\toprule
\textbf{Judge} &\textbf{Argument Summarization} &\textbf{Argument Completion} \\\midrule
\textbf{User1} &2.4 &3.1 \\
\textbf{User2} &4.2 &4.3 \\
\textbf{User3} &3.9 &3.3 \\
\texttt{\textbf{o3-mini}}$^\dagger$ &\textbf{4.6} &\textbf{4.9} \\
\bottomrule
\end{tabular}
}
\caption{The average meta-ratings assigned to human and LLM ratings for 10 samples per task. 
 $\dagger$\texttt{o3-mini-2025-01-31}.}
 \label{tab:evaluation_of_raters}
\end{table*}

%% file: tables/old_human_exp_compare_with_o3mini.tex
\begin{table*}[!htp]\centering
\resizebox{\textwidth}{!}{%
\begin{tabular}{lrr|r}\toprule
&\multicolumn{2}{c}{\textbf{Human Annotators }} &\textbf{o3-mini } \\\cmidrule{2-4}
&\textbf{Avg. of Means} &\textbf{Avg. of SDs } &\textbf{Mean} \\\midrule
All (42) &3.37 &0.65 &3.48 \\
With 3 annotations (35) &3.56 &0.78 &- \\
With 1 annotations (7) &2.43 &- &- \\
Scored heading is human-generated (23) &3.19 &0.67 &3.43 \\
Scored heading is LLM-generated (19) &3.6 &0.63 &3.53 \\
Scored heading is human-generated - With 3 annotations (19) &3.39 &0.82 &3.47 \\
Scored heading is LLM-generated - With 3 annotations (16) &3.77 &0.74 &3.62 \\
\bottomrule
\end{tabular}
}
\caption{A comparison of the average and standard deviation of the mean of three annotators’ ratings vs. the average
o3-mini ratings, broken down by number of annotations (1 or 3) and heading generation source (human vs. LLM).
The numbers in parentheses show the sample count. We use GPT-4o generations produced in a few-shot setup for
this study. The scale for o3-mini’s scores is: Deficient (1), Incomplete Position (2), Basic Legal Statement (3),
Strong Legal Position (4), and Masterful Legal Argument (5).}\label{tab:old_arg_summ_score_user_vs_o3mini}
\end{table*}

%% file: figures/user_variance_dist.tex
\begin{figure*}
    \centering
    \includegraphics[width=0.99\linewidth]{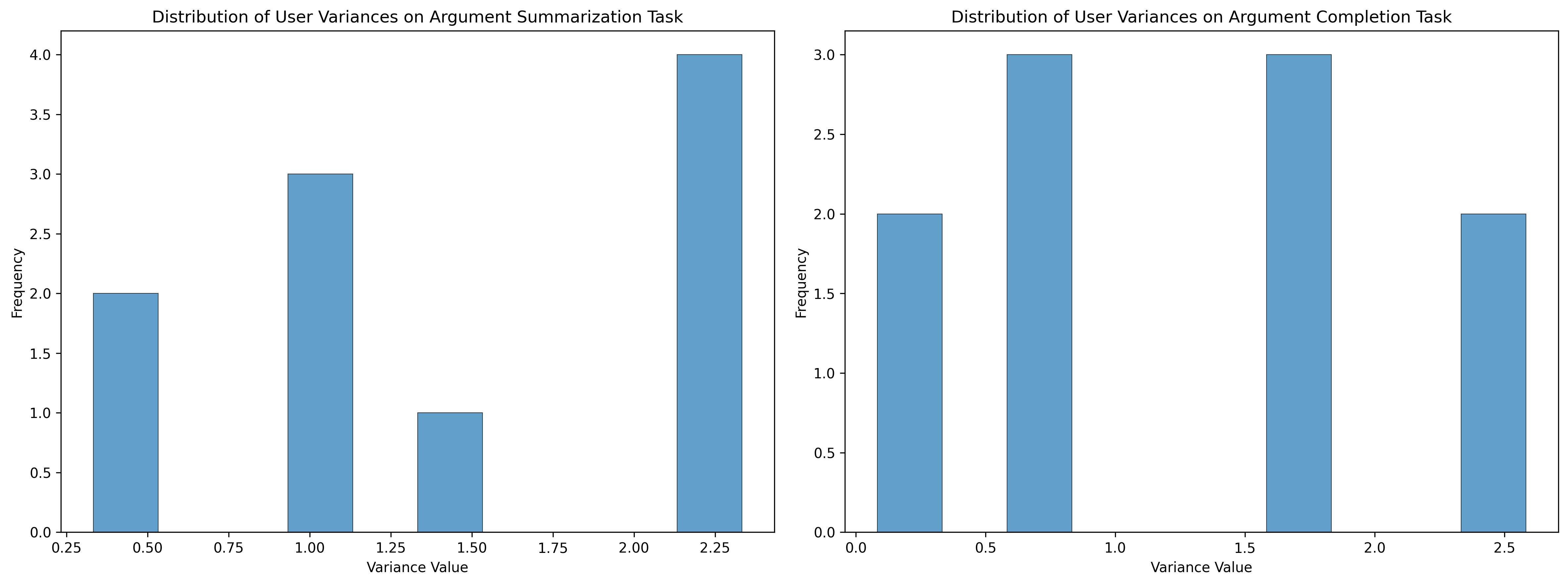}
    \caption{Distribution of per-sample variance among 3 human rates for 10 samples of argument summarization (left), with a mean variance of 1.43, and argument completion (right), with a mean variance of 1.22.}
    \label{fig:user_variance_dist}
\end{figure*}

%% file: figures/human_study_user_score_and_metareview-argsumm.tex
\begin{figure*}
    \centering
   \includegraphics[width=0.99\linewidth]{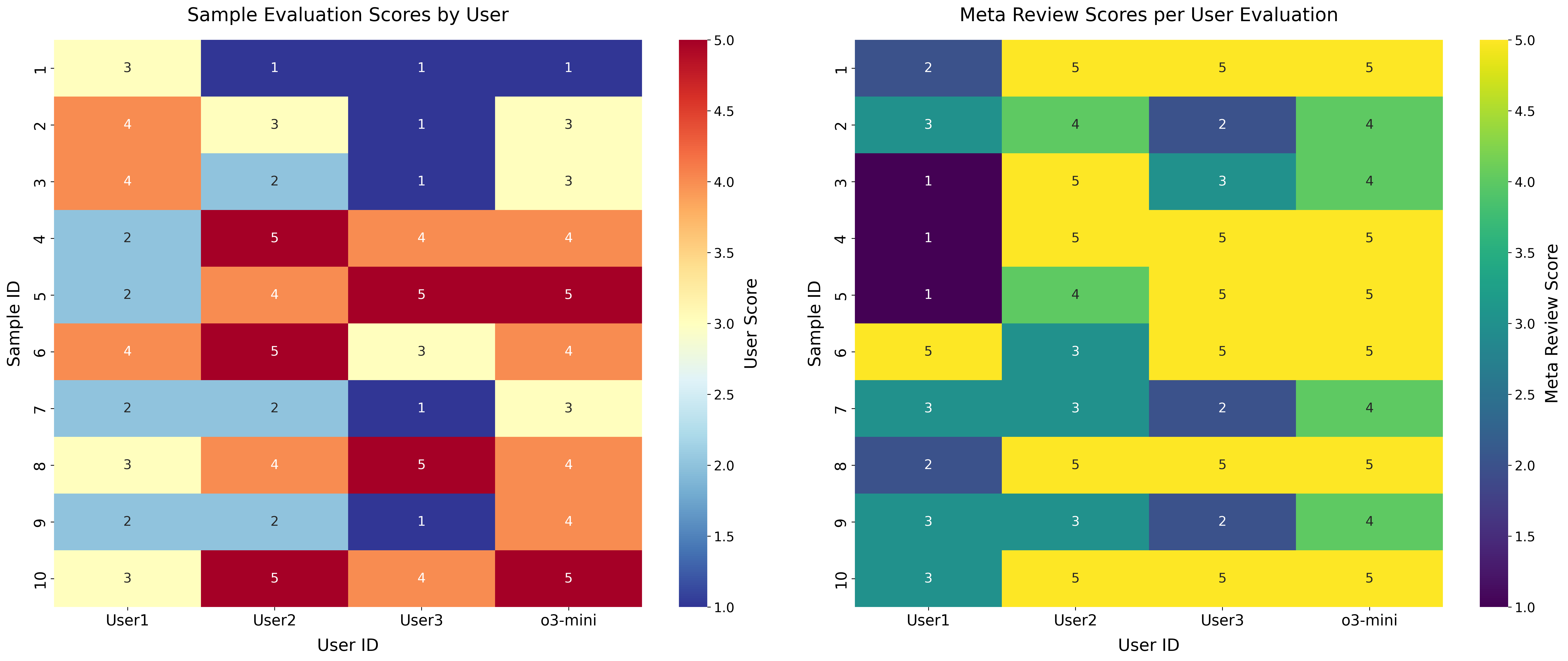}
    \caption{The left heatmap shows human ratings for 10 \textbf{headings serving as section summaries} across four evaluators: User1, User2, User3, and \texttt{o3-mini}. The right heatmap shows the corresponding meta-review scores assigned by one qualified author of this paper. \texttt{o3-mini} receives consistently high meta review scores, while human users show more variability.}
    \label{fig:arg-summ-metareview-heatmap}
\end{figure*}

%% file: figures/human_study_user_score_and_metareview-argcomp.tex
\begin{figure*}
    \centering
   \includegraphics[width=0.99\linewidth]{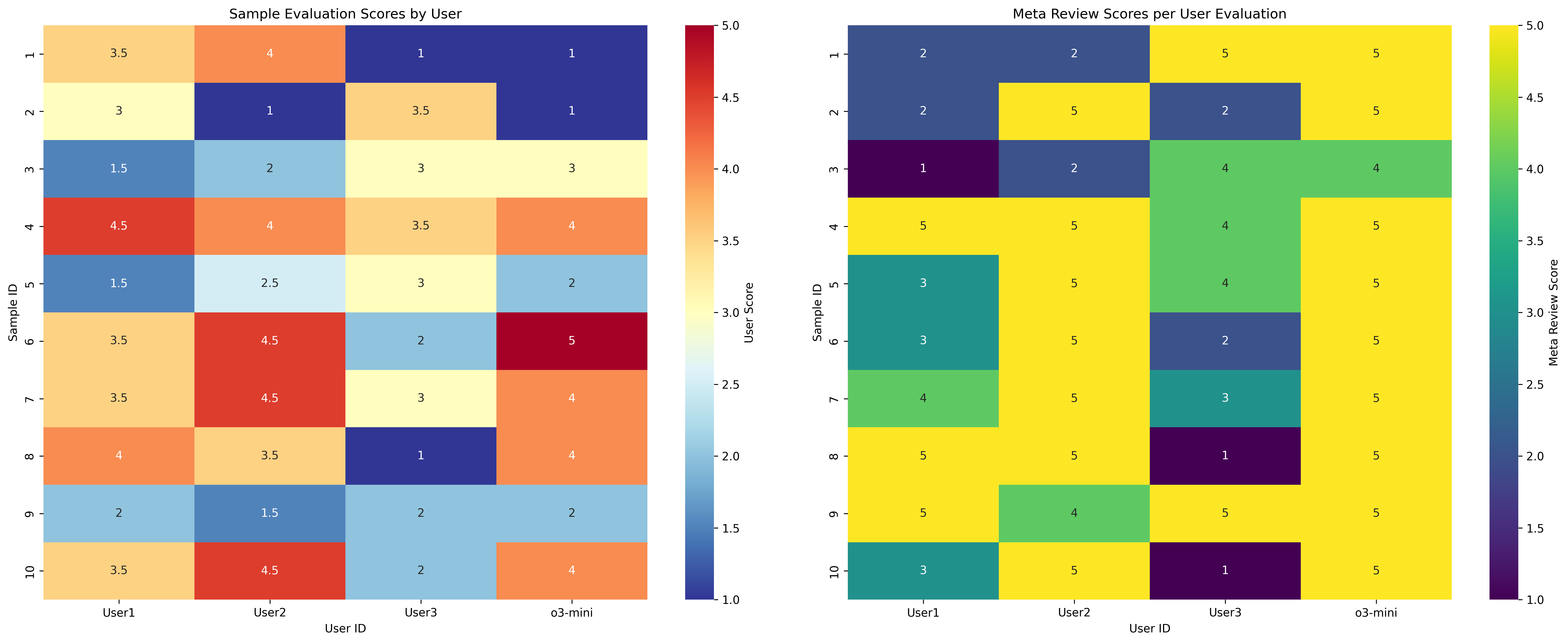}
\caption{The left heatmap shows human ratings for 10 \textbf{headings serving as completions of tables of contents} across four evaluators: User1, User2, User3, and \texttt{o3-mini}. The right heatmap shows the corresponding meta-review scores assigned by one qualified author of this paper. \texttt{o3-mini} receives consistently high meta review scores, while human users show more variability.}
    \label{fig:arg-comp-metareview-heatmap}
\end{figure*}

%% file: tables/user_justification.tex
\begin{table*}[!htp]\centering
\resizebox{\textwidth}{!}{%
\scriptsize
\begin{tabular}{p{2cm}p{1cm}p{6.5cm}}\toprule
\textbf{Task} &\textbf{User ID} &\textbf{Justification} \\\midrule
Arg. Summarization &1 &The text above is too short. It seems boring and very uninteresting. The text does not look like it would attract much readers due to its short length. I therefore conclude that the is weak and almost ineffective \\
\midrule
Arg. Summarization &3 &I first reviewed the Section Text before rating the Proposed Heading. I thought that the Proposed Heading was excellent, related well to the Section Text, and concisely and clearly identified the legal issues addressed. \\
\midrule
Arg. Completion &3 &I first reviewed the Conclusion before looking at the Table of Contents and the Proposed Heading. While I think the heading does state a relevant proposition and does fit within the overall structure of the Table of Contents, it is fairly short and not as effective as it could be. \\
\bottomrule
\end{tabular}
}
\caption{A few samples of the over-generic justifications provided by expert users who annotated samples for the two tasks of argument summarization and completion. }\label{tab:user_justification}
\end{table*}

%% file: tables/error_analysis.tex
\begin{table*}[t]\centering

\begin{tabular}{llrrrrr|r}\toprule
\textbf{Task} & \textbf{Model (Setup)}  & \multicolumn{6}{c}{\makecell{\textbf{Num. Examples Reviewed}\\\textbf{For Each LLM Judge Score}}}\\
& & \textbf{1} & \textbf{2} & \textbf{3} & \textbf{4} & \textbf{5} & \textbf{Total}\\
\midrule
\multirow{3}{*}{\textbf{Summarization}} & GPT-4o (Zero-shot) & - & - & - & 10 & 15 & 25\\
& GPT-4o (Few-shot) & - & - & - & 10 & 15 & 25 \\
& Llama-3.1-70b (Zero-shot) & 17 & 13 & 10 & 10 & 5 & 55 \\
\midrule
\multirow{3}{*}{\textbf{Completion}} & GPT-4o (Zero-shot) & - & 2 & 23 & 15 & 15 & 55 \\
& GPT-4o (Few-shot) & - & 1 & 13 & 15 & 15 & 44 \\
& Qwen-2.5-14b (Zero-shot) & 1 & 2 & 15 & 15 & 15 & 48 \\
\bottomrule
\end{tabular}

\caption{Summary of the examples reviewed for error analysis of the summarization and completion tasks.} \label{tab:error_analysis}
\end{table*}

%% file: tables/error_analysis_retrieval.tex
\begin{table*}[t]\centering
\resizebox{0.58\textwidth}{!}{%
\begin{tabular}{l|rrr}\toprule
\textbf{Model} &\textbf{1+ Match} &\textbf{All Match} &\textbf{Total Samples Reviewed}\\\midrule
\textbf{BM25} &31 &15 &40\\
\textbf{ColBERT} &32 &17 &40\\
\bottomrule
\end{tabular}
}
\caption{Summary of error analysis samples reviewed and results. The author with legal expertise reviewed the input text, gold reference case, and top 5 ranked cases retrieved by two models. 1+ Match means at 1 of the top 5 retrieved cases had the same topic as the gold reference case. All Match means that all top 5 has the same topic. }\label{tab:error_analysis_retr}
\end{table*}

%% file: figures/code_ex_1.tex
\begin{figure*}
    \centering
    \includegraphics[width=0.99\linewidth]{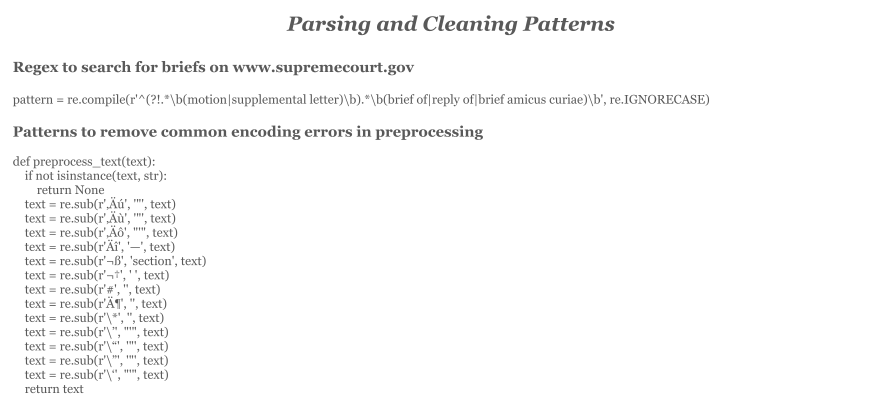}
    \caption{This figure shows the regular expression patterns used to identify briefs on supremecourt.gov and clean common encoding errors. Although the patterns themselves are straightforward, find the right elements to include in each one entailed significant trial and error.}
    \label{fig:code_ex_1}
\end{figure*}

%% file: figures/regex_extract.tex
\begin{figure*}
    \centering
    \includegraphics[width=0.99\linewidth]{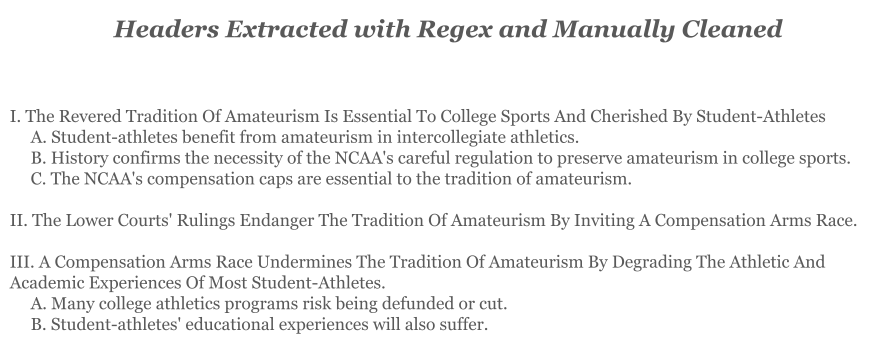}
    \caption{An example of the header arguments extracted by regular expressions and manually cleaned.}
    \label{fig:regex_extract}
\end{figure*}

%% file: figures/mistral_extract.tex
\begin{figure*}
    \centering
    \includegraphics[width=0.99\linewidth]{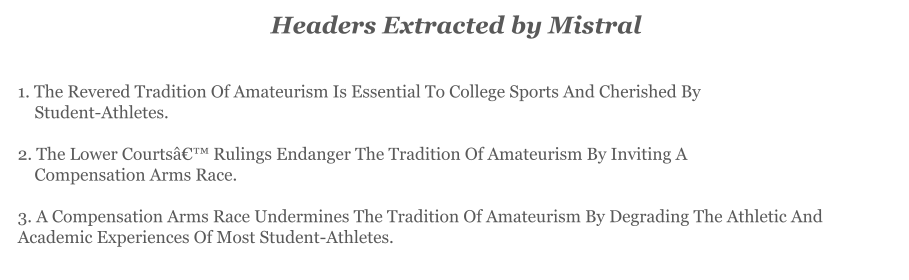}
    \caption{An example of the header arguments extracted Mistral. The model extracted the top level headers but failed to capture the child headers.}
    \label{fig:mistral_extract}
\end{figure*}

%% file: figures/code_ex_2.tex
\begin{figure*}
    \centering
    \includegraphics[width=0.99\linewidth]{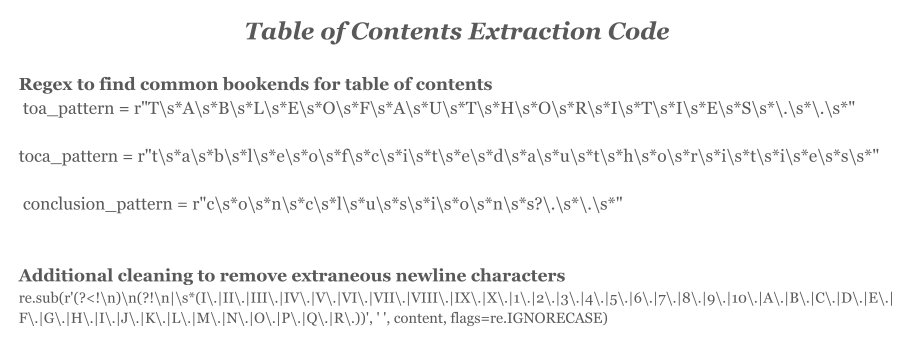}
    \caption{This figure shows regular expressions used to find terms that indicated the beginning of the table of contents. Expressions needed to be flexible to account for random whitespace inserted in between characters from the pdf extraction. Extraneous newlines caused numerous issues in the pipeline, so this code sought to remove them while keeping newlines that indicated the beginning of a new section of text.}
    \label{fig:code_ex_2}
\end{figure*}

%% file: figures/code_ex_3.tex
\begin{figure*}
    \centering
    \includegraphics[width=0.99\linewidth]{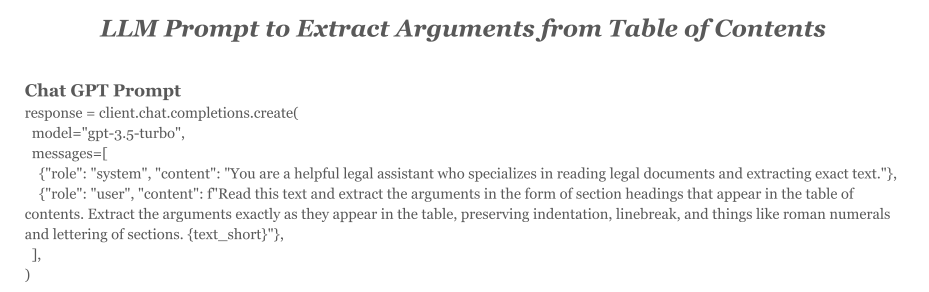}
    \caption{This figure shows the prompt with GPT 3.5-Turbo to assist in the argument extraction process.}
    \label{fig:code_ex_3}
\end{figure*}

%% file: figures/sum_data_ex1.tex
\begin{figure*}
    \centering
    \includegraphics[width=0.99\linewidth]{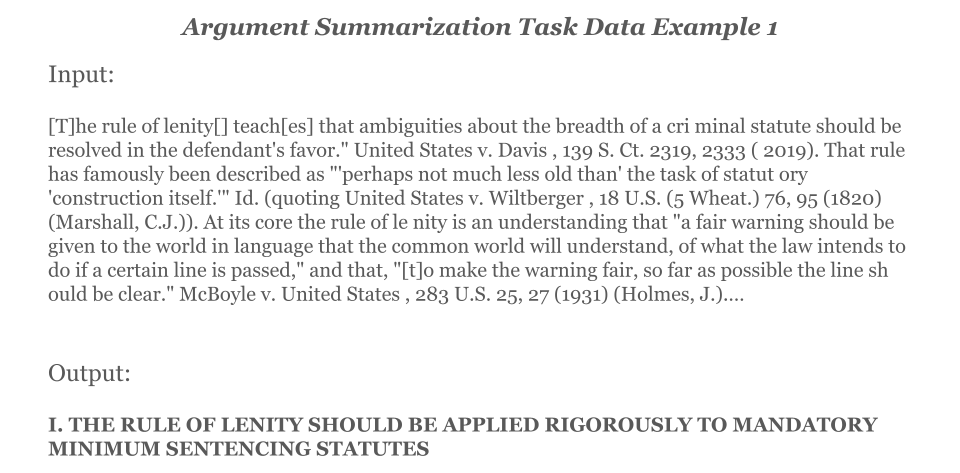}
    \caption{Summarization Task Data Example 1. This heading captures the argument of the brief about rule of lenity and its application to mandatory minimum sentencing. Although the full text is too long to reproduce, here we see that it starts from the premise of what the rule of lenity is and the values it protects.}
    \label{fig:sum_data_1}
\end{figure*}

%% file: figures/sum_data_ex2.tex
\begin{figure*}
    \centering
    \includegraphics[width=0.99\linewidth]{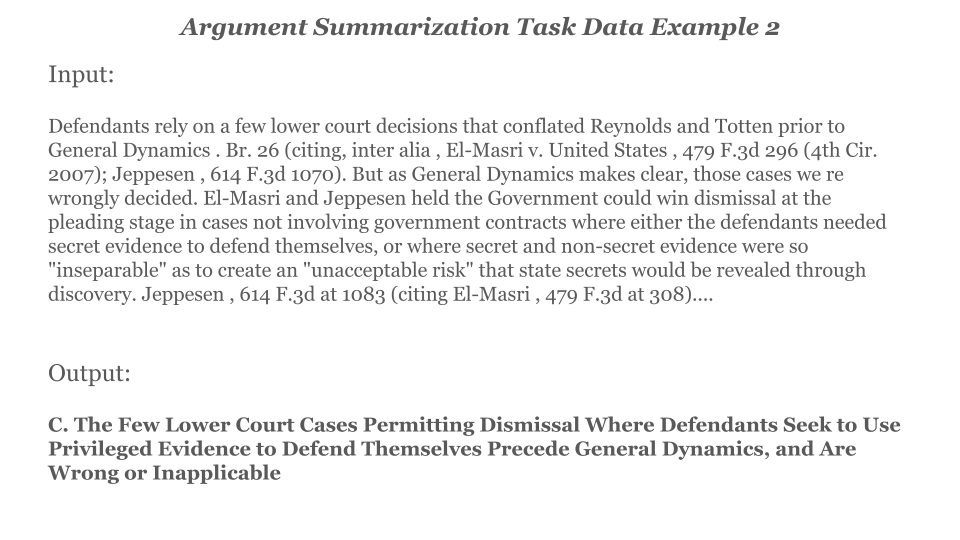}
    \caption{Summarization Task Data Example 2. Headings can themselves be long. Although they may be difficult for a non-lawyer to understand, their specificity clearly signals to judges the logical points that will be made in the text.}
    \label{fig:sum_data_2}
\end{figure*}

%% file: figures/sum_data_ex3.tex
\begin{figure*}
    \centering
    \includegraphics[width=0.99\linewidth]{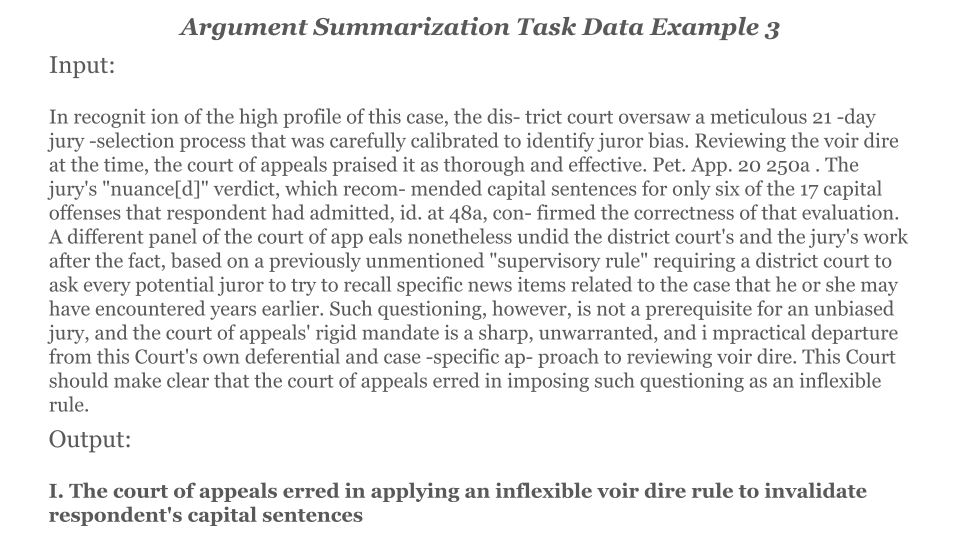}
    \caption{Summarization Task Data Example 3. Short sections often correspond to headers for top level arguments, where the text serves as an introduction to child arguments with more detailed and longer explanation.}
    \label{fig:sum_data_3}
\end{figure*}

%% file: figures/com_data_ex1.tex
\begin{figure*}
    \centering
    \includegraphics[width=0.99\linewidth]{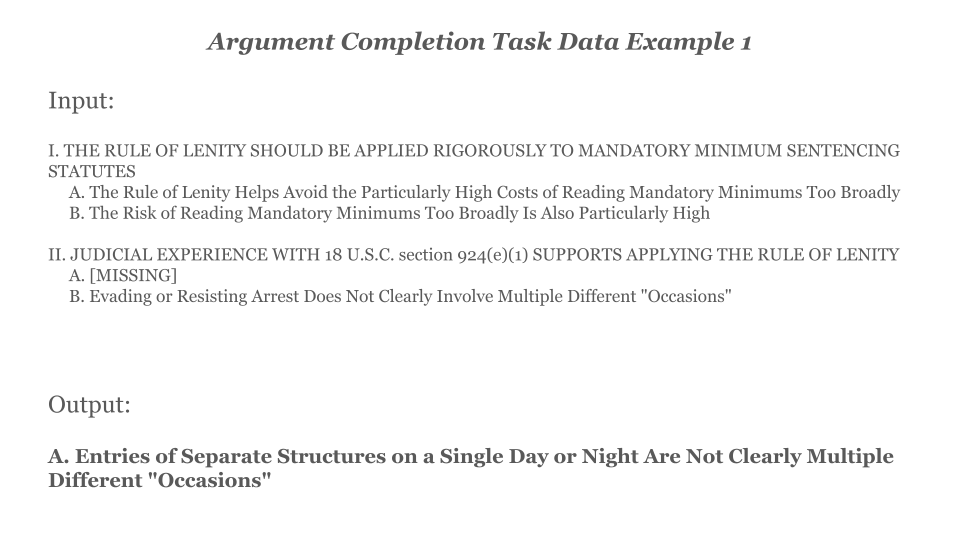}
    \caption{Argument Completion Task Data Example 1. The missing argument in II.A complements the point made in the sibling argument II.B, which is that these activities should not be construed as different "occasions" of a crime, and so should not trigger stricter penalties.}
    \label{fig:com_data_1}
\end{figure*}

%% file: figures/com_data_ex2.tex
\begin{figure*}
    \centering
    \includegraphics[width=0.99\linewidth]{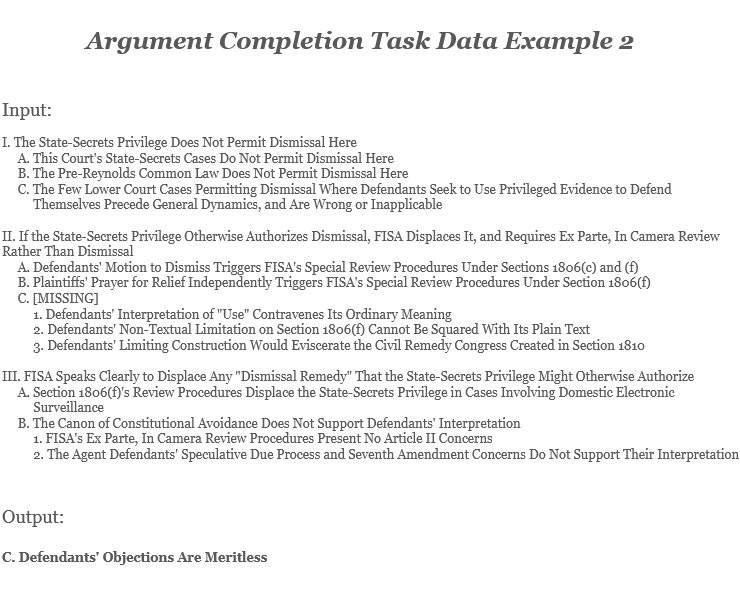}
    \caption{Argument Completion Task Data Example 2. Depending on the structure of the argument, sibling or child headers may give additional clues for the missing header.}
    \label{fig:com_data_2}
\end{figure*}

%% file: figures/com_data_ex3.tex
\begin{figure*}
    \centering
    \includegraphics[width=0.99\linewidth]{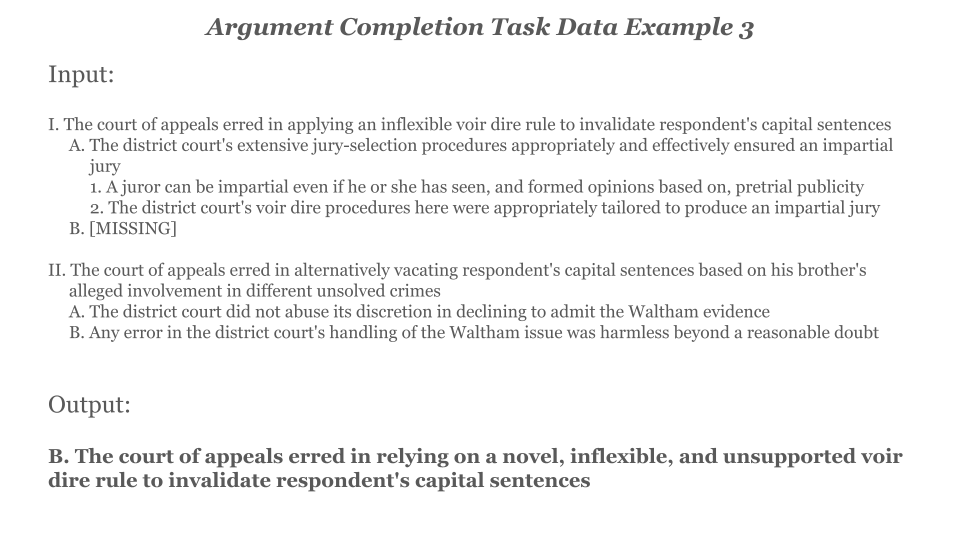}
    \caption{Argument Completion Task Data Example 3}
    \label{fig:com_data_3}
\end{figure*}

%% file: figures/com_data_ex4.tex
\begin{figure*}
    \centering
    \includegraphics[width=0.99\linewidth]{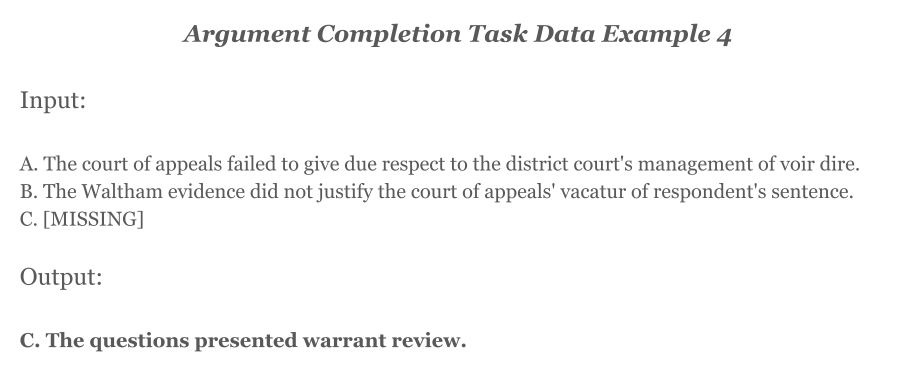}
    \caption{Argument Completion Task Data Example 4. Shorter arguments with less context are likely more difficult to complete.}
    \label{fig:com_data_4}
\end{figure*}

%% file: figures/cite_data_ex1.tex
\begin{figure*}
    \centering
    \includegraphics[width=0.99\linewidth]{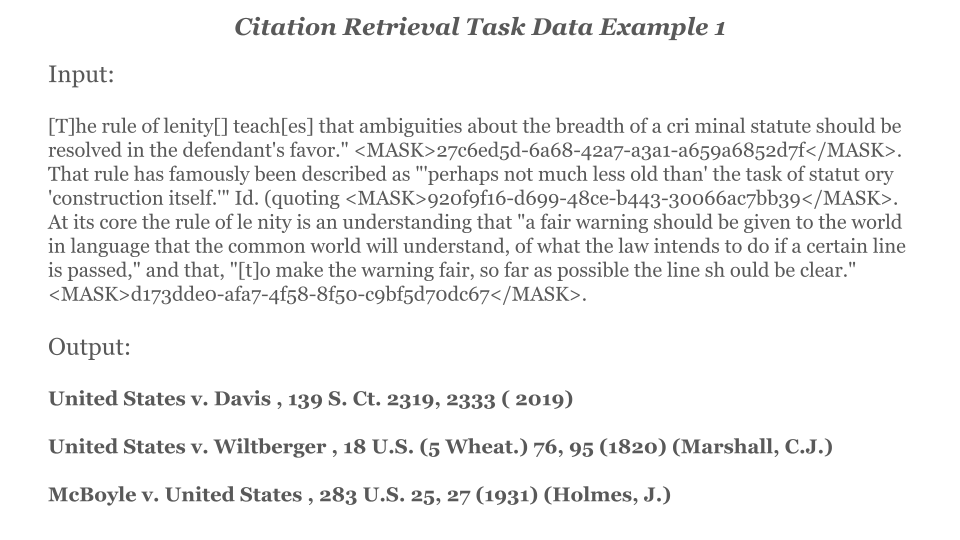}
    \caption{Citation Retrieval Task Data Example 1}
    \label{fig:cite_data_1}
\end{figure*}

%% file: figures/cite_data_ex2.tex
\begin{figure*}
    \centering
    \includegraphics[width=0.99\linewidth]{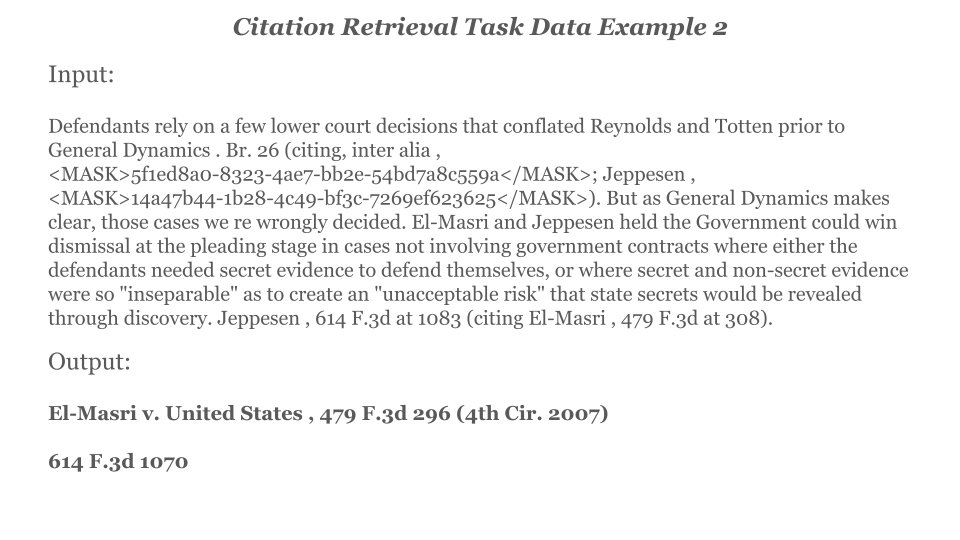}
    \caption{Citation Retrieval Task Data Example 2. The eyecite annotate\_citations() function works reasonably well but does sometimes miss citations. Because we only retrieve masked cases, cases that are missed are simply not measured.}
    \label{fig:cite_data_2}
\end{figure*}

%% file: figures/cite_data_ex3.tex
\begin{figure*}
    \centering
    \includegraphics[width=0.99\linewidth]{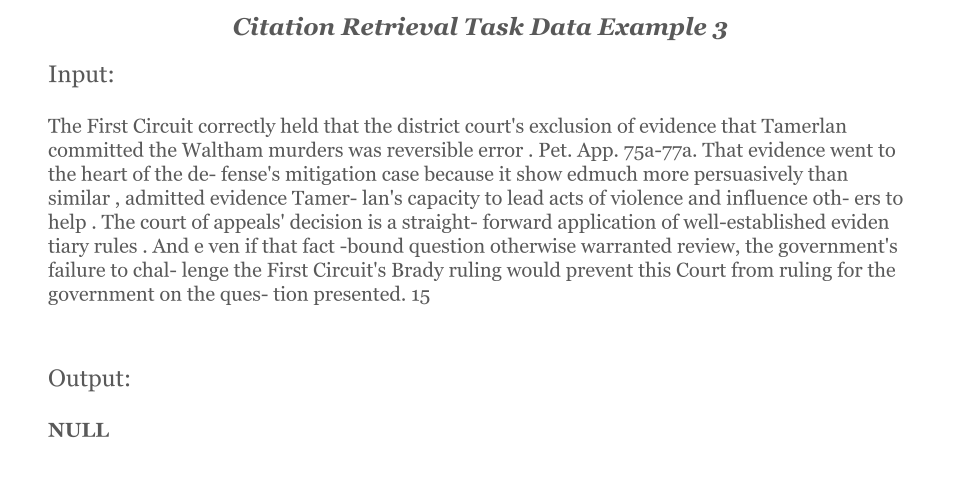}
    \caption{Citation Retrieval Task Data Example 3: The retrieval dataset only annotates citations to court cases. Citations to other types of documents (in this case to a document in the case's procedural history) are not part of the retrieval corpus. Sections of text with no relevant citations are excluded from our experiments.}
    \label{fig:cite_data_3}
\end{figure*}